\documentclass[10pt,journal,compsoc]{IEEEtran}

\ifCLASSOPTIONcompsoc
  \usepackage[nocompress]{cite}
  \usepackage{booktabs} 
  \usepackage[group-separator={,}]{siunitx}
  \usepackage{makecell}
  \usepackage[ruled]{algorithm2e}
  \usepackage{amsthm}
  \usepackage{amsmath}
  \usepackage{amsfonts}
  \usepackage{multirow}
  \usepackage{enumitem}
  \usepackage{array}
  \usepackage[skip=3pt]{caption}
  \usepackage{mathrsfs}
  \usepackage{stmaryrd} 
  \usepackage{subfigure}
  \usepackage{balance}
  \usepackage{graphicx}
  \usepackage{amsfonts}
  \usepackage{url}
  \usepackage{xspace}
  \usepackage{upgreek}
  \usepackage{color}
  \usepackage{amsmath}
  \usepackage{makecell}
  \usepackage{multirow}
  \usepackage{array}
  \usepackage{booktabs}
  \usepackage{amssymb}
  \usepackage{microtype}
\else
  \usepackage{cite}
\fi
\usepackage[edges]{forest}
\usepackage{bm}
\usepackage[pagebackref=true,breaklinks=true,letterpaper=true,colorlinks,citecolor=darkerblue,bookmarks=false]{hyperref}
\usepackage{url}

\ifCLASSINFOpdf
\else
\fi

\newcommand{\hide}[1]{} 

\newcolumntype{M}[1]{>{\centering\arraybackslash}m{#1}}

\usepackage{listings}
\usepackage{floatflt}
\usepackage{dblfloatfix}
\usepackage{makecell}
\usepackage[misc]{ifsym}
\usepackage{amsfonts}
\def\submissionarXiv{1}  
\def\submissionTPAMI{2}  
\def\submission{\submissionarXiv}  
\usepackage{soul}
\usepackage{multicol}
\usepackage{xcolor}  
\usepackage{colortbl}  
\definecolor{gray}{rgb}{0.5,0.5,0.5}
\definecolor{darkergreen}{RGB}{34, 139, 34}
\definecolor{darkerblue}{rgb}{0,0.08,0.45}
\definecolor{gray94}{gray}{.94}

\usepackage{pifont}%
\usepackage{xspace}
\usepackage{placeins}
\definecolor{maskBasic}{RGB}{255, 229, 154}
\definecolor{maskAdvanced}{RGB}{255, 230, 204}
\definecolor{encViT}{RGB}{249, 207, 204}
\definecolor{encCNN}{RGB}{225, 214, 231}
\definecolor{tarTokenizer}{RGB}{215, 232, 214}
\definecolor{tarFeatures}{RGB}{185, 224, 165}
\definecolor{tarPixel}{RGB}{151, 208, 119}
\definecolor{headContrastive}{RGB}{117, 227, 230}
\definecolor{headBoth}{RGB}{177, 221, 240}
\definecolor{headMIM}{RGB}{220, 232, 253}
\definecolor{background}{RGB}{249, 247, 237}

\newcommand{\MA}[1]{\sethlcolor{maskAdvanced}\hl{#1}}
\newcommand{\EV}[1]{\sethlcolor{encViT}\hl{#1}}
\newcommand{\EC}[1]{\sethlcolor{encCNN}\hl{#1}}
\newcommand{\TT}[1]{\sethlcolor{tarTokenizer}\hl{#1}}
\newcommand{\TF}[1]{\sethlcolor{tarFeatures}\hl{#1}}
\newcommand{\TP}[1]{\sethlcolor{tarPixel}\hl{#1}}
\newcommand{\HC}[1]{\sethlcolor{headContrastive}\hl{#1}}
\newcommand{\HB}[1]{\sethlcolor{headBoth}\hl{#1}}
\newcommand{\HM}[1]{\sethlcolor{headMIM}\hl{#1}}
\newcommand{\back}[1]{\rowcolor{background}{#1}}

\begin{document}
%

\title{Masked Modeling for Self-supervised Representation Learning on Vision and Beyond}
%
%
%
%

\author{
Siyuan Li*, Luyuan Zhang*, Zedong Wang, Di Wu, Lirong Wu, Zicheng Liu, Jun Xia, Cheng Tan,\\
Yang Liu, Baigui Sun, Stan Z. Li$^\dag$, \IEEEmembership{IEEE Fellow}
\IEEEcompsocitemizethanks{\IEEEcompsocthanksitem Siyuan Li and Luyuan Zhang are co-first authors. Stan Z. Li is the corresponding author.

\IEEEcompsocthanksitem Siyuan Li, Luyuan Zhang, Zedong Wang, Di Wu, Lirong Wu, Zicheng Liu, Jun Xia, Cheng Tan, and Stan. Z. Li are from the AI Lab, Research Center for Industries of the Future, Westlake University, Hangzhou, Zhejiang, China, 310030.\protect\\
E-mail:~lisiyuan@westlake.edu.cn;~zhangluyuan@smail.nju.edu.cn; wangzedong@westlake.edu.cn;~wudi@westlake.edu.cn;~wulirong@westlake.edu.cn;~liuzicheng@westlake.edu.cn;
junxia@westlake.edu.cn;~tancheng@westlake.edu.
cn;~stan.zq.li@westlake.edu.cn.

\IEEEcompsocthanksitem Siyuan Li, Yang Liu, and Baigui Sun are with the DAMO Academy, Hangzhou, Zhejiang, China.\protect\\
Email: ly261666@alibaba-inc.com; baigui.sbg@alibaba-inc.com.
%
}
}

%
%

\markboth{}%
{Shell \MakeLowercase{\textit{et al.}}: Bare Advanced Demo of IEEEtran.cls for IEEE Computer Society Journals}
%



\IEEEtitleabstractindextext{%
\begin{abstract}

As the deep learning revolution marches on, self-supervised learning has garnered increasing attention in recent years thanks to its remarkable representation learning ability and the low dependence on labeled data. 
Among these varied self-supervised techniques, masked modeling has emerged as a distinctive approach that involves predicting parts of the original data that are proportionally masked during training.
This paradigm enables deep models to learn robust representations and has demonstrated exceptional performance in the context of computer vision, natural language processing, and other modalities. 
In this survey, we present a comprehensive review of the masked modeling framework and its methodology. We elaborate on the details of techniques within masked modeling, including diverse masking strategies, recovering targets, network architectures, and more.
Then, we systematically investigate its wide-ranging applications across domains. 
Furthermore, we also explore the commonalities and differences between masked modeling methods in different fields. 
Toward the end of this paper, we conclude by discussing the limitations of current techniques and point out several potential avenues for advancing masked modeling research.
A paper list project with this survey is available at \url{https://github.com/Lupin1998/Awesome-MIM}.

\end{abstract}

}

\maketitle

\IEEEdisplaynontitleabstractindextext

%
\IEEEpeerreviewmaketitle

\section{Introduction}
\begin{figure*}[t!]
    \centering
    \vspace{-0.5em}
    \includegraphics[width=1.0\linewidth]{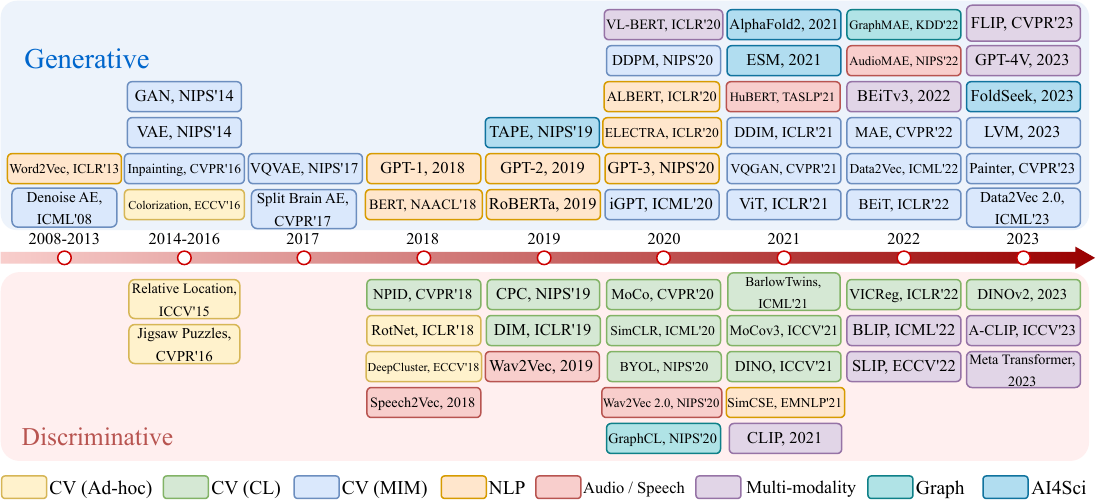}
    \caption{Research in self-supervised learning (SSL) can be broadly categorized into \textbf{Generative} and \textbf{Discriminative} paradigms. 
    We reviewed major SSL research since 2008 and found that SSL has followed distinct developmental trajectories and stages across time periods and modalities. Since 2018, SSL in NLP has been dominated by generative masked language modeling, which remains mainstream. In computer vision, discriminative contrastive learning (CL) dominated from 2018 to 2021 before masked image modeling gained prominence after 2022.
    }
    \vspace{-1.0em}
    \label{Timeline}
    \vspace{-0.5em}
\end{figure*}

\if\submission\submissionTPAMI  
    \IEEEPARstart{D}{}eep learning (DL) has made tremendous progress over the past decade, with an early emphasis on the supervised learning approaches~\cite{he2016deep, 2022convnet} that depend on labeled data.
\else
    \IEEEPARstart{D}{}eep learning (DL) has made tremendous progress over the past decade, with an early emphasis on the supervised learning approaches~\cite{he2016deep, 2017iccvmaskrcnn, 2022convnet, Li2022MogaNet} that depend on labeled data.
\fi
However, self-supervised learning (SSL) and pretraining techniques~\cite{liu2019roberta} have burgeoned, captivating the deep learning community with their advanced transferability and reduced dependence on labels. 
Fundamentally, \textit{SSL is to learn valuable representations from unlabeled data, \textit{e.g.}, intrinsic data structures, with designated pretext tasks.}
The development of SSL and pretraining techniques has been rapid, with a proliferation of variants across modalities and fields. To date, their evolutions have followed far different trajectories depending on specific modality and domain. Thus, it is crucial to provide an up-to-date survey of the rapidly growing masked modeling.
The development timeline of SSL is schematically illustrated in Figure \ref{Timeline}.

\textbf{Early Attempts}.  Due to the underwhelming results from discriminative pretext tasks, early-stage SSL methods were dominated by generative objectives. Research at that time focused heavily on generative modeling itself, such as image and text generation tasks, with pretraining treated as a byproduct rather than the major concern. Even today, generative approaches remain at the heart of SSL, including Autoencoder-based models~\cite{oord2018representation, esser2021taming}, GAN-based models~\cite{bau2018gan}, and diffusion-based models~\cite{Ho2020DenoisingDP}.
In contrast, former discriminative SSL frameworks were hinged on ad-hoc pretext tasks. Methods like~\cite{Doersch2015UnsupervisedVR}  and~\cite{Noroozi2016UnsupervisedLO} introduced other tasks like colorization and shuffle-reconstruction.~\cite{Lugmayr2022RePaintIU} pioneered the use of masked inputs for reconstruction, which served as a precursor to today's masked modeling. However, these approaches have not yet hit the mainstream.

\textbf{Language Domain}. In 2018, BERT~\cite{devlin2018bert} and GPT~\cite{hu2020gpt} introduced Masked Language Modeling (MLM) and Next Sentence Prediction (NSP) for natural language processing (NLP), ushering in more standardized objectives. Because of the remarkable performance of BERT and GPT, generative pretraining methods based on MLM and NSP have become the mainstream approaches for NLP. From 2018 to 2020, the NLP community mainly focused on refining pretraining strategies based on MLM and NSP. After contrastive learning (CL) was theoretically formalized, some 2021 works~\cite{Gao2021SimCSESC} explored discriminative pretraining for NLP. However, MLM-based research remains in a dominant position.

\textbf{Vision Domain}. In contrast to NLP, self-supervised pretraining in computer vision (CV) has followed a more complex and diverse development. In 2018, theoretical advances in  CL like~\cite{Lai2019ContrastivePC} and~\cite{Wu2018UnsupervisedFL} established their foundations, enabling significant performance gains in linear evaluation protocols. This catalyzed the rise of discriminative models for SSL in computer vision. From 2019 to 2021, CV research was dominated by contrastive approaches, with influential frameworks like \cite{he2019momentum}, \cite{chen2020simple}, and \cite{grill2020bootstrap} achieving impressive results. During this period, some generative models like iGPT~\cite{Chen2020GenerativePF} adopted auto-regressive pretraining with a GPT-2~\cite{hu2020gpt} backbone. However,  due to
performance limitations, generative SSL had minimal impact compared to CL. This changed in 2021 when Vision Transformers~\cite{Dosovitskiy2020AnII} (ViT) surpassed Convolutional Neural Networks (CNN)~\cite{he2016deep} and altered the CV self-supervision landscape. Post-ViT~\cite{Dosovitskiy2020AnII}, CV research began emulating BERT~\cite{devlin2018bert} by tokenizing images and then pretraining Transformers. MAE~\cite{he2022masked} formally introduced Masked Image Modeling (MIM), achieving strong performance. Since then, CV SSL research has focused on generative reconstruction and Masked Modeling (MM).

\textbf{Multimodality}. The earliest multimodal pre-trained models emerged in 2020, with VL-BERT~\cite{Su2019VLBERTPO} fusing modalities using a transformer architecture. In 2021, CLIP~\cite{radford2021learning} combined CV and NLP modalities, ushering in an era of CL for multimodal pretraining that became mainstream in academia. Proposed in 2022, BEiT.v3~\cite{2022BEiTV3} introduced Masked Modeling as a pretraining technique for multimodal models, while MetaTransformer~\cite{Zhang2023MetaTransformerAU} combined multiple approaches. Since then, Masked Modeling has played a pivotal role in multimodal research.

\textbf{Other Domains}. SSL has been broadly applied across modalities beyond NLP and CV, including Audio, Speech, Biology, Video, and others. Research on SSL pretraining for \textbf{Audio and Speech} has closely followed the paradigms in CV and NLP. When CL gained popularity in 2018, influential speech models like \cite{Chung2018Speech2VecAS} and \cite{nips2020wav2vec2} adopted CL for pretraining. Notably, \cite{nips2020wav2vec2} combined masked modeling as a data augmentation technique for CL. In 2021, \cite{2021mam} and then \cite{huang2022masked} in 2022 drew inspiration from masked image modeling in CV to implement masked spectrum modeling for audio. Since then, Masked Modeling has been a main direction in audio and speech research.
As AlphaFold~\cite{jumper2021highly} achieved a great breakthrough in accurate protein structure predictions in 2021s, masked modeling has been introduced into \textbf{Biology} and \textbf{Chemistry} to assist the scientists as the AI-for-Science (\textbf{AI4Sci}) research paradigm.

Masked Modeling has demonstrated compelling performance across modalities, including vision, language, speech, and beyond. With its widespread adoption, the landscape of Masked Modeling research has grown increasingly diverse. A multitude of masked modeling methods have emerged, creating a complex ecosystem of models tailored to different data types and tasks. Therefore, it is highly worthwhile to systematically review recent advances and provide structured categorization of the extensive Masked Modeling literature. In this paper, we conduct an extensive survey of the Masked Modeling research landscape. We thoroughly investigate the latest innovations in self-supervised representation learning across vision, NLP, speech, and other domains. Our main contribution is a comprehensive taxonomy that organizes the extensive body of Masked Modeling techniques into coherent groups according to training objectives, model architectures, and applications. This framing elucidates the relationships between existing methods and paves the way for developing new Masked Modeling techniques. Our review and classification provide a holistic reference to inform and accelerate future Masked Modeling research across modalities.

\begin{figure*}[t!]
    \centering
    \vspace{-1.0em}
    \includegraphics[width=0.90\linewidth]{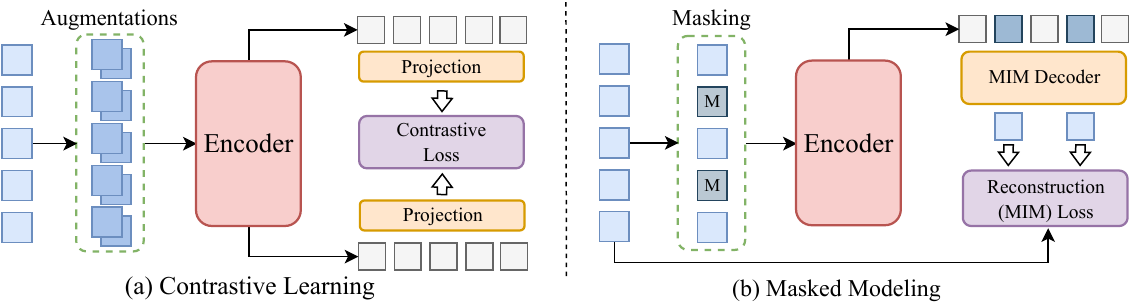}
    \vspace{-0.5em}
    \caption{Illustration of two popular self-supervised learning (SSL) frameworks. For simplicity, the input data can be serialized and transformed into a sequence of embedded tokens. (a) Contrastive learning (CL) learns discriminative representation from two augmented views of input data sequences by aligning two projected tokens. (b) Masked Modeling learns contextual information by the generative paradigm that reconstructs the masked tokens.
    }
    \label{fig:general_framework}
    \vspace{-1.0em}
\end{figure*}

Overall, compared to two published surveys~\cite{ijcai2023MIMsurvey, ieee2023MAEsurvey} on MIM, our contributions include:
\begin{itemize}[leftmargin=1.15em]
\vspace{-0.25em}
\item We provide a timely literature review and a comprehensive framework, taking CV as an instance, to holistically conceptualize Masked Modeling principles that can categorize different applications to date across domains and modalities under a common lens. 
\item We meticulously review and discuss the technical details within the Masked Modeling framework, such as masking strategies, targets, networks, and more, to let researchers get a better grasp of the involved techniques and thus gain a deeper understanding and insights. 
\item We systematically survey the downstream applications of Masked Modeling in vision, presenting the technical challenges and further showcasing their widespread applicability to other modalities and domains beyond vision, such as audio, speech, graph, biology, and more.
\item Through extensive algorithmic research and detailed evaluations, we provide a collection of comprehensive tables and awesome lists of masked modeling methods on GitHub. In the end, we identified the future directions of masked modeling research and further provided heuristic suggestions and reflections.
\vspace{-0.5em}
\end{itemize}

\section{Preliminary}
\label{section:Preliminary}

\subsection{Notations}
The notations used in this survey are provided in Table~\ref{tab:notions}.
In this paper, $x$ denotes a data sequence, such as a sentence in NLP, a patch sequence in CV, or a data sequence in another modality.
In CV tasks, $\pmb{x} = \in [\pmb{x}_i]_{i=1}^{N} \mathbb{R}^{N\times (P^2\times C)}$ denotes an image with $N$ patches, where $P^2$ is the patch resolution and $C$ denotes the embedding dimension.
In this paper, $\pmb{x}^k$ and $\pmb{x}_i^{k}$ denote the different sequences and patches, and $\pmb{x}^{v_i}$ denotes the different augmented views of the sequence.
In NLP tasks, $\pmb{x}=[\pmb{x}_i]_{i=1}^{L}$ denotes the original sentence and $\pmb{e}=[\pmb{e}_i]_{i=1}^L$ presents the embedded sequence.
Encoder and decoder are denoted as $f_{\theta}(\cdot)$ and $g_{\phi}(\cdot)$, where $\theta$ and $\phi$ are learnable parameters.
In masked modeling tasks, as some tokens or patches of $\pmb{x}$ are selected to mask, we use $\mathcal{M}=\{0,1\} ^{N}$ to present the mask set. A masked sequence can be written as $\pmb{x}\odot \mathcal{M}=[\pmb{x_1},\cdots,\pmb{x_{i-1}},0,\pmb{x_{i+1}},\cdots,\pmb{x_n}]$. The visible patches or tokens can be denoted as $\tilde{x}=x_{i=1,\mathbb{I}_{\mathcal{M}=1}}^N$ or $\tilde{e}=e_{i=1,\mathbb{I}_{\mathcal{M}=1}}^N$.

\subsection{Self-Supervised Learning}
\begin{figure*}[t]
    \centering
    \vspace{-0.5em}
\begin{minipage}{0.485\linewidth}
    \centering
    \vspace{-0.5em}
    \includegraphics[width=1.0\linewidth]{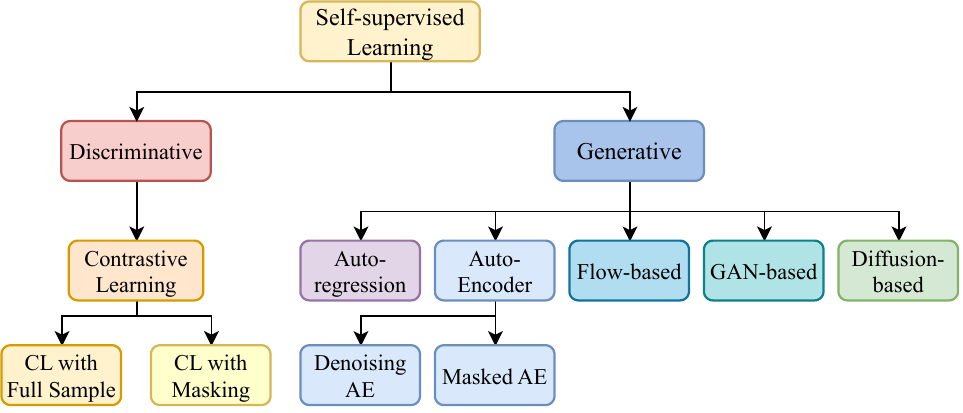}
    \caption{SSL is universally divided into generative and discriminative~\cite{liu2021self}. The generative model can be divided into AR, AE, Flow-based, GAN-based, and diffusion-based models where the AE model can be divided into Denoised AE and Masked AE.
    This survey is focused on AR and AE models for SSL and relevant tasks.
    }
    \label{fig:example}
    \vspace{-0.5em}
\end{minipage}
\begin{minipage}{0.51\linewidth}
    \centering
    \setlength{\tabcolsep}{0.3mm}
\resizebox{1.01\linewidth}{!}{
\begin{tabular}{ll|ll}
    \toprule
\multicolumn{2}{l|}{\textbf{Basic Notations}}                               & \multicolumn{2}{l}{\textbf{Functional Notations}}                            \\ \hline
$\mathbb{R}^{m\times n}$         & Two-dimensional tensor space             & $\pmb{q}_{\phi}(\cdot|\cdot)$  & The quantization tokenizer                  \\
$\mathbb{R}^{m\times n\times p}$ & Three-dimensional tensor space           & $\pmb{p}_{\psi}(\cdot|\cdot)$  & The decoder to train the tokenizer          \\
$\mathcal{N}$                    & Natural number set from $1$ to $N$       & $f_{\theta}(\cdot)$            & Encoder with parameter $\theta$             \\
$ \pmb{x}$, $\pmb{x}_i$          & A data sequence and its $i$-th element   & $f_{\hat{\theta}}(\cdot)$      & Teacher model with parameter $\hat{\theta}$ \\
$\pmb{x}_{m:n}$                  & The subsequence in $ \pmb{x}$            & $g_{\theta}(\cdot )$           & Decoder of with parameter $\theta$          \\
$\pmb{m}$                        & Encoding of masked patch/token           & $\nabla(\cdot)$                & Gradient function                           \\
$\pmb{z}$                        & A latent-space variable (feature)        & $\mathcal{T}(\cdot)$           & The transformation function                 \\
$\mathcal{M}=\{ 0,1\}^N$         & A set of masks for $N$ elements          & $\mathbb{I}_{(\cdot)}$         & An indicator function                       \\
$\mathcal{M}_i$                  & The $i$-th element in set $\mathcal{M}$  & $ \mathcal{G}(\cdot)$          & Adversarial training generator              \\
$\tilde{\pmb{x}}$                & A set of visible tokens after masking    & $\mathcal{D}(\cdot)$           & Adversarial training discriminator          \\
$\theta,\omega, \gamma, \cdots $ & Parameters of the deep networks          & $\mathcal{F}(\cdot)$           & Fourier transform function                  \\
$\tau$                           & Temperature parameter in CL              & $p(\cdot)$                     & Probability density function                \\
$\lambda$                        & Weights of loss functions                & $p(\cdot|\cdot)$               & Conditional probability distribution        \\ \cline{1-2}
\textbf{NLP}                     &                                          & $ \textrm{sg}(\cdot)$          & Stop-gradient operation                     \\
$\pmb{e}$                        & Embedded word tokens.                    & $\langle \cdot, \cdot \rangle$ & Inner product function                      \\
$\mathcal{V}$, $v_i$             & Vocabulary set and its $i$-th elements   & $ | \cdot |$                   & Cardinality of the set                      \\ \cline{1-2}
\textbf{CV}                      &                                          & $ \| \cdot \| $                & Norm of the vector                          \\
$\mathbf{X} $                    & Images X                                 & $ \mathcal{S}$                 & Similarity measurement function             \\
$\mathbf{X}^v $                  & Multiple views of the image $\mathbf{X}$ & $(\cdot)^{T}$                  & Transpose function                          \\
$\pmb{x}^{v_i}$                  & Patch sequence with multiple views.      & $\odot$                        & Element-wise multiplication                 \\
    \bottomrule
\end{tabular}
    }
    \caption{Mathmetical notations.}
    \label{tab:notions}

\end{minipage}
    \vspace{-1.0em}
\end{figure*}

This section will give a brief introduction to SSL methods, which are universally divided into two categories, \textit{i.e.}, generative and discriminative, as shown in Figure~\ref{fig:example}. Our classification on SSL is based on \cite{liu2021self}. 

\textbf{Generative model} usually encodes the input $x$ into a latent variable $z$ and decodes the latent variable $z$ to reconstruct the input $x$ with an encoder-decoder architecture. Auto-regressive models typically model a series of regressions one by one for one input.

\textbf{Auto-Regressive} (AR) models typically model a series of regressions one by one for one input, where the current output depends on the previous inputs or outputs in the sequence. \textbf{GPT}~\cite{hu2020gpt} and \textbf{Transformer}~\cite{vaswani2017attention} are AR models. The learning object of the AR model can be formulated as:
\begin{align}
    \max\limits_{\theta} p_{\theta}(\pmb{x}) = \sum_{t=1}^{T} \log p_{\theta}(\pmb{x}_t | \pmb{x}_{1:t-1}),
    \label{eq:AR}
\end{align}

\textbf{Auto-Encoder} (AE) reconstructs the input from the corrupted input. 
The learning object of the AE model is:
\vspace{-0.25em}
\begin{align}
    \min \mathcal{L}\big(\pmb{x},g_{\textrm{dec}}(f_{\textrm{enc}}(\pmb{x}))\big).
    \label{AE}
    \vspace{-0.5em}
\end{align}
We further divide the AE model into \textbf{Denoising AE} and \textbf{Masked AE}. The \textbf{Denoising AE} model is trained to reconstruct clean data from noisy or corrupted input. By removing noise or corruption, the model learns robust representations. And a \textbf{Masked AE} is trained to predict missing or masked portions of the input data. By reconstructing the missing parts, the model learns contextual representations.

\textbf{Flow Based} model aims to learn densities $p(x)$ from data. 
Suppose a latent variable $z$ follows a known distribution $p_Z(x)$ and define $z=f_{\theta}(x)$. The learning objective is to maximize the likelihood:
\begin{equation}
\begin{split}
&\max\limits_{\theta} \sum\limits_{i} \log p_{\theta} (x^{(i)}) \\
= &\max\limits_{\theta} \sum\limits_{i} \log p_{Z} (f_{\theta}(x^{(i)})) 
+ \log \bigg|\frac{\partial f_{\theta}}{\partial x} (x^{(i)}) \bigg|.
\end{split}
\label{Flow}
\end{equation}

\textbf{GAN-Based} model (adversarial learning)   involves training two models in competition with each other, typically a generator $\mathcal{G}$ and discriminator $\mathcal{D}$. The learning object is: 
\begin{equation}
  \begin{split}
\min_{\mathcal{G}} \max_{\mathcal{D}} V(\mathcal{D}, \mathcal{G}) &= 
     \mathbb{E}_{x \sim p_{\text{data}}(x)}\big[\log \mathcal{D}(x)\big]\\
     &+ \mathbb{E}_{z \sim p_z(z)}\big[\log(1 - \mathcal{D}(\mathcal{G}(z)))\big].
     \end{split}
     \label{GAN}
\end{equation}

\textbf{Diffusion-based} model initially processes images through a series of Gaussian noise treatments, followed by restoration of the image through the model. The diffusion-based model process is divided into forward and reverse processes. The forward process treats the image with cumulative Gaussian noise, which can be modeled as follows:
\begin{align}
\begin{split}
    q(x_t|x_{t-1}) & =\mathcal{N}(x_t;\sqrt{1-\beta_t}x_{t-1},\beta_t\mathbf{I}), q(x_{1:T}|x_0)\\
    & =\prod_{t=1}^T q(x_t|x_{t-1}),
\end{split}
\end{align}
in which $\beta_t$ is mean coefficient. The reverse process of the diffusion-based model, which involves denoising and inference, has a learning objective as follows:
\begin{align}
    p_\theta(X_{0:T})&=p(x_T)\prod_{t=1}^T p_\theta(x_{t-1}|x_t);\\ p_\theta(x_{t-1}|x_t)&=\mathcal{N}(x_{t-1};\mu_\theta(x_t,t),\Sigma_\theta(x_t,t)). 
\end{align}

\textbf{Discriminative model} are typically formulated using CL objectives. The core idea in CL is to train encoders to produce similar representations for semantically related instances while distinguishing unrelated samples~\cite{liu2021self}. 
Contrasting at the \textbf{context-instance} level involves comparing the local feature, which is encoded, with the global representation from the identical sample. In contrast, the \textbf{instance-instance} contrast method is more focused on the representation at the instance level, examining the commonalities across multiple samples~\cite{liu2021self}. \textbf{InfoNCE}~\cite{oord2018representation} is the basic learning objective:
\begin{equation}
    \hspace{-0.85em}
    \mathcal{L}_{\text{infoNCE}} = -\mathbb{E}_{(\pmb{x}^i, \pmb{x}^j) \sim p(\pmb{x})} \left[ \frac{\exp(f(\pmb{x}^i)^T f(\pmb{x}^j) / \tau)}{\sum_{k=1}^{K} \exp(f(\pmb{x}^i)^T f(\pmb{x}^k) / \tau)} \right].
    \label{Con}
\end{equation}

\vspace{-0.5em}
\subsection{Masked Modeling}
\textbf{Masked Language Modeling}.
MLM was first introduced in BERT. The central idea of MLM is to randomly mask tokens within a sentence and replace them with a Mask vector. The encoder then predicts the masked vector. We formally define the problem of MLM as follows: 
A sentence $ \pmb{x}=[\pmb{x_i}]_{i=1}^{L}$ is first tokenized as $\pmb{e} = [\pmb{e_i}]_{i=1}^{L}$ through a tokenizer $\pmb{q}_{\phi}(\cdot|\cdot)$, in which $L$ denotes the number of the tokens in this sentence. The masked sequence of the embedded sentence  $\pmb{e}\odot \mathcal{M}$ is fed into a Transformers encoder $f_{\theta}(\cdot)$.  $m_i=f_{\theta}(\tilde{e})$ is the hidden state of the last layer at the masked position and can be regarded as a fusion of contextualized representations of surrounding tokens. And the MLM task is ~\cite{lan2019albert} :
\begin{equation}
    \hspace{-1.0em}
    \mathcal{L}_{\textrm{MLM}}(x)= -\frac1{\|\mathcal{M}\|}\sum_{i\in\mathcal{N}}\mathbb{I}_{ \{\mathcal{M}_i=1\}}\log\frac
    {\exp(m_i \cdot e_i)}
    { \sum_{k=1}^{|\mathcal{V}|} \exp(m_i \cdot e_k)},
    \label{mlm}
\end{equation}

\textbf{Masked Image Modeling}.
The core concept of  MIM aligns with that of MLM. It involves masking certain pixel regions of the input image and reconstructing the original image based on the unmasked portions. Given that images lack the tokenizer structure inherent in natural language, the intuitive approach is to reconstruct pixel values directly. However, due to the high redundancy and dimensionality of image pixel information, pixel-level reconstruction is often challenging. This has historically hindered the progress of MIM. It wasn't until the introduction of the ViT, which segments images into patches that MIM began to emerge as a feasible approach. We formally define the problem of MIM as follows: A image $ \mathbf{X} \in \mathbb{R}^{H \times W \times C}$ is partitioned into multiple patches $\pmb{x} \in \mathbb{R} ^{N\times (P^2C)}$, $\pmb{x}=[\pmb{x_i}]_{i=1}^N$ where $N$ denotes the number of patch. Masked sequence can be denoted as $\pmb{x}\odot \mathcal{M}$. The remaining unmasked patches $\tilde{\pmb{x}}$ is used to reconstruct the original pixel through an encoder$f_{\theta}(\cdot)$ and a decoder $g_{\theta}(\cdot)$. We use $m_i$ to denote the hidden layer at the masked portion as NLP and $m_i=f_{\theta}(\tilde{x})$, The learning object is:
\begin{align}
\mathcal{L}_{\textrm{MIM}}= \frac{1}{\|\mathcal{M}\|}\sum_{i \in \mathcal{N}} \mathbb{I}_{ \mathcal{M}_{i}=1}\|m_i-\pmb{x_i}\|^2. 
\end{align}

\begin{figure*}[t!]
    \centering
    \vspace{-0.5em}
    \includegraphics[width=0.95\linewidth]{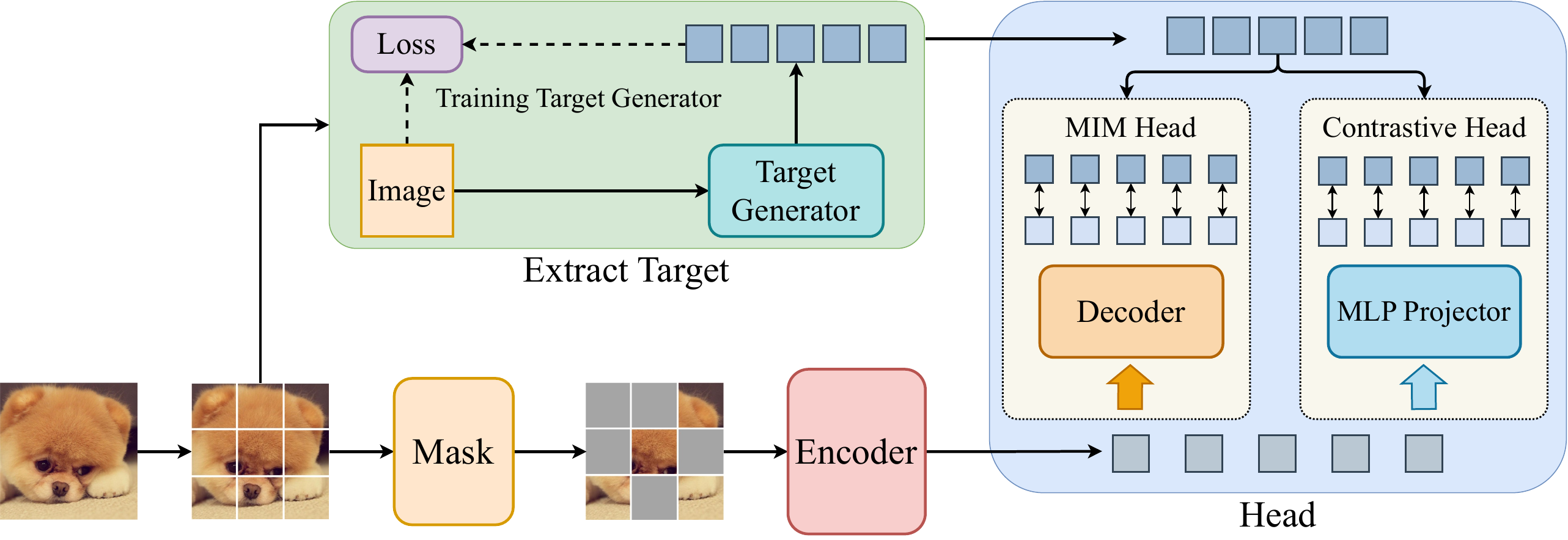}
    \caption{The overview of the basic MIM framework, containing four building blocks with their internal components and functionalities. All MIM research can be summarized as innovations upon these four blocks, \textit{i.e.}, Masking, Encoder, Target, and Head. The general frameworks of masked modeling for other modalities are similar to this framework.
    }
    \vspace{-1.0em}
    \label{overview}
    \vspace{-0.5em}
\end{figure*}

\textbf{Beyond}. Beyond CV and NLP, Masked Modeling can also be applied to various data structures and multimodal domains. The core idea is to mask parts of the input vector with mask tokens and then reconstruct the data through an encoder-decoder framework. 
Masked Data Modeling can be formally described as:
given an input sequence $x$ of any modality, we generate the corrupted sample $x\odot \mathcal{M}$ by replacing elements in $x_m$ with mask tokens [MASK]. We use $\mathcal{S}(\cdot,\cdot)$ to denote the similarity between the predicted mask tokens and the original data. The learning object is:
\begin{align}
\mathcal{L}_{\textrm{MDM}}= \frac{1}{\|\mathcal{M}\|} \sum_{i \in \mathcal{N}} \mathbb{I}_{\{\mathcal{M}_i=1\}}\mathcal{S}(m_i,x_i).
\end{align}

\section{Basic framework: A unified perspective}
\label{sec:basic}

This section will introduce a unified perspective for Masked Modeling, offering a comprehensive categorization of Masked Modeling research. 
\textbf{Since MM has been most thoroughly explored and developed in CV with the most extensive techniques and has laid the foundation for developments across domains, this survey takes MIM as an example to elucidate Masked Modeling from the perspective of CV.}

\subsection{A Unified Perspective}
Based on the current research on MIM for SSL pre-training, this paper conducts an in-depth investigation. It proposes a unified research framework and paradigm for MIM, providing a detailed classification of existing studies. 
The framework mainly consists of four modules, namely: \textbf{Mask}, \textbf{Target}, \textbf{Encoder}, and \textbf{Head}. An overview of our framework is visually presented in Figure~\ref{overview}.

\begin{itemize}[leftmargin=1.15em]
    \item \textbf{Mask}: Mask module is to generate a mask set $\mathcal{M}$ for the masked image $\pmb{x}\odot \mathcal{M}$. Typical mask strategies include Random Mask, Attention Mask, Contextual Mask, \textit{etc}.
    \item \textbf{Target}: The Target module's role is to generate supervisory signals. The target module can be formulated as: $\mathcal{T}(f_{\omega}(\pmb{x}))$, $f_{\omega}(\cdot)$ is a model with parameter $\omega$. Within this module, tokenizers like VQ-GAN~\cite{esser2021taming} and dVAE~\cite{Ramesh2021ZeroShotTG} can be utilized as tools to extract these signals, and different supervision targets can lead to different model preferences.
    \item \textbf{Encoder}: The Encoder $f_{\theta}(\cdot)$ is the target for pre-training and can adopt various network architectures (\textit{e.g.}, Transformer, CNN, or a hybrid of both). The encoder's input can be visible patches and both visible and masked patches.
    \item \textbf{Head}: The Head module is to compute losses between the supervisory signals and the predictions. The primary task of MIM is to predict the original tokens, so the most widely used head is the MIM head to reconstruct the original image or features. Meanwhile, combining with the Contrastive head can also enhance the MIM performance.
\end{itemize}
Based on the unified perspective we proposed, the MIM problem can be mathematically represented as:
\begin{equation}
    \mathcal{L}_{\textrm{MIM}} = \mathcal{S} (\mathcal{T}_1(f_{\omega}(\pmb{x})),\mathcal{T}_2(g_{\gamma}(f_{\theta}(\pmb{x}\odot \mathcal{M})))).
\end{equation}
Permuting and combining these four modules, we have meticulously categorized the research on MIM. The detailed classification is elaborated in Figure \ref{tab:mim_classification}.

\if\submission\submissionarXiv  
\begin{figure}[htb!]
    \centering
    \vspace{-0.5em}
    \includegraphics[width=1.0\linewidth]{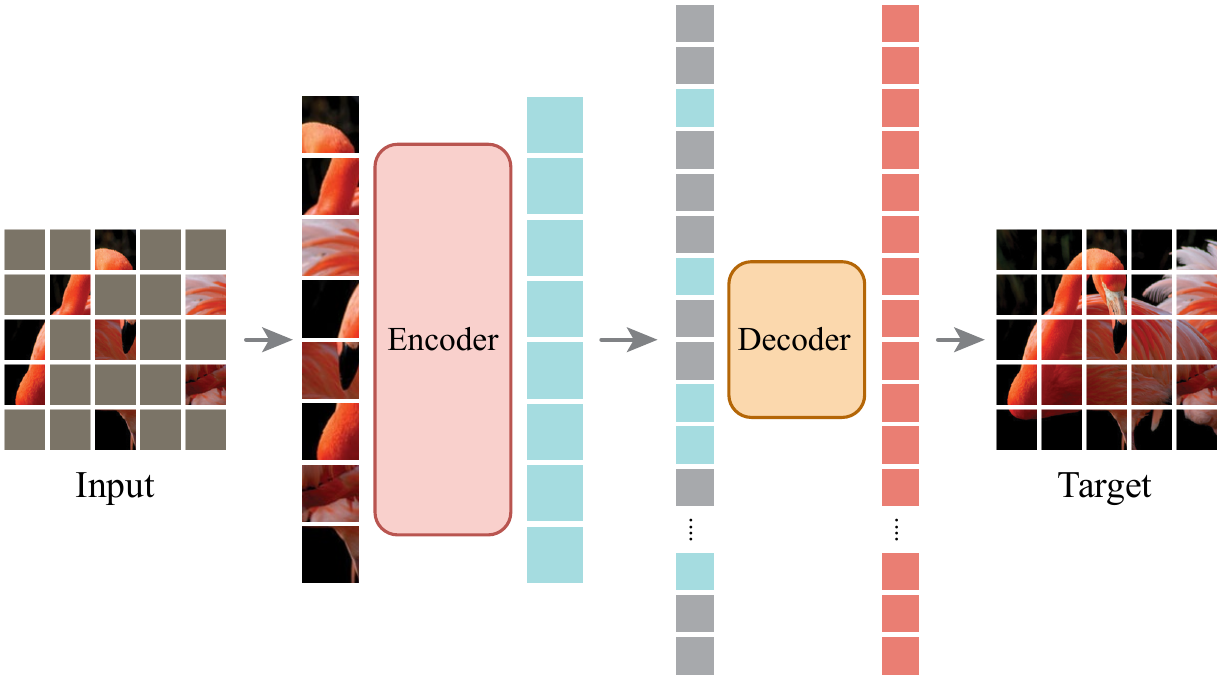}
    \caption{MAE proposed a basic framework for MIM pre-training, where the visible patches are encoded while the encoded features are decoded together with masked patches to reconstruct the pixel. The figure is reproduced from \cite{he2022masked}.}
    \label{fig:MAE}
    \vspace{-0.5em}
\end{figure}
\fi

\subsection{Basic Framework}

\textbf{iGPT}~\cite{Chen2020GenerativePF}: The input image $\mathbf{X}$, when arranged according to pixel values and subsequently downsampled, forms a pixel sequence $\pmb{x}$ that is fed into a Transformer structure identical to GPT-2~\cite{hu2020gpt}. This model predicts the value of the next pixel $\pmb{x}_t$ based on the current pixel value $\pmb{x}_{1:t}$. Given that iGPT predicts pixel values in sequence, its masking approach can be considered as ``\textbf{Basic Masking}``, with the target being the \textbf{Token}. Based on GPT, the encoder of the iGPT is \textbf{Transformer}, and the decoder is a \textbf{Linear MIM Head}. The loss of iGPT can be formulated as Eq.~\ref{eq:AR}.

\textbf{MAE}~\cite{he2022masked}: 
\if\submission\submissionarXiv  
    The overview of MAE can be seen in Figure~\ref{fig:MAE}.
\fi
The input image $ \mathbf{X} \in \mathbb{R}^{H \times W \times C}$ is partitioned into multiple patches $\pmb{x} \in \mathbb{R} ^{N\times (P^2C)}$, where approximately 75\% of the patches are \textbf{Randomly Masked}. The remaining unmasked patches $\tilde{x}$ are then fed into the \textbf{Transformer Encoder} $f_{\theta}(\cdot)$ which generates the features. These features, in conjunction with the masked patches, are input into the \textbf{Transformer Decoder} $g_{\omega}(\cdot)$ to reconstruct the \textbf{Pixels} of the original image. The quality of the reconstruction is measured using the MSE loss function in MAE:
\begin{align}
   \frac{1}{\|\mathcal{M}\|}\|g_{\omega}(f_{\theta}(\tilde{x})))-\tilde{\pmb{x}}\|^2.
   \label{eq:MAE}
\end{align}

\begin{table}[ht]
    \vspace{-1.0em}
    \centering
\resizebox{\linewidth}{!}{
    \begin{tabular}{lcc}
    \toprule
    Model    & MAE                    & iGPT              \\ \hline
    Mask     & Basic (Random)         & Basic (AR Mask)   \\
    Encoder  & Transformer            & Transformer       \\
    Target   & Pixel                  & Token             \\
    Head     & MIM Head (Transformer) & MIM Head (Linear) \\ \hline
    Category & BTPM                   & BTTM              \\
    Type     & AE                     & AR                \\
    \bottomrule
    \end{tabular}
    }
    \caption{Four parts of iGPT and MAE based on the basic framework. As two typical MIM methods, iGPT is based on AR while MAE represents Masked AE.}
    \label{table:MAEandiGPT}
    \vspace{-0.5em}
\end{table}

MAE and iGPT represent two typical basic frameworks in MIM research: iGPT is based on the AR paradigm like GPT~\cite{hu2020gpt}, while MAE is grounded in the Masked AE paradigm like BERT~\cite{devlin2018bert}. 
The four modules of iGPT include: \textbf{Basic Masking(Auto-Regressive Masking) + Transformer + Tokenizer + MIM Head}, whereas MAE is \textbf{Basic Masking (Random) + Transformer + Pixel + MIM Head}.
Table~\ref{table:MAEandiGPT} summarizes the difference between iGPT and MAE.

\begin{figure*}[t!]
    \vspace{-1.0em}
    \centering
    \includegraphics[width=1.0\linewidth]{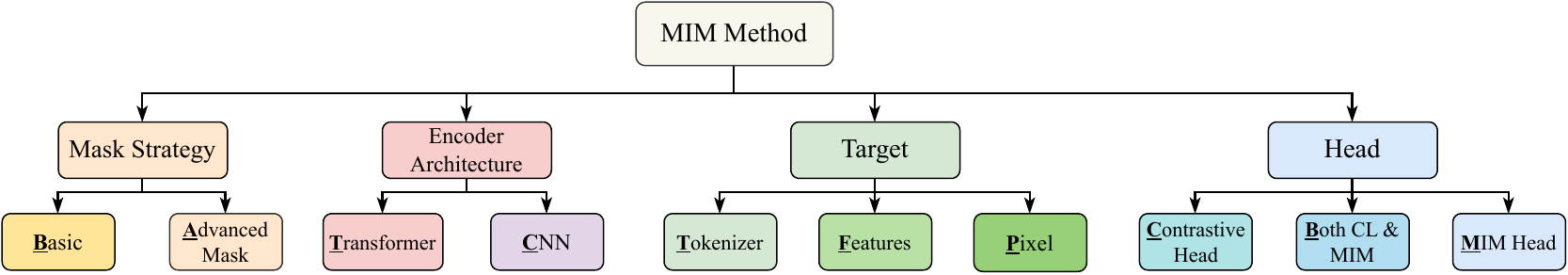}
    \vspace{-1.0em}
    \label{Class}
    \vspace{-0.5em}
\end{figure*}
\begin{table*}[t]
    \vspace{-0.5em}
    \centering
    \setlength{\tabcolsep}{0.3mm}
\resizebox{\linewidth}{!}{
\begin{tabular}{cccccccccccccccc}
    \normalsize{\HB{~B~}\EV{~T~}\TP{~P~}\HM{~M~}} &
    \normalsize{\HB{~B~}\EV{~T~}\TF{~F~}\HM{~M~}} &
    \normalsize{\HB{~B~}\EV{~T~}\TT{~T~}\HM{~M~}} &
    \normalsize{\HB{~B~}\EV{~T~}\TT{~T~}\HB{~B~}} &
    \normalsize{\HB{~B~}\EV{~T~}\TF{~F~}\HB{~B~}} &
    \normalsize{\HB{~B~}\EV{~T~}\TP{~P~}\HC{~C~}} &
    \normalsize{\MA{~A~}\EV{~T~}\TF{~F~}\HB{~B~}} &
    \normalsize{\MA{~A~}\EV{~T~}\TP{~P~}\HM{~M~}} &
    \normalsize{\MA{~A~}\EV{~T~}\TF{~F~}\HM{~M~}} &
    \normalsize{\HB{~B~}\EC{~C~}\TP{~P~}\HM{~M~}} &
    \normalsize{\HB{~B~}\EV{~T~}\TP{~P~}\HB{~B~}} &
    \normalsize{\HB{~B~}\EV{~T~}\TF{~F~}\HC{~C~}} &
    \normalsize{\MA{~A~}\EV{~T~}\TP{~P~}\HC{~C~}} &
    \normalsize{\MA{~A~}\EV{~T~}\TF{~F~}\HC{~C~}} \\
%
\back{ \begin{tabular}[l]{@{}l@{}} MAE\cite{he2022masked}\\ SimMIM\cite{Xie2021SimMIMAS}\\ RePre\cite{Wang2022RePreIS}\\ DMAE\cite{Wu2022DenoisingMA}\\ RCMAE\cite{Lee2022RCMAE}\\ RMAE\cite{Nguyen2023RMAERM}\\ Hiera\cite{icml2023Hiera}\\ BootMAE\cite{Dong2022BootstrappedMA}\\ SdAE\cite{Chen2022SdAESM}\\ TTT-MAE\cite{Gandelsman2022TestTimeTW}\\ MaskVLM\cite{Kwon2022MaskedVA}\\ MAE-lite\cite{Wang2022ACL}\\ ...
\end{tabular}}
&
\begin{tabular}[l]{@{}l@{}}CAE\cite{Chen2022ContextAF}\\ SIM\cite{Tao2022SiameseIM}\\ dBOT\cite{liu2022dBOT}\\ MaskDistill\cite{2022maskdistill}\\ CAE.V2\cite{Zhang2022CAEVC}\\ FastMIM\cite{Guo2022FastMIM}\\ Data2Vec\cite{Baevski2022data2vecAG}\\ MFM\cite{2022MFM}\\ MP3\cite{Casas2021MP3AU}\\ MaskFeat\cite{Wei2021MaskedFP}\\ MultiMAE\cite{Bachmann2022MultiMAEMM}\\ ...
\end{tabular}
&
\begin{tabular}[l]{@{}l@{}}iGPT\cite{Chen2020GenerativePF}\\ iBOT\cite{iclr2022ibot}\\ BEiT\cite{Bao2021BEiT}\\ BEiT.V2\cite{2022BEiTV2}\\ BEiT.V3\cite{2022BEiTV3}\\ MaPeT\cite{Baraldi2023LearningTM}\\ RandSAC\cite{Hua2022SelfsupervisionTR}\\ MaskGIT\cite{Chang2022MaskGITMG}\\ CIM\cite{Zheng2022CIMCI}\\ mcBEiT\cite{Li2022mcBEiTMD}\\ MVP\cite{Wei2022MVPMV}\\ PeCo\cite{Dong2021PeCoPC}\\ ...
\end{tabular}
&
\begin{tabular}[l]{@{}l@{}}MAGE\cite{cvpr2023mage}\end{tabular}
&
\begin{tabular}[l]{@{}l@{}}MaskCLIP\cite{2022MaskCLIP}\\ Ge2AE\cite{Liu2022TheDI}\end{tabular}
&
\begin{tabular}[l]{@{}l@{}}ConMIM\cite{2022ConMAE}\\ LayerGrafted\cite{iclr2023layergrafted}\end{tabular}
&
\begin{tabular}[l]{@{}l@{}} 
SDMAE\cite{Mao2022SDMAE}\\
\end{tabular}

&
\begin{tabular}[l]{@{}l@{}}MST\cite{Li2021MSTMS}\\ ADIOS\cite{shi2022adversarial}\\ UnMAE\cite{Li2022UniformME}\\ SemMAE\cite{2022SemMAE}\\
LoMaR\cite{chen2022efficient}\\
i-MAE\cite{Zhang2022iMAE}\\ ccMIM\cite{iclr2023ccMIM}\\ AutoMAE\cite{Chen2023AutoMAE}\\ HPM\cite{cvpr2023HPM}\\ I-JEPA\cite{cvpr2023IJEPA}\\ MixMIM\cite{2022MixMIM}\\ ObjMAE\cite{Wu2022ObjectwiseMA}\\...
\end{tabular}
&
\begin{tabular}[l]{@{}l@{}}AttMask\cite{eccv2022attmask}\\
MILAN\cite{Hou2022MILAN}\\
DMJD\cite{Ma2022DisjointMW}\\
MaskAlign\cite{}\\
data2vec2.0\cite{2022Data2Vec2}
\end{tabular}
& 
\begin{tabular}[l]{@{}l@{}} ConvNeXt.V2\cite{Woo2023ConvNeXtV2}\\ SparK\cite{Tian2023SparK}\\
ConvMAE\cite{Gao2022ConvMAEMC} \end{tabular} & 
\begin{tabular}[l]{@{}l@{}} CAN\cite{Mishra2022ASE}\end{tabular}
& 
\begin{tabular}[l]{@{}l@{}} MSN\cite{Assran2022MaskedSN}\\
ExtreMA\cite{Wu2022ExtremeMF}\\
MimCo\cite{2022MimCo}\\ FLIP\cite{Li2022FLIP}\\ MOMA\cite{Yao2023MOMADF}\\ D-iGPT\cite{ren2023rejuvenating}
\end{tabular}
&
\begin{tabular}[l]{@{}l@{}} CMAE\cite{2022CMAE}\end{tabular}
&
\begin{tabular}[l]{@{}l@{}} ACLIP\cite{Yang2022AttentiveCLIP} \end{tabular}
%
%
\end{tabular}
    }
    \caption{Comprehensive categories of existing MIM methods according to the basic framework with four modules. We divided the \textit{Mask strategy} into Basic Mask and Advanced Mask, the \textit{Encoder Architecture} into CNN and Transformer, the learning \textit{Target} into Pixel, Tokenizer, and Feature, and the \textit{Head} into MIM Head, Contrastive Head, and their combination. We use the initials of each module to form a category name; for example, MAE is categorized as \textbf{BTPM} because it uses a Transformer as the encoder structure, a Random Mask as the masking strategy, a Pixel as the target, and MIM Head for reconstruction. Note that we only list the widely known methods for \textbf{BTPM}, \textbf{BTFM}, \textbf{BTTM}, and \textbf{ATPM} because they cover most of the existing MIM algorithms. Refer to Table~\ref{tab:full_mim_categoty} for detailed information and categories.
    }
    \label{tab:mim_classification}
    \vspace{-1.5em}
\end{table*}

\section{Method}
\label{section:method}
In this section, we will sequentially introduce the four essential modules for the MIM Framework, \textit{i.e.,} Mask Strategy, Targets, Architecture of the encoder, and MIM Head. Within each module, there are many studies; we will provide a more detailed classification and summary.
Then, we will discuss some research on MIM theory and several fundamental directions where MIM is applied.

\subsection{Masking Strategy}
This subsection will also spotlight typical masking strategies employed in MIM. For classification purposes, we bifurcate masking strategies into basic and advanced masking. \textit{Basic masking, which encompasses pixel-wise predictions based on AR models and the Random Mask introduced by MAE, has been elaborated upon in Sec.~\ref{sec:basic}}. Consequently, our ensuing discussion will primarily focus on Advanced Masking techniques. As illustrated in the accompanying figure, Advanced Masking can be further subdivided into four types: Hard Sampling, Mixture, Adversarial Mask, and Contextual Mask.

\textbf{Remark}: Despite improving performances, Mixture Mask and Adversarial Mask usually require more computational costs.
Therefore, an attention-based mask strategy might achieve a better trade-off between mining hard samples and computational overheads.

\subsubsection{Hard Sampling}
In the \textbf{AttMask}~\cite{eccv2022attmask} framework, a teacher model $f_{\theta'}$ is employed to extract the attention maps $\hat{a}$ and image features $f_{\theta}(\pmb{x})$ from the input images $\mathbf{X}$ and patches $\pmb{x}$. The student model $f_{\theta}$ then masks the regions with high attention scores in the attention maps. 
The reconstruct loss in AttMask is:
\begin{align}
    \hspace{-1.0em}
    \mathcal{L}_{\textrm{MIM}}=\sum_{v }\sum_{i \in \mathcal{N}} \mathbb{I}_{\{\mathcal{M}_i=0\}}f_{\theta}(\pmb{x}^v \odot \mathcal{M})_i \log f_{\theta'}(\pmb{x}^v \odot \mathcal{M})_i.
\end{align}
Employing attentive masking, \textbf{AttMask delivers excellent results and has relatively lower computational overhead}. Based on Table~\ref{tab:mim_classification}, AttMask is categorized as \textbf{Advanced Mask + Transformer + Features + MIM Head (ATFM).}

\textbf{HPM}~\cite{cvpr2023HPM} (ATFM) introduces a teacher-student framework. The teacher model $f_{\theta'}$ predicts the reconstruction loss for each patch $x_i$, while the student model $f_{\theta}$  masks and reconstructs the image $\pmb{x}$ using an "easy to hard" approach guided by the teacher model. 
The object of HPM concludes a reconstruction loss and a prediction loss, and reconstruction loss is formulated as \ref{eq:MAE}.

Meanwhile, \textbf{SemMAE}~\cite{2022SemMAE} (Advanced Mask + Transformer + Pixel + MIM Head, ATPM) implements a semantic-based masking strategy through semantic information learned by ViT, \textbf{MILAN}~\cite{Hou2022MILAN} (ATFM) combines attention mask with an online feature as the target. \textbf{ObjMAE}~\cite{Wu2022ObjectwiseMA} (ATFM) proposes an object-wise mask strategy that discards non-objective patches.

\subsubsection{Mixture}

\textbf{MixedAE}~\cite{Chen2023MixedAF} (ATPM): Based on MAE, MixedAE introduces a technique of blending portions from different images as input to the network. 
MixedAE enhances the model's representational capacity by incorporating CL. The loss function for this CL can be formulated as Eq.~\ref{eq:contrastive}.
\textbf{MixMIM}~\cite{2022MixMIM} (ATPM) utilizes both mixed masking and attention mask as masking methods and improves the network architecture to a  hierarchical Transformer. \textbf{i-MAE}~\cite{Zhang2022iMAE} (ATPM) designs a mixed masking strategy for its input and simultaneously introduces a linear layer to separate the mixed input before reconstruction to improve the performance.

\subsubsection{Adversarial}

\textbf{ADIOS}~\cite{shi2022adversarial} (ATPM) combines MIM with adversarial learning.
Generator $\mathcal{G}$ produces images with different masks based on the original image, while Discriminator $\mathcal{D}$ aligns the generated images with the original ones. 
Since ADIOS does not rely on the block construction of the Transformer, it can be implemented in the backbone of CNNs. \textbf{AutoMAE}~\cite{Chen2023AutoMAE} (ATPM), on the other hand, introduces a Mask Generator based on the MAE architecture to generate different mask strategies. 
The encoder adaptively reconstructs the original image based on different mask methods.

\subsubsection{Contextual Masking}
\textbf{UnMAE}~\cite{Li2022UniformME} (ATPM) proposes a Uniform Masking strategy for masking, with the selection of the masked portion consisting of two parts: Uniform Sampling and Secondary Masking. The former randomly samples a patch from a 2x2 grid, while the latter randomly masks a portion of the already sampled area. 
\textbf{LoMaR}~\cite{chen2022efficient} (ATPM), on the other hand, builds upon MAE by using small-window patches for local reconstruction prediction, improving efficiency and accuracy compared to MAE.

\begin{figure}[t!]
    \centering
    \includegraphics[width=1.0\linewidth]{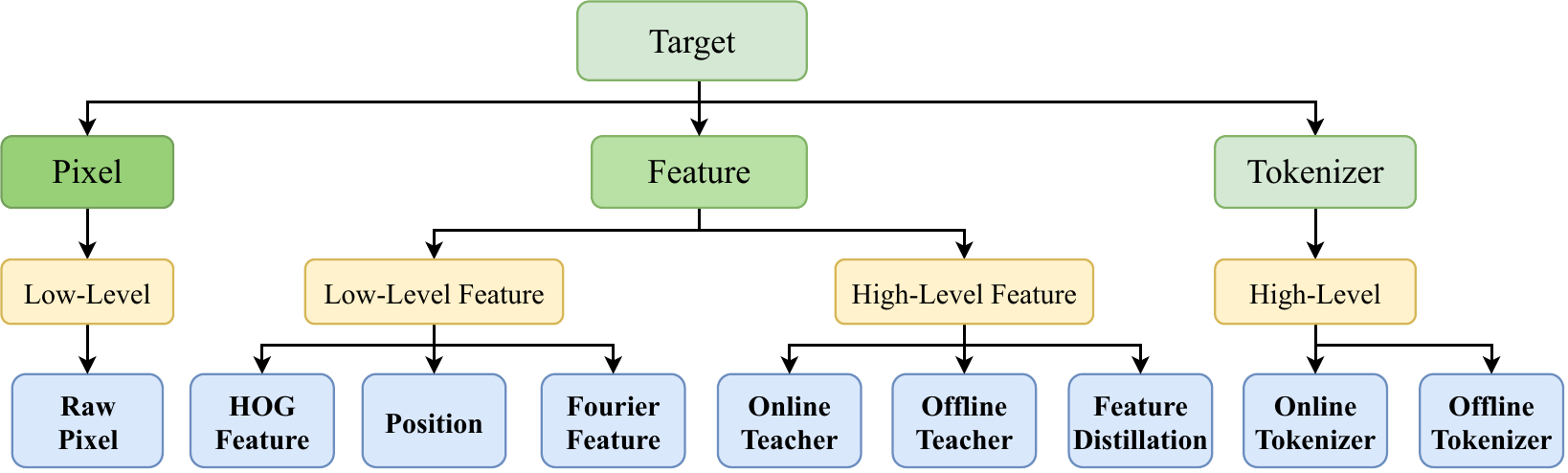}
    \caption{The types of the MIM target include three categories, that is Pixel, Feature and Tokenizer. 
    }
    \label{fig:target}
    \vspace{-1.0em}
\end{figure}

\subsection{Different Targets}
This subsection will delve into the learning targets in MIM training. We categorize these targets into three main types: tokenizer, pixel, and features. Delving deeper, these categories can be further detailed, with comprehensive explanations provided in Figure~\ref{fig:target}.

\subsubsection{Raw Pixel}
Raw Pixel is the most fundamental target in MIM. Classic models like MAE and \textbf{SimMIM}~\cite{Xie2021SimMIMAS} (BTPM) are based on Raw Pixel for image reconstruction. \textbf{I-JEPA}~\cite{cvpr2023IJEPA} (ATPM) uses a Context Patch as the input for the Encoder, and the reconstruction target is the three different patches adjacent to the Context Patch. By reconstructing through the Context Patch, I-JEPA can achieve better contextual representation capabilities while also reducing computational overhead.

\subsubsection{Tokenizer}


A tokenizer is a mapping function $\pmb{q}_{\phi}( \pmb{z}|\pmb{x})$ that encodes image $\mathbf{X} \in \mathbb{R}^{H \times W \times C}$ into $z = [z_{i}]_{i=1}^{|\mathcal{V}|} \in \mathcal{V}^{h \times w}$, where the vocabulary $\mathcal{V} = \{ i \}_{i=1}^{|\mathcal{V}|}$ contains token indices. These latent variables represent high-level semantic features of certain parts of the image. Hence, we can represent an image based on the dictionary $\mathcal{V}$, which can be used as the supervisory signal for  MIM. 
The tokenizer $\pmb{q}_{\phi}( \pmb{z}|\pmb{x})$ maps pixels $x$ into discrete tokens $z$ according to a visual codebook~\cite{Oord2017NeuralDR} (\textit{i.e.}, vocabulary), and decoder $\pmb{p}_{\psi }(\pmb{x}|\pmb{z})$ learns to reconstruct the image based on visual tokens $\pmb{z}$~\cite{Bao2021BEiT}. The learning objective of the tokenizer is:
\begin{align}
\min \mathbb{E}_{\pmb{z \sim }\pmb{q}_{\phi }(\pmb{z}|\pmb{x})}(\log \pmb{p}_{\psi }(\pmb{x}|\pmb{z})).
\end{align}
The training of tokenizers concludes \textbf{dVAE} variants~\cite{Ramesh2021ZeroShotTG}.

\textbf{BEiT}~\cite{Bao2021BEiT} (Basic Mask + Transformer +Tokenizer + MIM Head, BTTM) : 
In first stage, BEiT discretely encodes image $ \mathbf{X}\in\mathbb{R}^{H \times W \times C}$  into $z = [z_i]_{i=1}^{|\mathcal{V}|} \in \mathcal{V}^{h \times w}$, where the vocabulary $\mathcal{V} = \{ i \}_{i=1}^{|\mathcal{V}|}$ contains discrete token indices. 
After the tokenizer is pre-trained,  The encoder $f$ encodes the unmasked regions of an image, and encoded features are then passed through the MIM Head, with discrete image tokens serving as the supervision signal for learning.
The learning object of BEiT is:
\begin{align}
\max \sum_{\textrm{dataset}}  \mathbb{E}_{\mathcal{M}} \left[ \sum_{i \in \mathcal{N}} \mathbb{I}_{\{\mathcal{M}_i=0\}}\log p_{\text{MIM}}( z_i | \pmb{x}\odot \mathcal{M} ) \right],
\end{align}
where $\mathcal{D}$ denotes the traning corpus.

\textbf{iBOT}~\cite{iclr2022ibot} (BTTM) formulate MIM as a knowledge-distillation task and perform self-distillation using a teacher-student framework, which means iBOT uses an online Tokenizer. 
The teacher model is updated by the student model with EMA as Eq \ref{eq:EMA}. Building on the framework of BEiT, \textbf{BEiTv2}~\cite{2022BEiTV2} (BTTM) employs distillation on VQ to transform the discrete semantic space into compact codes. Building further upon BEiTv2, \textbf{BEiTv3}  (BTTM) integrates MOE and multimodality to design specialized tokenizers for vision, language, and vision-language tasks and scales up the model.
\textbf{Peco}~\cite{Dong2021PeCoPC} (BTTM) utilizes a perceptual prediction target to train a perceptual codebook. \textbf{mc-BEiT}~\cite{Li2022mcBEiTMD} (BCTM) represents a masked patch with a soft probability of vector instead of a unique token id. \textbf{CIM} \cite{fang2022corrupted} (BTTM) proposed an encoder-enhancer architecture in which a small pre-trained BEiT is used as an encoder and a CNN-based model can be applied to the enhancer. Pixel reconstruction and GAN loss are used in CIM, respectively.

\subsubsection{Low-Level Features}

\textbf{HOG Features}. \textbf{MaskFeat}~\cite{Wei2021MaskedFP}  (Basic Mask + Transformer + Feature + MIM Head, BTFM) proposes a framework based on MAE. Notably, the supervision signal for training the model is derived from the HOG features of the original image. 
\textbf{FastMIM}~\cite{Guo2022FastMIM} (BTFM) designs a Hierarchical Transformer and utilizes HOG features as the target.

\textbf{Position}. \textbf{DILEMMA}~\cite{Sameni2022RepresentationLB} (BTFM) employs a teacher model to generate position encoding. The student model is trained to predict new positions and judge whether the prediction is true or not.
\textbf{MP3}~\cite{Zhai2022PositionPA} (BTFM) trains a masked Transformer to predict the position of patches using MAE as a loss function. \textbf{SDMAE}~\cite{Xu2022MaskedAA} (ATFM) combines position prediction loss, pixel loss, and global contrastive loss to train its backbone. \textbf{DropPos} \cite{Wang2023DropPosPV} (BTFM) randomly selects a subset of patches and replaces their positional encodings with mask tokens. The positional encodings are then reconstructed.

\textbf{Fourier Features}. Models combined with Fourier Features can generally be divided into two main categories.
\textbf{Calculating Loss In Fourier domain}:
\textbf{Ge2AE}~\cite{Liu2022TheDI} (Basic Mask + Transformer + Feature + Both Head, BTFB) reconstructs in the Fourier domain while computing both contrastive loss and reconstruction loss. 
\textbf{A2MIM}~\cite{2022a2mim} (BCFM) utilizes
The intermediate layer features of the CNN-based and ViT-
based encoder to reconstruct ground truth in the spatiotemporal domain
and frequency domain.
The discrete Fourier transform of each channel is defined as:
\begin{equation}
\vspace{-0.5em}
\begin{aligned}
    \mathcal{F}_{(u,v)} = \sum_{H,W}
    x(h,w) e^{-2\pi j (\frac{uh}{H} + \frac{vw}{W})}.
\end{aligned}
\label{eq:DFT}
\end{equation}
The frequency domain learning objective is formulated as:
\begin{equation}
\begin{aligned}
\mathcal{L}_{freq} = &\sum_{C,H,W} \omega \big\lVert \mathcal{F}(x \odot \mathcal{M} +\\
    &\mathrm{de} (x) \odot (1 - \mathcal{M})) - \mathcal{F} (x)\big\lVert,
\end{aligned}
\label{eq:fft_loss}
\end{equation}
where $\omega=\omega(u,v)$ is  a dynamic frequency weighting matrix.
\textbf{Masking In Fourier Domain}: \textbf{MFM}~\cite{2022MFM}    (BTFM) masks in the frequency domain, adds noise, and then reconstructs the image. \textbf{MSCN}~\cite{Jing2022MaskedSC} (BTFM), after masking in the frequency domain, integrates with CL and employs a contrastive loss.
\textbf{PixMIM}~\cite{Liu2023PixMIMRP} (BTFM) reconstructs the image in both the spatial and frequency domain.

\subsubsection{High-Level Features}
\if\submission\submissionarXiv  
    This branch of research takes high-level features extracted from images as the MIM targets, which are often associated with the teacher model or distilled image features. This type of research can be categorized into offline teachers, online teachers, and those combined with knowledge distillation (KD).
\else
    This type of research can typically be categorized into off-line teachers, on-line teachers, and those combined with knowledge distillation (KD).
\fi

\textbf{Offline Teacher.}
\textbf{MILAN} \cite{Hou2022MILAN} (ATFM) utilizes CLIP~\cite{radford2021learning} to generate attention maps to guide the model to mask and generate features as the target.
\textbf{MOMA}~\cite{Yao2023MOMADF} (Basic Mask + Transformer + Feature + Contrastive Head, BTFC) builds upon the MAE  and uses pre-trained Multiple Teacher features as the prediction target. \textbf{Img2vec}~\cite{pan2023img2vec} (BTFM) uses a pre-trained ConvNet as the teacher model to extract features. Based on the MAE framework, it reconstructs patches and combines CL to compute the global loss. \textbf{TinyMIM}~\cite{Ren2023TinyMIM} (BTFM)
discovered that using the intermediate layer features of the teacher model often yields better results, with a smaller gap to downstream tasks. 

\textbf{Online Teacher.}
\textbf{data2vec}~\cite{Baevski2022data2vecAG} (BTFM) utilizes contextualized representations of the online teacher model and combines several modalities, including NLP, CV, and Speech. Data2vec updates its parameter with the EMA:
\begin{align}
  \pmb{\hat{\theta}} \leftarrow \tau \pmb{\hat{\theta}} +(1-\tau) \pmb{\theta}.
  \label{eq:EMA}
\end{align}
\textbf{data2vec.v2}~\cite{2022Data2Vec2} (ATFM), building on the foundation of data2vec, introduces a multi-mask training method to enhance efficiency and reduce computational costs. \textbf{dBOT}~\cite{liu2022dBOT} (BTFM), based on iBOT, has designed a multi-stage distillation scheme, concluding that teacher models with different parameters tend to have consistent performance in student models after multi-stage distillation.
\textbf{BootMAE}~\cite{Dong2022BootstrappedMA} (BTPM), while using online features as prediction targets, also adds the task of reconstructing image pixels. Unlike directly calculating the loss between features, \textbf{RC-MAE} \cite{Lee2022RCMAE} (BTPM) inputs the masked image into two Transformer encoders with EMA-updated parameters. It then computes the contrastive loss of the reconstructed image, supplemented by a task of pixel-level image reconstruction.
\textbf{MaskDistill}~\cite{2022maskdistill} (BTFM)
\textbf{MaskCLIP}~\cite{2022MaskCLIP} (Basic Mask + Transformer + Feature + Both Head, BTFB) integrates multiple techniques, including MIM, multi-modality, online features, and CL.

\textbf{Feature Distillation}. 
\textbf{DMJD}~\cite{Ma2022DisjointMW} (ATFM) proposes a disjoint mask and simultaneously trains the encoder using features distillation and prediction reconstruction methods. \textbf{CAE.v2} \cite{Zhang2022CAEVC} (BTFM) distills CLIP and is supplemented with a task to predict CLIP features.
\textbf{SdAE}~\cite{Chen2022SdAESM} (BTPM) delves into creating effective views for the teacher branch and proposes a multi-fold masking strategy to reduce computational costs.

\subsection{Different Network Architecture}

\begin{figure}[ht]
    \centering
    \vspace{-0.5em}
    \includegraphics[width=0.96\linewidth]{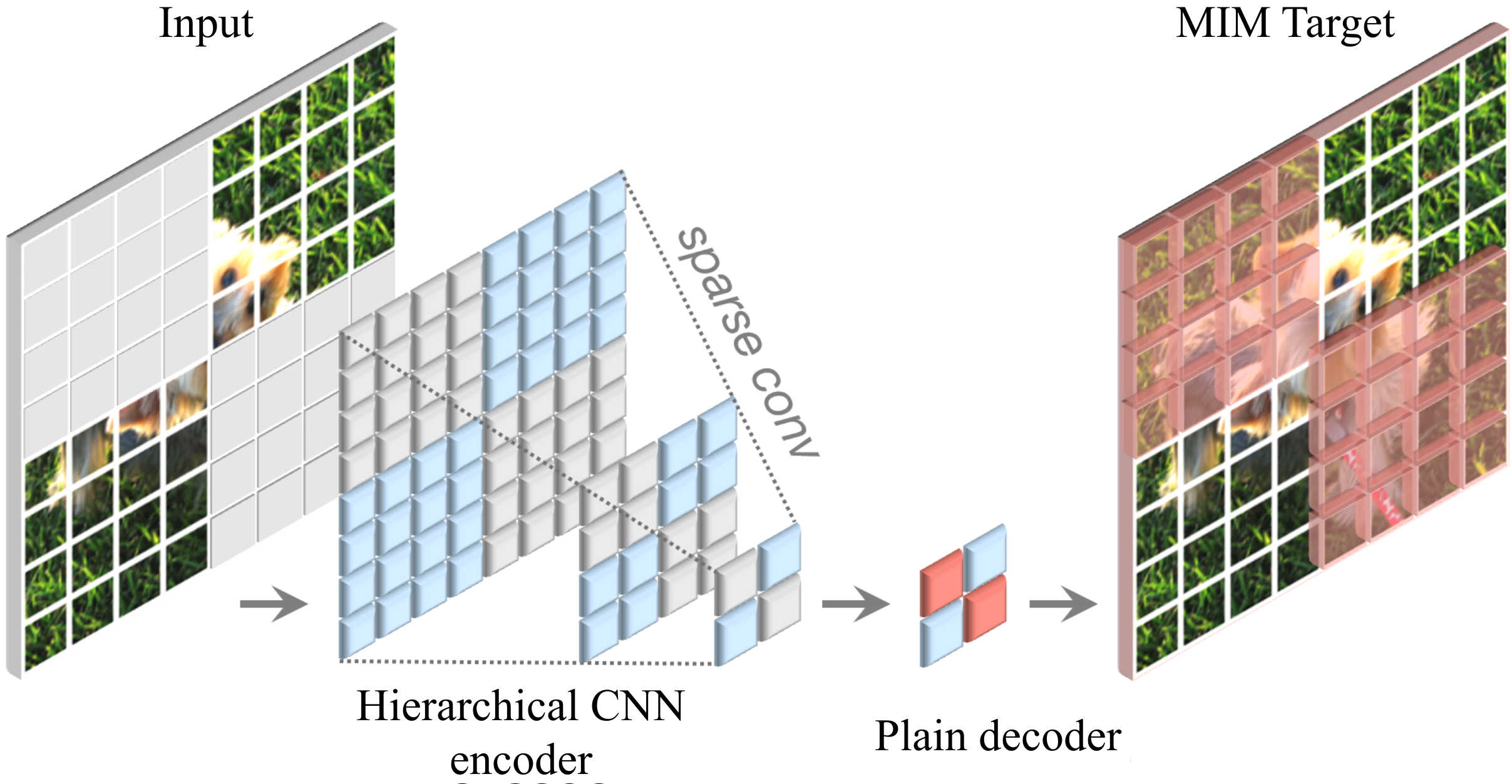}
    \caption{
    Illustration of MIM for CNN encoders with the sparse convolutions and masking~\cite{Woo2023ConvNeXtV2, Tian2023SparK}, where the encoder only aggregates information of visible tokens. The figure is reproduced from \cite{Woo2023ConvNeXtV2}.
    }
    \label{fig:sparse_conv}
    \vspace{-0.5em}
\end{figure}

\textbf{Transfer encoder to hierarchical vision transformer}:
\textbf{GreenMIM}~\cite{Huang2022GreenHV} (BTPM) inputs the masked image $\mathbf{X}\odot \mathcal{M}$ into a Hierarchical Transformer encoder. 
To reduce unnecessary computations in areas that are masked or do not contain useful information, the sparse convolution is introduced to discard invisible patches and only processes on the visible patches, achieving patch merging, similar to Figure~\ref{fig:sparse_conv}. 
\textbf{HiViT}~\cite{Zhang2022HiViTHV} (BTPM) removes local inter-unit operations, resulting in structurally simple hierarchical vision Transformers.
\textbf{Hiera}~\cite{icml2023Hiera} (BTPM) eliminates the need for many of the complex components found in other hierarchical vision Transformers and achieves superior accuracy. \textbf{ConvMAE}~\cite{Gao2022ConvMAEMC} (Basic Mask + CNN + Pixel + MIM Head, BCPM) proposes a multi-scale hybrid convolution-Transformer, employs a masked convolution to prevent information leakage in the convolution blocks and a block-wise mask to reduce the computational cost. \textbf{SparseMAE}~\cite{iccv2023sparsemae} (BCPM) introduces sparse MHSA and FFN blocks for sparse pre-training.

\textbf{Make MIM Compatible with CNN}:
\textbf{CIM}~\cite{fang2022corrupted} (Basic Mask + CNN + Tokenizer + MIM Head, BCTM) employs an auxiliary generator equipped with a compact trainable BEiT to corrupt the input images, thereby enhancing the network's capability to predict whether each visual token has been replaced by a sample from the generator. Due to CIM's approach of using an auxiliary generator to corrupt the input, there's no need for specific input formats, which are compatible with CNNs. 
\textbf{A2MIM} \cite{2022a2mim} (BCFM) 
posits that masking at the block embedding layer aligns well with the attention mechanism of Transformers, offering robustness against occlusion. For CNNs, masking at the network's input stages leads to low-order interactions, undermining CNN's context extraction capability. 
\if\submission\submissionarXiv  
    Therefore, A2MIM suggests masking intermediate features encompassing semantic and spatial information, allowing the mask token to encode interactions with a moderate number of tokens.
\fi

\textbf{Specially designed CNN architectures}:
\textbf{Spark} \cite{Tian2023SparK} (BCPM) pinpointed the incompatibility of convolutional networks' hierarchical nature with the challenges of recognizing irregularly masked images and BERT's single-scale pre-training, impeding MIM implementation on CNNs. To resolve this, Spark treated unmasked pixels as 3D point clouds, employing sparse convolution for encoding, suitable for irregular masking. Additionally, they introduced a hierarchical decoder, aligning with CNN's structure, to reconstruct images from multi-scale features.
As shown in Figure \ref{fig:sparse_conv}, \textbf{ConvNext.v2} \cite{Woo2023ConvNeXtV2} (BCPM) 
features a convolutional masked encoder based on ConvNext, converting standard convolution to sparse convolution. Its decoder uses a streamlined ConvNext block for the simultaneous processing of encoded and masked tokens, integrating MIM into CNN architecture. 

\subsection{Head}
This subsection will discuss the Head of MIM research. We distinguish the heads into three categories: Contrastive Head, MIM Head, and Both Contrastive Head and MIM Head. We will bifurcate our discussion into two primary segments, focusing separately on the MIM and the Contrastive Head.
\if\submission\submissionarXiv  
    It's essential to highlight that both the MIM Head and Contrastive Head can have diverse internal architectures. The specifics of these structures are visually represented in the provided Figure~\ref{fig:contrastive_head}. Our discussion is bifurcated into two primary segments, focusing separately on the MIM Head and the Contrastive Head.
\fi

\begin{figure}[ht]
    \centering
    \vspace{-0.5em}
    \includegraphics[width=1.0\linewidth]{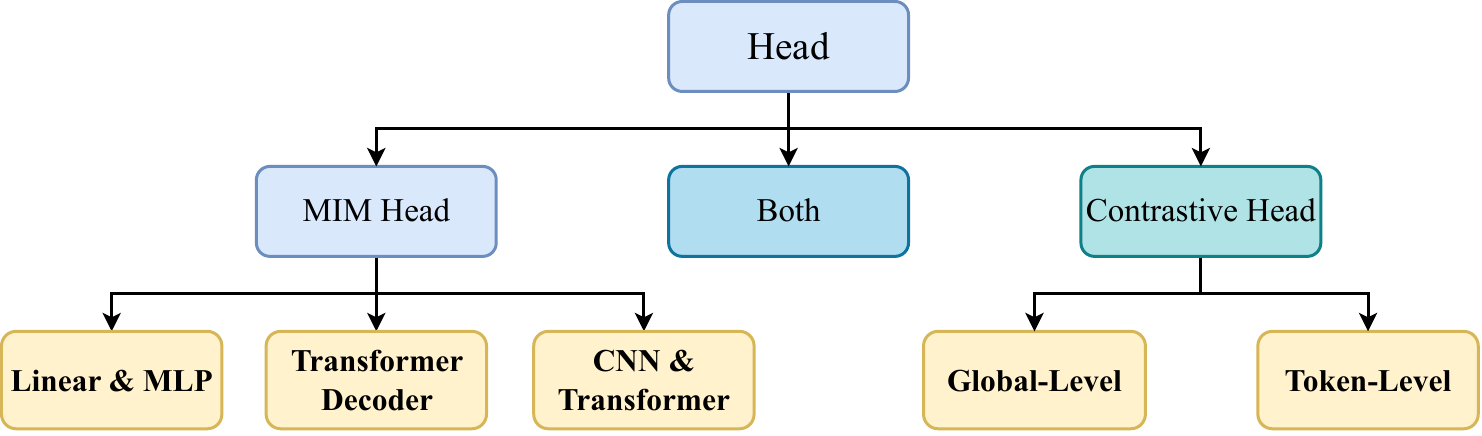}
    \caption{The types of MIM Head include Linear or MLP, Transformer, or a combination of CNN and Transformer. The Contrastive Head section is categorized based on the algorithm type into Token-level and Global-level.}
    \label{fig:Head}
    \vspace{-0.75em}
\end{figure}

\subsubsection{MIM Decoder}
\label{sec:mim_decoder}
\textbf{Linear or MLP}: \textbf{SimMIM} \cite{Xie2021SimMIMAS} (BTPM) essentially adopts the framework of MAE but with several significant modifications. In SimMIM, the encoder processes both the visible patches and the masked tokens simultaneously. Remarkably, SimMIM's decoder achieves satisfactory results using just a \textbf{Linear Prediction Head}. A detailed comparison between SimMIM and MAE can be found in the provided table.
Other MIM models utilize linear layers as the MIM decoder, \textit{e.g.,} \textbf{BEiT}, \textbf{BEiT.v2}, and \textbf{data2vec}, \textit{etc}.

\textbf{Transformer Decoder and Combined Decoder}: The Transformer decoder is most widely used in MIM,  while the combined decoder of Transformer and CNN further improves the MIM performances as shown in Figure \ref{fig:Head}. \textbf{LocalMAE} \cite{Wang2023MaskedIM} (BTFM) employs intermediate features from multiple stages for multi-scale reconstruction. In the reconstruction segment, LocalMAE introduces a Transformer-Deconvolution-MLP architecture for the task.

\begin{figure*}[t]
\vspace{-0.5em}
\centering
\begin{minipage}{0.32\linewidth}
    \centering
    \vspace{-0.0em}
\resizebox{1.0\linewidth}{!}{
    \setlength{\tabcolsep}{0.7mm}
    \begin{tabular}{lcc}
    \toprule
        Model   & MAE         & SimMIM      \\ \hline
        Mask    & Random      & Random      \\
        Encoder & Transformer & Transformer \\
        Target  & Raw Pixel   & Raw Pixel   \\
        Input & Visible  & Visible  and Masked \\ 
        Head    & Transformer     & Linear       \\
        Method & Auto-Encoder & Auto-Encoder \\
    \bottomrule
    \end{tabular}
    }
    \caption{The most significant difference between SimMIM and MAE lies in whether the input to the encoder includes the masked tokens and the structure of the MIM Head. An in-depth explanation of this aspect can be found in the designated Sec.~\ref{sec:mim_decoder}.}
    \label{simMIM_and_MAE}
\end{minipage}
~\begin{minipage}{0.67\linewidth}
    \centering
    \includegraphics[width=1.0\linewidth]{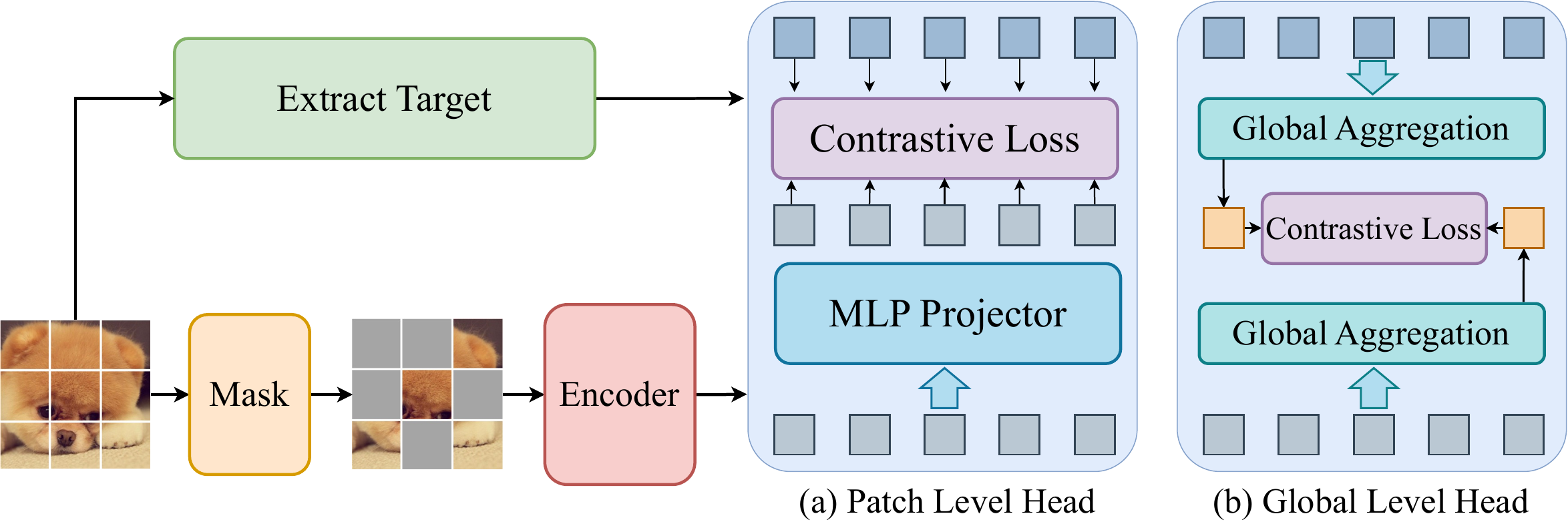}
    \vspace{-1.75em}
    \caption{Two categories of MIM methods combined with  CL: token-level and global-level CL heads. For the token-level head, tokens are subjected to an MLP Projector and compute the CL loss. The global-level head aggregates global information on MIM targets and tokens before calculating the CL loss.}
    \label{fig:contrastive_head}
\end{minipage}
\vspace{-1.0em}
\end{figure*}

\textbf{Remark}: 
The effectiveness of image reconstruction in certain models using a simple Linear Head, as opposed to others requiring a complex Transformer decoder, hinges on the inclusion of masked tokens in the Encoder's input. When masked tokens are part of the input, they interact with visible patches within the Encoder, facilitating early image information capture and enabling effective reconstruction with just a \textbf{Linear Head}. In contrast, without masked tokens in the Encoder, these tokens must interact within a sophisticated \textbf{Transformer decoder} to reconstruct the image.
Figure~\ref{simMIM_and_MAE} compares SimMIM and MAE in detail.

\subsubsection{Combined with Contrastive Head}
There are typically two approaches combining CL and masked language modeling:
The first incorporates masked images as a data augmentation technique and applies them within the CL framework to benefit CL.
The second utilizes the standard masked language modeling framework and adds CL objectives in the prediction head to benefit masked language modeling.
In this section, we will detail both lines of work and elaborate on the network architecture for the contrastive prediction head.

\textbf{Mask as Data Augmentation}: \textbf{MSN}~\cite{Assran2022MaskedSN}
(BTFC) utilizes masked images as an augmentation technique and incorporates them into the framework of PCL~\cite{Li2020PrototypicalCL}.
\textbf{MSCN}~\cite{Jing2022MaskedSC} (BTFM) and \textbf{Mimco}~\cite{2022MimCo} (BTFC) incorporate masked images as data augmentation into the frameworks of SimCLR and BYOL respectively, to benefit CL methods. This achieves an integration of masked modeling and CL.

\textbf{Add CL Loss}: This line of work builds upon masked modeling and incorporates a contrastive prediction head by adding or replacing the original MIM head. It can be categorized into two groups: token-level CL and global-level CL. Details are illustrated in Figure \ref{fig:contrastive_head}.
\textbf{Token Level Contrastive}:
\textbf{ConMIM}~\cite{2022ConMAE} (Basic Mask + Transformer + Pixel + Contrastive Head, BTPC) utilizes two Transformer encoders, one for masked images and another for unmasked images. The branch that takes the masked images as input predicts the original images. The features obtained from the prediction are contrasted with those from the unmasked images through CL.
The CL loss is defined as:
\begin{align}
    \vspace{-0.5em}
    \mathcal{L}_{\textrm{ con}}(x) = - \log \frac{\exp{(\langle f(\pmb{x}_{i}),\pmb{x}_{j}\rangle /\tau)}}{\sum_{k=1}^{2N} \mathbb{I}_{\{k\neq i\}}\exp{(\langle f(\pmb{x}_{i}),\pmb{x}_{k}\rangle /\tau)}},
    \label{eq:contrastive}
\end{align}
\textbf{Global Level Contrastive}:
\textbf{ccMIM}~\cite{iclr2023ccMIM} (ATPM) employs attention  to rank each patch in the image $x$ and selects the more challenging parts as masked set $\mathcal{M}$ for reconstruction. 
Subsequently, global-level CL is performed on the CLS token.
\textbf{CAN}~\cite{Mishra2022ASE} (Basic Mask +Transformer +Pixel +Both, BTPB) adds Gaussian noise to the masked images. Building upon MAE, it performs pooling before reconstructing the image and computes a global-level CL loss.

\textbf{Architecture of Contrastive Head}: The CL Head usually utilizes the classical CL projection heads, consisting of multiple \textbf{MLP} or \textbf{FNNs}. They typically have an appended \textbf{BN} layer, as seen in models like \textbf{SimCLR}~\cite{chen2020simple} and \textbf{BYOL}~\cite{grill2020bootstrap}. A characteristic feature of these heads is that they often upscale the dimensions, having a larger number of channels. 
For research that employs the \textbf{Transformer Decoder} as the Contrastive Head, considerations usually revolve around the depth and width of the Transformer blocks.

\vspace{-1.0em}
\subsection{Theoretical Foundation}
Supervised learning, offers strong mathematical theoretical guarantees, outlining specific conditions for assured learning success. It generally assumes training and test datasets to be independently and identically distributed. As training iterations increase, one can often achieve lower training and test losses. This is because supervised learning is relatively straightforward. In contrast, unsupervised learning lacks the simple and intuitive theoretical guarantees present in supervised learning. Intuitively, we believe that the essence of unsupervised learning is a form of information \textbf{compression}. The compression algorithms learned from the training set represent the universal knowledge and structure inherent within the data. The way to evaluate these compression algorithms is to determine whether they extract all the knowledge from unlabeled data, i.e., whether they provide as much assistance as possible and yield the maximum benefit. We will elucidate and summarize the theoretical foundations of MIM from three perspectives.

\textbf{From CL}:
\textbf{Layer Grafted}~\cite{iclr2023layergrafted} (BTPC) finds that MIM and CL are suitable for lower and higher layers, respectively. The model designs a gradient surgery experiment by computing the cosine similarity between gradients of two tasks following \cite{Yu2020GradientSF} and verifying that the MIM loss and CL loss have different targets to optimize. The cosine similarity is:
\begin{align}
    \pmb{C}_{\textrm{MIM},\textrm{CL}}(x) =\frac{\nabla_\theta L_\textrm{MIM}\left(x\right)^T}{\left\|\nabla_\theta L_\textrm{MIM}\left(x\right)\right\|} \frac{\nabla_\theta L_\textrm{CL}\left(x\right)}{\left\|\nabla_\theta L_\textrm{CL}\left(x\right)\right\|}.
\end{align}
They propose a "sequential cascade" approach where early layers are first trained under one MIM loss, and then later layers continue to be trained under another CL loss.
\if\submission\submissionTPAMI  
    and then later layers continue to be trained under another CL loss.
\else
    and then later layers continue to be trained under another CL loss:
    \begin{equation}
        \mathcal{L}_{\textrm{MIM}} \rightarrow \mathcal{L}_{\textrm{CL}}.
    \end{equation}
\fi
\cite{Zhang2022HowMM} demonstrates that the mask loss exhibits a lower bound compared to the align loss in CL, making it more effective than aligning within CL. 
\begin{align}
    \vspace{-0.5em}
    \mathcal{L}_{\textrm{MAE}} \geq \frac{1}{2}\mathcal{L}_{\textrm{align}} - \epsilon +\textrm{const}.
\end{align}
Subsequently, a uniform loss, akin to that in CL, is incorporated into the mask loss.

\textbf{From Masking}: 
\cite{cvpr2023understandingMAE} models MIM as a hierarchical latent variable model. The objective of MIM is to recover the latent variable $z$ shared between visible patches and invisible patches based on the lower-level visible patches. This latent variable encapsulates the information shared between the visible patch and the invisible portions. Both a very low mask ratio and an extremely high mask ratio tend to make the model focus on recovering low-level latent variable information, making it challenging to learn higher-level semantic features. Therefore, the mask ratio in MAE
can assist the model in capturing higher-level latent variable information, enhancing its representation capability.

\textbf{From Empirical Study}:
Many studies have extensively explored certain characteristics of masked language modeling through numerous experiments and obtained some valuable conclusions. \cite{Xie2022RevealingTD} and \cite{2022AnES} verified through extensive experiments that, compared to other self-supervised methods like jigsaw puzzles and image inpainting, masked language models demonstrate better transferability and superior performance on tasks like pose estimation, depth prediction, video object tracking, and object detection. \cite{Xie2022OnDS} showed that masked models tend to underperform and are prone to overfitting on small datasets. As the dataset grows larger, the performance improvement of masked language models accelerates. \cite{2022UnderstandMIM} suggested that the efficacy of masked language modeling stems largely from the masking operation itself as the key to good performance, while different masking strategies contribute limited improvements.

We summarize some conclusions:
\begin{itemize}[leftmargin=1.0em]
\vspace{-0.25em}
    \item \textbf{From CL}:
    MIM, focusing on low-level features with local bias, contrasts with CL's high-level feature focus, elucidating the latter's earlier development. Previously, CNNs, with their inherent local bias, complemented CL, mutually enhancing effectiveness. However, the similar local biases of MIM and CNNs resulted in less optimal MIM performance on CNN architectures. The emergence of ViT, favoring global information capture, aligns better with MIM, elevating its prominence in SSL algorithms.
    \item \textbf{From Masking}: Masking, essential in MIM, uses higher ratios in the visual domain compared to NLP due to images' greater redundancy. Smaller mask ratios barely affect image semantics, so larger ratios obscure key information, making reconstruction harder and fostering robust model representations.
    \item \textbf{From Empirical Study}: Models based on MIM exhibit certain characteristics and preferences. For instance, they rely more on large-scale data for training and tend to learn better representations with larger datasets. Masked modeling performs better on tasks that require more detailed visual information, such as video object tracking and pose estimation. These tasks demand the model's ability to capture low-level information.
\end{itemize}

\subsection{Auto-Regressive For Generation}

\begin{figure*}[t!]
    \centering
    \vspace{-1.0em}
    \includegraphics[width=0.99\linewidth]{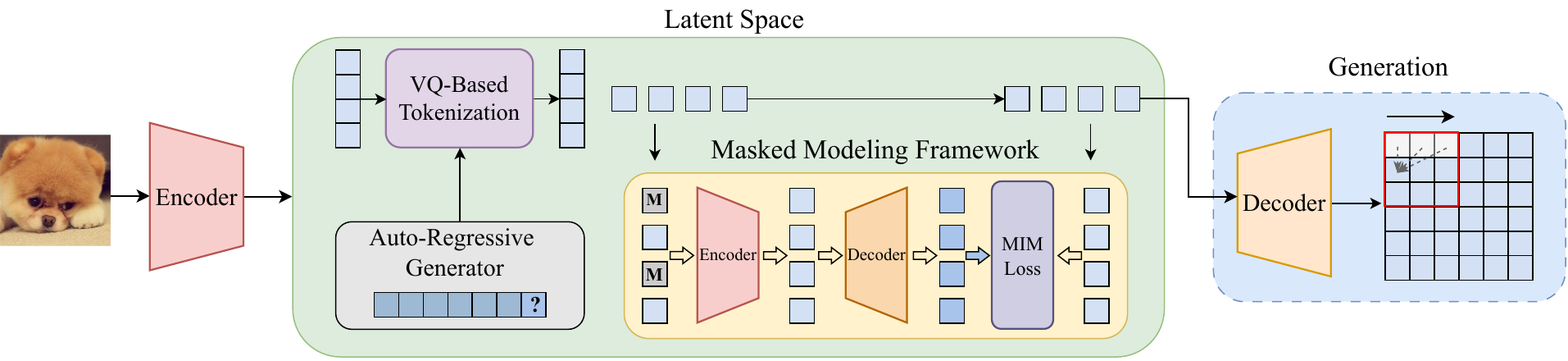}
    \caption{Research on autoregression (AR) for generation and pre-training can be summarized by this flowchart. Some studies focus on improving the quality and speed of image generation, while others combine pre-training with image generation, performing further operations in the latent space. The figure is reproduced based on \cite{esser2021taming, cvpr2023MAGVIT, cvpr2023mage}.
    }
    \vspace{-1.0em}
    \label{fig:AR_framework}
    \vspace{-0.5em}
\end{figure*}

Most MIM research utilizes AE for generative SSL, but AR modeling remains crucial in generative SSL. Significant research merges AR generation with MIM for representation learning and generative tasks. This section covers AR generative models and explores the integration of SSL by AR. Figure \ref{fig:AR_framework} shows the differences between these paradigms.

\subsubsection{VQ-Based Generation}
\label{subsec:ar}
Vector Quantization (VQ) is a significant technique in generative models, where it quantizes the continuous features learned by the encoder into discrete vectors in a codebook. 

\textbf{VQ-VAE}~\cite{oord2018representation}  introduces a generative framework that encompasses both generation and training processes. During training, VQ-VAE encodes image pixels into feature vectors, searching for the token in the codebook that is closest to the feature vector. The image is then reconstructed through the decoder. Therefore, the training loss includes the quantization loss of the vectors and the reconstruction loss:
\begin{align}
\begin{split}
\mathcal{L}_{\textrm{VQ-VAE}} = &\|x - g(v_q)\|^2 + \|sg[f(x)]- v_q\|^2 \\+ 
 & \beta \|f(x) - sg[v_q]\|^2,
\end{split}
\end{align}
where $\beta$ is a hyperparameter used to control the weights of the two losses. The generation process involves producing feature vectors through \textbf{PixelCNN}~\cite{van2016conditional}, followed by vector quantization of these feature vectors, and then generating new images via the decoder. Subsequent research based on VQ-VAE has two main focuses: one is to improve the training process to enhance the quality of image generation, and the other is to improve the generation process to increase the speed of image generation.

\textbf{Improve Generation Quality}: 
\textbf{VQ-GAN} is based on the VQ-VAE architecture, using GPT-2 as the generator in the workflow to produce discrete encodings. To enhance the reconstruction performance of the Decoder, an adversarial loss is added to the reconstruction loss.
The learning object consists of the reconstruct loss and adversarial loss:
\begin{align}
    \mathcal{Q}^* = \min_{f,g, \mathcal{V}} \max_{\mathcal{D}}
    \mathbb{E}_{x\sim p(x)} \Big[
    &\mathcal{L}_{\textrm{VQ}}(f, g, \mathcal{V}) \nonumber \\
    + &\lambda \mathcal{L}_{\textrm{GAN}}(\{f, g, \mathcal{V} \}, \mathcal{D})
    \Big].
\end{align}
Based on GPT-2, the process of generation is:
\begin{align}
      \max\limits_{\theta} p_{\theta}(\pmb{v}) = \sum_{t=1}^{T} \log p_{\theta}(\pmb{v}_t | \pmb{v}_{1:t-1}).
\end{align}

\textbf{Improve Generation Speed}:
Based on the VQ-VAE and VQ-GAN, \textbf{MaskGIT}~\cite{Chang2022MaskGITMG} learns to predict randomly masked tokens by attending to tokens from all directions. In the inference stage, the model initially generates all tokens of the image simultaneously and subsequently refines the image iteratively based on prior generations.
\textbf{RandSAC}~\cite{Hua2022SelfsupervisionTR} adopts a strategy of segmenting tokens into hierarchical sections. Within each section, it employs a parallel
prediction mechanism akin to BERT, while between different sections, it utilizes a sequential prediction approach reminiscent of GPT. Randomizing the sequencing of sections and leveraging parallel training, significantly enhances efficiency.

\subsubsection{Combining Pre-training with Image Generation}
\textbf{iGPT}:  
By predicting pixel values through the Transformer's autoregressive approach, iGPT achieves image generation capabilities. The unsupervised learning on large-scale unlabeled data makes iGPT a pre-trained model, which can achieve good results on downstream tasks through fine-tuning. \textbf{MAGE}~\cite{cvpr2023mage} first maps images to tokens in a discrete latent space using VQ-GAN, then performs masked image modeling by masking tokens in the latent space.  In this way, MAGE can learn representations via masked image modeling in the latent space while achieving image generation. 
\textbf{RCG}~\cite{Li2023SelfconditionedIG} trains a representation generator by adding noise to the encoded representation and then removing it. Subsequently, it utilizes the generated representation within the MAGE architecture to achieve pixel generation, which unifies pre-training and representation learning.

\subsection{Vision Fundation Model}
As DL research increasingly focuses on integrating multi-modal data, it has made multimodal research a key area in AI. We divide multimodal studies into three categories: The first focuses on using multimodal data for pre-training to enhance visual network architectures and maximize model potential, as detailed in Table \ref{tab:scaling_up}. The second revolves around generating multimodal data, including text-to-image conversion, summarized in Table \ref{tab:multimodality}. The third involves developing a vision generalist model that consolidates various visual tasks within a singular network architecture.

\subsubsection{Pre-train With Multimodality}

\textbf{Masked Modeling Methods}.
\textbf{VL-BERT} \cite{Su2019VLBERTPO} incorporates visual and linguistic inputs into a BERT-based architecture, allowing early and unrestricted interactions between modalities for joint representation learning.
\textbf{MaskVLM}~\cite{Kwon2022MaskedVA} applies to mask to image-text pairs, and then the masked images and masked texts are separately inputted into the image encoder and text encoder. Furthermore, a multimodal encoder is designed to encode the masked text and image, followed by simultaneous reconstruction of both the image and text. \textbf{BEiT.v3} integrates MOE and multimodality to design specialized tokenizers for vision, language, and vision-language tasks and scales up the model.

\begin{figure}[t]
    \centering
    \includegraphics[width=1.0\linewidth]{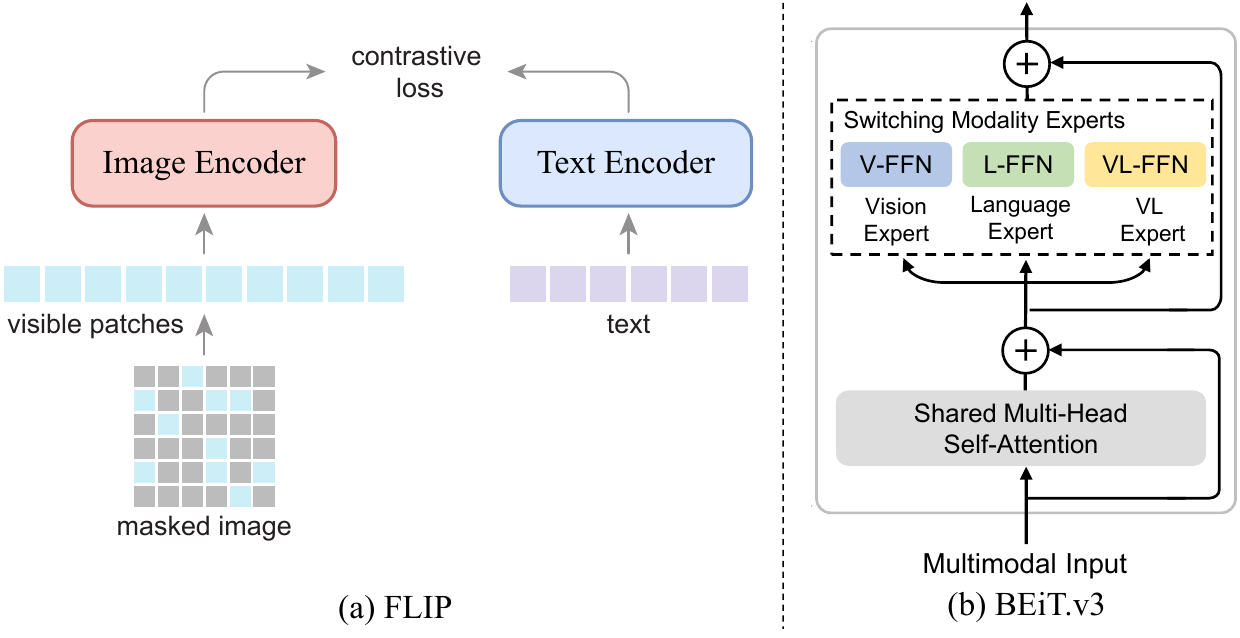}
    \vspace{-1.25em}
    \caption{
    Illustration of masked modeling with multimodality. (a) FLIP~\cite{Li2022FLIP} applies masking augmentations to the CLIP~\cite{radford2021learning} framework for text-image alignment. (b) BEiT.v3~\cite{2022BEiTV3} designs a mixture-of-export encoder for text-image. The figures are reproduced from \cite{2022BEiTV3} and \cite{Li2022FLIP}.
    }
    \label{fig:beit_flip}
    \vspace{-1.0em}
\end{figure}

\textbf{Contrastive Methods}. \textbf{A-CLIP}~\cite{Yang2022AttentiveCLIP} comprises an online update vision encoder and a language encoder. After images go through extracted feature maps and are masked, they undergo V-L CL and compute loss with CLIP features. In Figure~\ref{fig:beit_flip}, \textbf{FLIP}~\cite{Li2022FLIP} uses visible image patches and text, which compute a contrastive loss after passing through different encoders. \textbf{MaskCLIP} \cite{2022MaskCLIP} incorporates textual encoding into the masked image modeling architecture and computes contrastive loss between language and images to improve model performance through CL.

\textbf{Scaling up}.
DL models often see substantial performance improvements when the number of model parameters reaches a certain scale. Models based on MAE also exhibit phenomenal changes when their parameter size is expanded to a certain extent. A series of studies have scaled up the MAE parameters and tested their performance in various downstream tasks. Models such as \textbf{EVA} \cite{Fang2022EVAET}, \textbf{EVA-02} \cite{Fang2023EVA02AV}, \textbf{WSP} \cite{Singh2023WSP}, and others have achieved excellent results with large parameters. Table \ref{tab:scaling_up} summarizes information and performances of this category of models.

\begin{figure*}[t!]
    \centering
    \vspace{-1.0em}
    \includegraphics[width=1.0\linewidth]{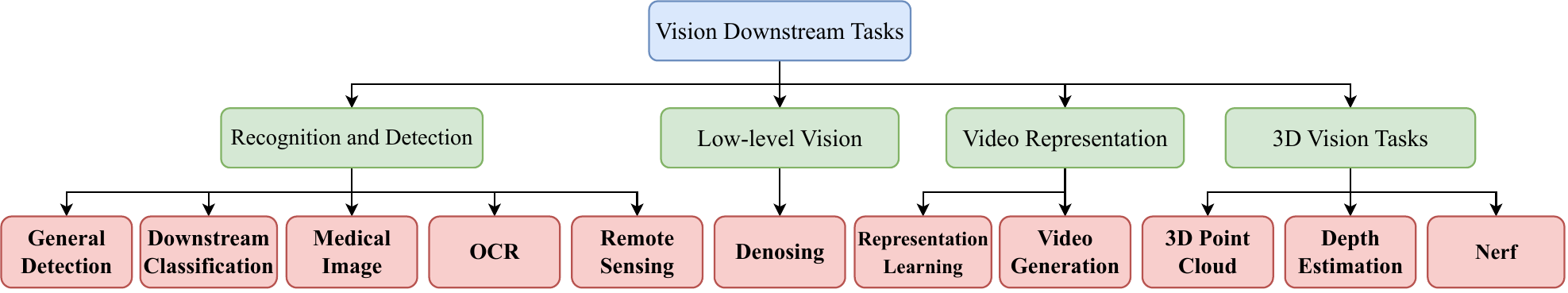}
    \caption{Illustration of various downstream tasks in computer vision. We summarize them by the label (task) types and data modalities. For example, tasks under \textbf{recognition and detection} utilize sample-level (\textit{e.g.,} classification) or sparse objective-level labels (\textit{e.g.,} detection and OCR) on 2D images, while \textbf{low-level vision} tasks prefer pixel-level supervision.
    }
    \vspace{-1.0em}
    \label{fig:cv_downstream}
    \vspace{-0.5em}
\end{figure*}

\subsubsection{Multimodality for Image Generation}
Another significant research direction in CV for multimodal models involves using multimodality for image generation. This encompasses various tasks, including Text-to-Image Generation and Image Generation. The study of image generation primarily falls into two approaches. The first employs an autoregressive method, predominantly based on VQ, and falls under \textbf{VQ-based} algorithms such as \textbf{DALLE}]~\cite{Ramesh2021ZeroShotTG}. We have delved further into this in Sec.~\ref{subsec:ar}. The other category primarily utilizes diffusion with multimodality for image generation. Common models in this category include , \textbf{DALLE-2}~\cite{Ramesh2022HierarchicalTI}, \textbf{DALLE-3}~\cite{BetkerImprovingIG}, \textbf{Stable Diffusion}~\cite{Rombach2021HighResolutionIS}, \textbf{GPT-4V}~\cite{wen2023road}, among others.

\subsubsection{Vision Generalist Model}
Vision Generalist Model unifies multiple tasks within a single model, selecting different tasks through prompt input and setting the model's output to a specific target, thereby achieving the unification of various tasks.
\textbf{Painter} \cite{Wang2022ImagesSI} considers an image paired with its corresponding task output, such as text or features, as a sample pair. Such a pair can encompass multiple modalities. The corresponding task output of the image is masked, and then the image, serving as the task's prompt, is fed into the encoder to reconstruct the corresponding task output. 
\textbf{InstructDiffusion} \cite{Geng2023InstructDiffusionAG} and \textbf{InstructCV} \cite{Gan2023InstructCVIT} build upon the foundation of stable diffusion, using prompts and the original image to reconstruct different task objectives, achieving a unification of various task architectures.
\textbf{LVM} \cite{bai2023sequential} uses a VQ-GAN encoder to convert images into tokens for training with an autoregressive Transformer. It generates outputs by forming partial visual sentences for specific tasks. Additionally, the authors introduce a large-scale LAION-5B dataset for in-context learning with visual sentences as a unified data unit. 

\section{Vision Downstream Task}
\label{sec:cv_downstream}
In this section, we will introduce the specific applications of MIM in vision downstream tasks. Broadly speaking, we categorize the applications of MIM in vision downstream tasks into four parts: recognition and detection, low-level vision, video representation, and 3D vision tasks. Figure \ref{fig:cv_downstream} provides a classification of CV downstream tasks.

\subsection{Video Representation}
Research on MIM pre-training for videos can be divided into two parts: one part is based on the Masked AE framework (\textit{e.g.,} adapting to the MAE framework to video, and the other is based on the AR framework.

\subsubsection{AE-Based Representation Learning}
AE-based models usually aim for video reconstruction as the task objective to achieve the purpose of representation learning. However, videos have higher dimensionality compared to images. Therefore, the focus is on adapting video data to fit within architectures like MAE and BEiT.
To apply the 2D MAE framework to videos, a common approach is to mask out space-time tubes instead of spatial patches. This treats the video as a sequence of 2D frames and masks contiguous patches across time. More advanced methods mask at the 3D voxel level for finer spatio-temporal masking. Additional modifications, like introducing a motion-specific encoder, can help capture temporal dynamics.

Based on the framework of \textbf{MAE}, \textbf{VideoMAE}~\cite{Tong2022VideoMAEMA} performs spatial-temporal masking during pre-training by randomly occluding cubic patches in spatiotemporal spaces.
\if\submission\submissionarXiv  
\begin{figure}[t!]
    \centering
    \includegraphics[width=1.0\linewidth]{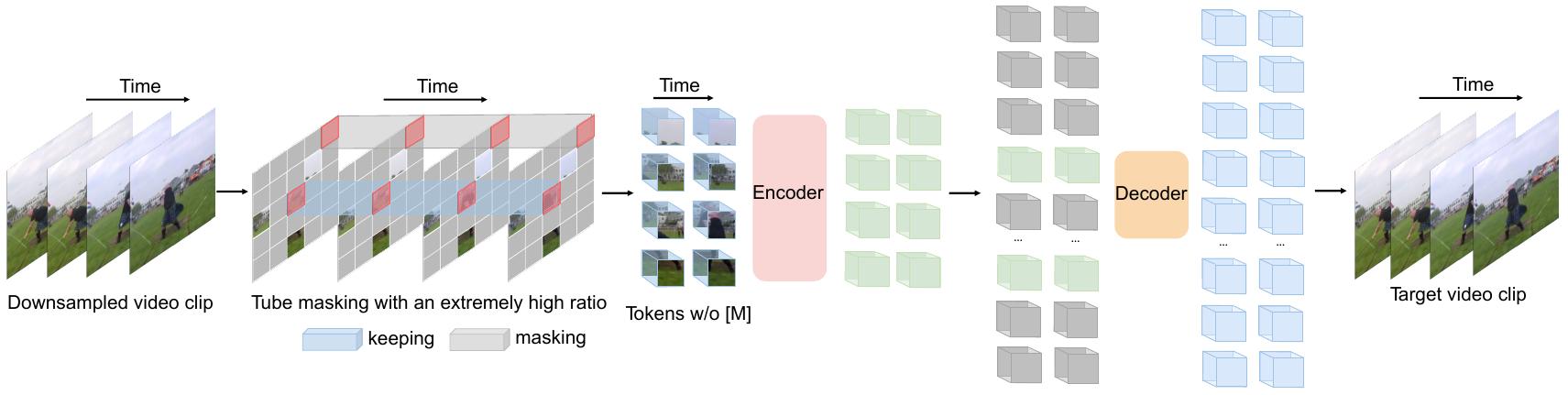}
    \caption{Illustration of MIM on videos. Taking VideoMAE~\cite{Tong2022VideoMAEMA} as an example, it employs an asymmetric encoder-decoder architecture with random spatiotemporal cubic masks and reconstructs the missing ones. The figure is reproduced from \cite{Tong2022VideoMAEMA}.
    }
    \label{fig:videomae}
    \vspace{-1.0em}
\end{figure}
Figure \ref{fig:videomae} shows the framework of VideoMAE.
\fi
\textbf{AdaMAE} \cite{CVPR2023AdaMAE} adopts an adaptive sampling method that, based on semantic context, utilizes an auxiliary sampling network to sample visible tokens. It estimates a classification distribution concerning spatio-temporal block tokens, selecting tokens that increase the expected reconstruction error as visible tokens. \textbf{VideoMAE.v2} \cite{cvpr2023VideoMAEV2}  introduces a dual-masking strategy where the encoder operates on a subset of video tokens, and the decoder deals with another subset of video tokens. 
\textbf{MotionMAE} \cite{Yang2022motionmae} reconstructs masked video patches and predicts motion structure, leveraging an asymmetric MAE architecture to outperform existing baselines in action classification and video object segmentation by effectively capturing both static and dynamic information in videos. \textbf{OmniMAE} \cite{2022OmniMAE} uses masked autoencoding with spatiotemporal patches to train on both images and videos, achieving competitive results in downstream tasks by reconstructing missing patches and applying pixel reconstruction loss. 
\textbf{MAM2} \cite{2022MAM2}  enhances self-supervised video transformer pre-training by separately decoding motion cues using RGB difference as a prediction target, achieving competitive video recognition performance with fewer pre-training epochs.

\subsubsection{AR-Based Video Generation}
AR-based models typically aim at video prediction or video generation tasks, often employing VQ or GPT architectures to model video data. Given that video information is more redundant and higher-dimensional compared to image information, autoregressive models usually predict sequentially along one dimension at a time. Therefore, it is necessary to convert video data into tokens. In AR-based models, the design of the tokenizer is often crucial. Typically, some methods break videos into 2D patches across space and time to get space-time tokens. More sophisticated tokenizers divide the video into 3D voxels and vector quantize these voxel features to obtain discrete visual tokens. 

Different from existing methods applying VQ-encoders on super voxel (3D-VQ), \textbf{MGVIT} \cite{cvpr2023MAGVIT} expand all 2D convolutions inVQ-GAN to 3D convolutions with a temporal axis, and combines 3D-VQ with VQ-GAN to design a new 3D-VQGAN architecture.
\textbf{MaskViT} \cite{Gupta2022MaskViTMV} employs an MAE-based architecture for video prediction, utilizing spatial and spatiotemporal window attention to enhance memory and training efficiency. \textbf{FMNet} \cite{acmmm2022fmnet} predicts the depth of masked frames using adjacent frames, and by reconstructing the masked temporal features, it improves temporal consistency.

\subsection{Detection And Recognition}

\subsubsection{General Detection}
\textbf{iTPN}~\cite{Tian2022IntegrallyPT} enhances the pre-training phase by incorporating a feature pyramid, unifying the reconstruction and recognition neck, and supplementing MIM with masked feature modeling, providing multi-stage supervision. 

\textbf{MIMdet}~\cite{Fang2022UnleashingVV} finds that a MIM pre-trained Vanilla ViT encoder can perform surprisingly well in challenging object-level recognition scenarios, even with randomly sampled partial observations. \textbf{imTED}~\cite{Zhang2022IntegrallyMP} migrates a pre-trained Transformer encoder-decoder to a target detector, constructing a fully pre-trained feature extraction pathway to enhance the detector's generalization capability while introducing a multi-scale feature modulator for scale adaptability.

\subsubsection{Downstream Classification}
\textbf{Face Recognition}. \textbf{FaceMAE}~\cite{Wang2022FaceMAEPF} randomly masks face images to train the MIM head as MAE~\cite{he2022masked}. An instance relation matching module is tailored to minimize the distribution gap between real faces and the reconstructed ones.

\textbf{Knowledge Distillation}.
\textbf{G2SD} \cite{cvpr2023G2SD} introduces two KD processes to enhance the potential of smaller ViT models. During the generic distillation phase, the smaller model's decoder is encouraged to align its feature predictions with the hidden representations of the larger model, thereby transferring task-agnostic knowledge. In the specific distillation phase, the smaller model's predictions are constrained to be consistent with the larger model's predictions, transferring task-specific features that ensure task performance. \textbf{DMAE}~\cite{Bai2022MaskedAE} introduces a computationally efficient KD framework that leverages MAE to align intermediate feature maps between teacher and student models, enabling robust knowledge transfer and improved performance with high masking ratios and limited visible patches.

\textbf{Efficient Fine-tuning}. 
\textbf{Robust Fine-tuing}~\cite{Xiao2023MaskedIA} presents a technique that uses masked image patches for counterfactual sample generation, enhancing model robustness by breaking spurious correlations during fine-tuning of large pre-trained models.
\textbf{MAE-CT}~\cite{Lehner2023ContrastiveTA}
employs Nearest Neighbor CL to refine the top layers of a pre-trained MAE, enabling it to form semantic clusters and improve performance on classification tasks without the need for labeled data.
\textbf{MAE-CIL} \cite{Zhai2023MaskedAA} explores a bilateral MAE framework for Class Incremental Learning, enhancing image reconstruction quality and representation stability through a novel fusion of image-level and embedding-level learning,

\subsubsection{Medical Image}
\textbf{SD-MAE} \cite{Luo2022SelfdistillationAM}  performs region masking and reconstruction on histology images to learn useful representations. Additionally, self-distillation is introduced by making the student model mimic the outputs of the teacher autoencoder via a hint loss. 
\textbf{MedMAE} \cite{Zhou2022SelfPW} migrates MIM to medical images and appends task-specific Heads for specific tasks. It achieves commendable results in various tasks such as chest X-ray disease classification, abdominal CT multi-organ segmentation, and MRI brain tumor segmentation.
\textbf{FreMAE} \cite{2023FreMAE} explores the potential of using Fourier Transform for masked image modeling in medical image segmentation, integrating both global structural information and local details. This is achieved by leveraging the frequency domain and multi-stage supervision. \textbf{GCMAE} \cite{2022gcmae} employs MIM for representation learning in the computational pathology domain, effectively extracting both global and local features from pathological images.

\subsubsection{OCR}
\textbf{DocMAE}~\cite{icme2023DocMAE} proposes a self-supervised framework  that leverages masked autoencoders to learn rectification models for document image correction without human annotation.  \textbf{MaskOCR}~\cite{2022MaskOCR} presents a novel pre-training approach that uses masked image modeling to learn robust encoder-decoder architectures for text recognition in a self-supervised manner without text annotations.

\subsubsection{Remote sensing}
Based on \textbf{MAE}, \textbf{SatMAE}~\cite{2022SatMAE}  incorporates a temporal embedding and independently masks image patches across time to harness the temporal information present in the data. This approach allows the model to learn from the changes in the data over time, providing a richer and more nuanced understanding of the imagery.
\textbf{CMID} \cite{TGRS2023CMID} is capable of learning both global semantic separable and local spatial perceptible representations by combining CL with MIM in a self-distillation manner. This approach addresses the limitations of existing RS SSL methods, which typically focus on either global or local representations, and is better suited to the varied and complex representations required for different RS downstream tasks.

\subsubsection{Low-Level Vision}
Deep learning has achieved remarkable results in various image tasks, but they often struggle to generalize across different noise distributions. \textbf{MaskedDenoising}~\cite{Chen2023MaskedIT}  masks feature in the self-attention layer to address inconsistencies between training and testing based on MAE. 
\textbf{DreamTeacher} \cite{Li2023DreamTeacherPI} employs two KD methods for pre-training image backbones and performing image denoising: feature distillation and label distillation. Feature distillation transfers features from the generative model to the target backbone, while label distillation transfers task-specific labels. 

\subsection{3D Vision Task}

\subsubsection{Depth Estimation}
\textbf{Mesa} \cite{Khan2023MeSaMG} introduces a novel pre-training framework that synergizes masked, geometric, and supervised learning to enhance the representation of later layers in monocular depth estimation models.\textbf{UniPAD} \cite{Yang2023UniPADAU} introduces a SSL paradigm that utilizes 3D volumetric differentiable rendering for encoding 3D space and reconstructing 3D shapes, significantly enhancing performance in autonomous driving tasks like 3D object detection and semantic segmentation.

\subsubsection{3D Point CLoud}
Research on 3D point clouds can primarily be divided into three categories: one applies the foundational architecture of  MIM to 3D point cloud data, another combines it with CL, and the last category utilizes different network architectures based on the MIM framework.

\begin{figure}[t!]
    \centering
    \vspace{-0.25em}
    \includegraphics[width=1.0\linewidth]{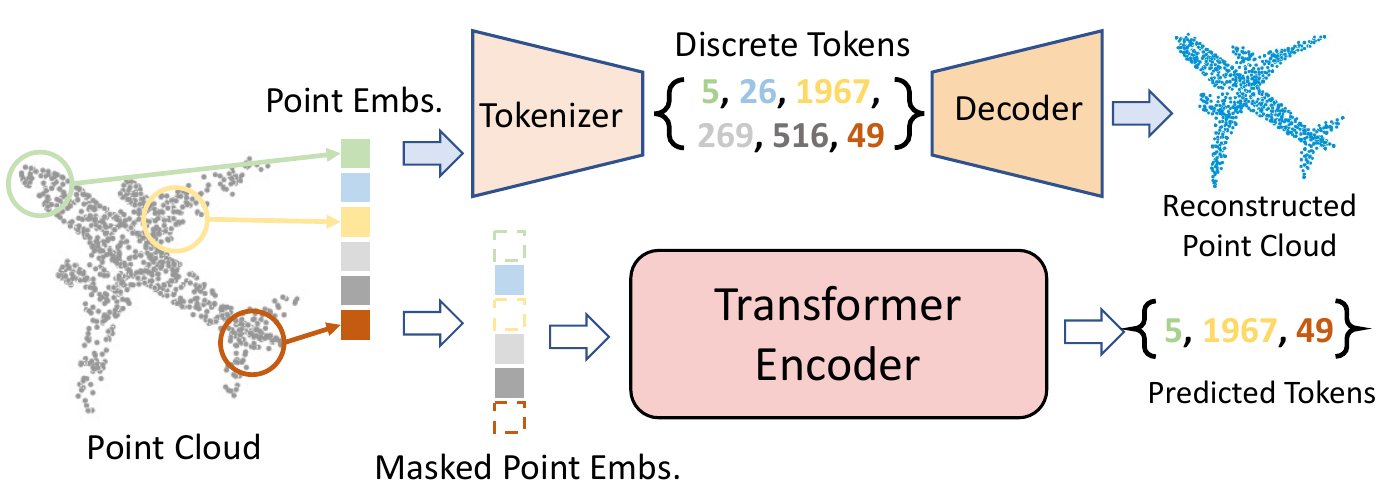}
    \caption{Illustration of MIM on Point Cloud. Taking PointBERT~\cite{cvpr2022pointbert} as an example, it partitions point clouds into local point chunks and employs a point cloud Tokenizer to generate discrete tokens. The figure is reproduced from \cite{cvpr2022pointbert}. 
    }
    \label{fig:PointBert}
    \vspace{-1.0em}
\end{figure}

\textbf{Basic MIM}.
To adapt the 2D MAE framework to 3D point clouds, a common approach is voxelization - converting the irregular point cloud into a regular 3D voxel grid that can then be masked. One method masks contiguous 3D voxels to extend patch masking. Encoder architectures like sparse 3D CNNs help capture 3D spatial context. Alternately, some methods work directly on raw point clouds using specialized encoders.
For tokenization, point clouds are often voxelized first before applying 3D convolutional autoencoders to learn discrete voxel tokens. Other approaches cluster point cloud features into visual words without voxelization. Hybrid tokenizers combine both voxel and raw point features. 
\textbf{MAE-Based}:
\textbf{Voxel-MAE} \cite{2022VoxelMAE} introduces a distance-based random masking strategy and an occupancy prediction pretext task, which helps the model predict the occluded occupancy structure of 3D scenes. \textbf{PointMAE} \cite{Zhang2022PointM2AEMM} divides the input point cloud into patches, randomly masks them, and uses a Transformer encoder to learn high-level latent features from unmasked patches. \textbf{I2P-MAE} \cite{cvpr2022Learning3R}  focuses on geometric feature reconstruction and identifies three self-supervised learning objectives specific to point clouds: centroid prediction, normal estimation, and curvature prediction.
\textbf{ACT}~\cite{iclr2023act} utilizes pre-trained 2D image or language Transformers as teachers for 3D representation learning, transferring their latent features to a 3D Transformer student through masked point modeling. \textbf{MaskPoint}~\cite{Liu2022MaskedPT} introduces a discriminative masked pre-training framework that represents point clouds as discrete occupancy values and performs binary classification between points of masked objects and sampled noise. \textbf{GeoMAE}~\cite{cvpr2023GeoMAE} employs a Transformer to process a set of randomly masked points, and then uses a lightweight Transformer to predict the centroid, normals, and curvature for each voxel in the point, enabling the model to infer the fine-grained geometric structure.
\textbf{BEiT-Based}: \textbf{PointBERT}~\cite{cvpr2022pointbert} partitions point clouds into local point chunks and employs a point cloud Tokenizer to generate discrete tokens. It randomly masks certain chunks of the input point cloud and recovers the original point tokens at the masked positions, as shown in Figure~\ref{fig:PointBert}.

\textbf{Combined with CL}.
\textbf{PointCMP}~\cite{cvpr2023PointCMP} integrates the learning of both local and global spatiotemporal features using a two-branch structure. A mutual similarity-based augmentation module is introduced to generate hard samples at the feature level. 
\textbf{ReCon}~\cite{Qi2023ContrastWR} trains a generative student to guide a contrastive student using an encoder-decoder style RECON-block that transfers knowledge through cross attention with stop-gradient. This approach avoids overfitting and pattern difference issues, achieving remarkable results in 3D representation learning and improving performance on downstream tasks.

\textbf{Different Architecture}.
\textbf{Point-M2AE}~\cite{Zhang2022PointM2AEMM}: The encoder and decoder are redesigned into a pyramid structure to capture the spatial geometry and semantic information of 3D shapes. Then, a multi-scale masking strategy is designed to generate consistently visible regions across different scales.

\if\submission\submissionTPAMI
    \vspace{-1.0em}
    \section{Masked Modeling on Other Modalities}
    \label{sec:beyond}
    This section further extends masked modeling pre-training to other mainstream domains beyond CV and NLP and summarizes the essential design and applications.
    \vspace{-1.0em}
\else
    \section{Masked Modeling on Other Modalities}
    \label{sec:beyond}
    This section further extends masked modeling pre-training to other mainstream domains beyond CV and NLP and summarizes the essential design and applications.
\fi

\subsection{Audio and Speech}
\textbf{Combining CL with Masked Modeling.}
The concept of applying the masked modeling mechanism for SSL can be expanded to audio signals. \textbf{VQ-wav2vec}~\cite{nips2020wav2vec2} introduces \textbf{BERT}-style masked modeling as pre-training on top of \textbf{wav2vec}~\cite{2019vqwav2vec}. In \textbf{wav2vec}, the input audio signal is first mapped into dense latent representations by an encoder network. Aggregating latent representations from multiple time steps, the context network generates a contextualized representation. A CL is adopted as the objective function motivated by Contrastive Predictive Coding (CPC)~\cite{oord2018representation}. \textbf{VQ-wav2vec}~\cite{nips2020wav2vec2} introduces a quantization module to replace the dense latent representations with discrete representations, similar to \textbf{VQ-VAE}. The resulting discretized audio representations facilitate a seamless application of the original \textbf{BERT}-style masked modeling, which requires a discrete vocabulary. \textbf{wav2vec 2.0} adopts a Transformer as the context network in contrast to the \textbf{wav2vec}, which uses CNNs for both networks. The output from the convolutional encoder is randomly masked before feeding into the Transformer. \textbf{InfoNCE} is adopted to maximize the similarity between the contextualized representation at the masked time stamps and the corresponding quantized version of the localized representation where negative samples are drawn from other masked time steps. Apart from creating the discrete inputs as input to \textbf{BERT} using a quantization module, \textbf{Hidden Unit BERT (HuBERT)}~\cite{taslp2021hubert} discretize the prediction target by coming up with cluster labels provided by applying K-means to Mel Frequency Cepstral Coefficients (MFCC) of the input audio. HuBERT adopts the same architecture design as in \textbf{wav2vec 2.0}, where the CNN audio encoder and the Transformer \textbf{BERT} encoder are adopted. The categorical cross-entropy loss is employed to assess the hidden cluster assignment performance for masked and unmasked tokens, similar to a frame-level acoustic unit discovery problem. It is essential to highlight that while the masking operation is a common element in \textbf{VQ-wav2vec}, \textbf{wav2vec 2.0}, and \textbf{HuBERT}, only \textbf{VQ-wav2vec} and \textbf{HuBERT} incorporate a \textbf{BERT}-style masked modeling approach, whereas \textbf{wav2vec 2.0} employs the \textbf{BERT}-style masking operation as a means to enhance the performance of CL. 

\textbf{Masked Audio Modeling as MIM.}
In contrast to the common practice in MIM, where the prediction task usually takes the form of regression, regardless of whether the prediction target involves tokenizers, pixels, or features, it is worth noting that \textbf{VQ-wav2vec} and \textbf{HuBERT}, rigorously adhere to categorization. The pivotal connection uniting MIM and masked audio modeling (MAM) is the transformation from raw audio signals to a visual representation of either spectrogram or mel-spectrogram. Treating the spectrogram as a greyscale image, the problem of MAM can be naturally and directly transformed into the problem of MIM~\cite{liu2020mockingjay, chi2021audio, 2021mam, 2022MAEAST, 2022MaskSpec, huang2022masked}. The difference between these works again resides in the design of the modules for \textbf{Mask}, \textbf{Target}, \textbf{Encoder}, and \textbf{Head}. Since the spectrogram itself has already extracted features of the audio signal, the main difference is whether the masked patches are fed into the encoder. Only unmasked patches are fed into the encoder in \textbf{Audio-MAE}, while works like \textbf{Mockingjay}~\, cite{liu2020mockingjay} and \textbf{Audio ALBERT}~\cite{chi2021audio} pass both masked and unmasked patches into the encoder. \textbf{Audio-MAE}~\cite{huang2022masked} explores different masking strategies of unstructured masking (random patch masking), time masking (column-wise masking), and frequency masking (row-wise masking).
\if\submission\submissionarXiv  
    The framework of Audio-MAE is shown in Figure \ref{fig:AudioMAE}.
\fi
Combining MAM and MIM, \textbf{Audiovisual MAE}~\cite{iccv2023audiovision} simultaneously applied the masked modeling to audio and image for video pre-training.

\vspace{-1.0em}
\subsection{Graph Representation}
Graph data are in real-world practice, \textit{e.g.}, social networks. Masked modeling has also achieved overwhelming success in graph data analysis. Initially, \textbf{AttrMasking}~\cite{hu2019strategies} first masks some proportions of nodes and edges within each graph and trains the GNN encoder to predict them. Analogously, \textbf{GROVER}~\cite{rong2020self} attempts to predict the masked subgraphs. Subsequently, \textbf{GPT-GNN}~\cite{hu2020gpt} proposes an autoregressive framework to perform node and edge reconstruction iteratively, which generates one masked node (atom) and its connected edges (bonds) and optimizes the likelihood of the node and edges generation in the next iteration. More recently, inspired by the huge success of MAE~\cite{he2022masked} in CV, \textbf{GraphMAE}~\cite{hou2022graphmae} masks some input node features with special tokens and enforces the graph autoencoder to reconstruct the masked ones. \textbf{GraphMAE2}~\cite{hou2023graphmae2} argues that GraphMAE is usually vulnerable to disturbance in the features. To mitigate this issue, they designed the multi-view random re-mask decoding and latent representation prediction to regularize the feature reconstruction.
Similarly, \textbf{MGAE}~\cite{tan2022mgae} observes that a high masking ratio of the input graph edges could benefit the downstream tasks. They also propose a tailored cross-correlation decoder to reconstruct the large number of masked edges. With the increasing attention paid to Graph Transformer, \textbf{GMAEs}~\cite{zhang2022graph} designs an asymmetric Graph Transformer~\cite{min2022transformer} framework, where the encoder is a deep Transformer and the decoder is a shallow Transformer. Equipped with the masking mechanism, GMAE is more memory-efficient than classical Transformers. Despite the fruitful progress, the masking operations create an undesirable dispensary between pre-training and finetuning because the masks would not appear in the downstream tasks. It remains promising to tackle this crucial issue.

\vspace{-1.0em}
\subsection{Biology and Chemistry}
Masked modeling has recently been extended to various biological applications to accelerate biochemical experiments, especially for research on proteins and molecules.

\textbf{Sequence Modeling for Protein.}
Considering an amino acid in the protein sequence as a word in the sentence, a number of self-supervised tasks proposed for natural language can be naturally extended to protein sequences. \textbf{TAPE}~\cite{rao2019evaluating} proposes to predict the type of the next amino acid based on a set of masked sequence fragments. \textbf{ESM-1b}~\cite{rives2021biological} randomly masks out a single or a set of contiguous amino acids and then predicts the masked amino acids from the remaining sequences. Unlike random masking, \textbf{AC-MLM}~\cite{mcdermott2021adversarial} combines adversarial training with masked language modeling and proposes to mask amino acids in a learnable and adversarial manner. Taking into account the dependence between masked amino acids, \textbf{Pairwise MLM (PMLM)}~\cite{he2021pre} proposes to model the probability of a pair of masked amino acids instead of predicting the probability of a single amino acid. Different from these generative methods, \textbf{CPCProt}~\cite{lu2020self} applies different masking transformations on the input sequences to generate different views and then applies \textbf{InfoNCE} to maximize the similarity of two jointly sampled pairs. The antibody is a special kind of protein, and \textbf{ABGNN}~\cite{gao2023pre} enables pre-training of antibody sequences by masking the residues on the Compound Determining Regions (CDRs) and predicting the types of masked residues.

\textbf{Sequence-structure Co-modeling for Protein.}
The amino acid sequences of proteins can be folded into stable 3D structures in the real physicochemical world, forming a special kind of sequence-structure data. The concept of the masked modeling mechanism for SSL can also be expanded to protein structure pre-training. \textbf{GearNet} \cite{zhang2022protein} proposes multiview contrasting that randomly samples two sub-structures from each protein by masking, encoders them into two representations, and finally maximizes the similarity between representations from the same protein while minimizing the similarity between representations from different proteins. \textbf{GraphComp}~\cite{you2022cross} proposes graph completion, which takes as input a protein graph with partially masked residues and then makes predictions for those masked tokens. \textbf{AlphaFold2} \cite{jumper2021highly} takes masked language modeling as a pre-training task and full-atomic structure prediction as a downstream task. It was found by \cite{hu2022exploring} that the representations from AlphFold2's \textbf{Evoformer} could work well on various protein-related downstream tasks, including fold classification, stability prediction, \textit{etc}. Moreover, \textbf{Masked Inverse Folding (MIF)} \cite{yang2022masked} trains a model to reconstruct the original amino acids conditioned on the masked sequence and the masked backbone structure. Similar to MAGE~\cite{cvpr2023mage}, more recently proposed SSL methods~\cite{su2023saprot, Gao2023VQPLVQ} like \textbf{FoldSeek}~\cite{nature2023foldseek} first expand the codebook for amino acid sequences with VQVAE and than perform masked modeling for the latent Transformer encoder.

\if\submission\submissionarXiv  
\textbf{Graph Representation for Molecules.}
Most molecule data can be represented as SMILE sequences or 2D/3D graphs. Therefore, many methods developed for languages or graphs can also be directly transferred to molecules. \textbf{AttrMasking}~\cite{hu2019strategies} randomly masks the input node and edge attributes (\textit{e.g.}, atom types in the molecular graph) and applies GNNs to predict the masked attributes. For sequence-based masking, \textbf{SMILES-BERT}~\cite{wang2019smiles} and \textbf{Molformer}~\cite{ross2022large} randomly mask the characters in the SMILES sequences and then reconstruct them from the encoded features. To alleviate the problem of imbalance atom types in nature, \textbf{Mole-BERT}~\cite{xia2022mole} designs a context-aware tokenizer that encodes atoms as chemically meaningful discrete codes for masking modeling on embedded codes as BEiT~\cite{Bao2021BEiT}.
\fi

\if\submission\submissionarXiv  
\begin{figure}[t!]
    \centering
    \includegraphics[width=1.0\linewidth]{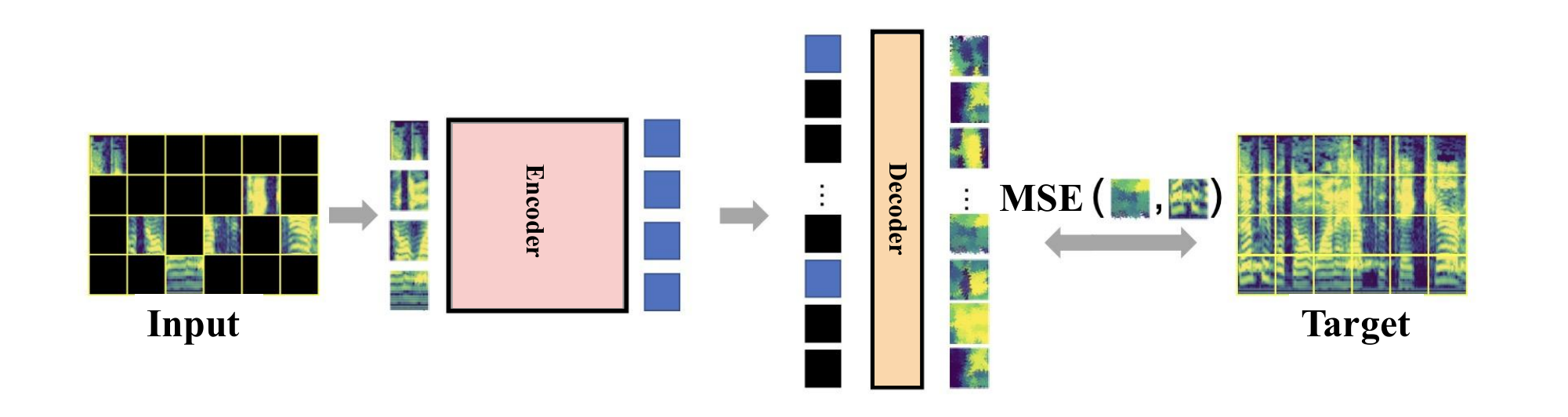}
    \caption{Illustration of MIM on Audio. Taking Audio-MAE~\cite{nips2022AudioMAE} as an example, it directly applies the MAE framework to audio. The figure is reproduced from \cite{nips2022AudioMAE}. 
    }
    \label{fig:AudioMAE}
    \vspace{-0.5em}
\end{figure}
\fi

\if\submission\submissionarXiv
    \if\submission\submissionTPAMI
    \subsection{Discussions and Future Directions}  
\else
    \section{Discussions and Future Directions}  
\fi
\label{section:discussion}

\textbf{How to design an efficient MIM Model?}
This paper sets out from its main arguments to offer recommendations and heuristic considerations for designing efficient Masked Image Modeling models. The essence of Masked Modeling lies in the reconstruction using masked data. In NLP, the masked tokens are often several consecutive tokens, an operation grounded in a critical principle: preventing information leakage and enabling the model to work with minimal prior information, thereby increasing the difficulty of the reconstruction task. Therefore, when designing the structure of Masked Modeling, the Masked part should adhere to the principle of preventing information leakage. The \textbf{attention-based masking} strategy, while considering the avoidance of data information leakage, utilizes the least computational resources. Furthermore, as introduced in section 3, Masked Modeling's task of reconstructing low-level features and details compensates for the inadequacies of \textbf{Transformers} in detail modeling. Coupled with the Transformer's inherent global modeling capabilities, the combination of Masked Modeling and Transformer enables the model to accommodate both low-level modeling capabilities and global modeling abilities, thereby further raising the upper limit of model performance.
The selection of Head and Target parts should be contingent upon the specific task. Different Targets will induce varying biases in the model and yield different effects in diverse tasks. Feature maps are generally more suitable for detection tasks. As for whether the Head part should be combined with contrastive learning, this should depend on the choice of Target. If the selected Target necessitates the extraction of a feature map, contrastive learning could be conveniently used to enhance model performance. Conversely, if the model uses Pixels as the Target, employing contrastive learning would not significantly improve performance and would incur substantial computational costs.

\textbf{Explainability of MIM.}
Compared to contrastive learning, Masked Modeling still lacks a more comprehensive explanation. The task of contrastive learning, utilizing the InfoNCE loss function, offers a complete loss function and a relatively unified architecture with clearer task objectives. In contrast, Masked Modeling involves complex processing techniques within its various modules and across different modalities. For Masked Modeling, employing different masking strategies and tokenization methods to compress data can result in significant structural and computational differences, making it challenging to develop a comprehensive and unified theoretical explanation.
Currently, most theoretical explanations are specific to particular tasks or based on empirical studies, and they fail to generalize across various modalities. The prevailing explanatory approaches mainly unfold in three directions: interpretation based on hierarchical structures, explanations derived from the theoretical foundations of contrastive learning, and interpretations from the perspective of information compression. Although these research efforts provide a certain degree of interpretability to Masked Modeling, they still lack a profound theoretical basis. This makes the interpretability of Masked Modeling a challenging research direction.

\textbf{Downstream Task.}
Current research on downstream tasks mainly focuses on applying the  MAE architecture to specific downstream task structures. However, with the robust growth of Masked Modeling, more complex technologies are gradually being introduced into these tasks. In video research, GPT and MAE are two critical backbones, but a series of studies combining VQ-based models with Masked Modeling are increasingly emerging in the field. These studies employ VQ technology for more efficient data compression and tokenize data to achieve higher-quality reconstruction. Therefore, we believe that research on 3D point clouds will follow this development trend, combining VQ-based models with Masked Modeling to achieve better information compression efficiency.

\textbf{Other Domains Beyond CV and NLP.} 
Multimodal research is currently a significant direction in artificial intelligence, and the application of Masked Modeling in multimodal contexts is one of the most promising future directions. Early multimodal research primarily employed contrastive learning, aligning different modalities and computing contrastive loss. With the advancement of diffusion techniques, studies aligning different modalities through diffusion are also increasing. Masked Modeling holds potential in multimodal applications. The current research paradigm mainly involves aligning different modalities after masking them, increasing task complexity. A new research paradigm is also emerging, where data from different modalities are aligned to a central modality, and then Masked Modeling is applied using the central modality's data.
Moreover, applying Masked Modeling to various modalities technically poses more challenges. Extending masking to 3D, 4D, or even higher-dimensional data and tokenizing higher-dimensional data are technical details that need attention and resolution when expanding Masked Modeling to higher dimensions. Therefore, integration with multimodal approaches will be an important research direction for Masked Modeling.
\fi

\if\submission\submissionTPAMI
    \vspace{-1.0em}
    \section{Conclusion}
    \label{section:conclusion}
    \textbf{Discussions.} We provide discussions on some future directions of masked modeling in Appendix~\ref{section:discussion}, including how to design an efficient MIM Model, explainability of MIM, CV downstream tasks, and other domains beyond CV and NLP.

    \textbf{Conclusions.}
    This survey, grounded in CV, proposes a unified architecture for Masked Modeling, successfully integrating various technical details and data modalities within this framework. Additionally, we have meticulously organized and elucidated technologies related to Masked Modeling, such as CL, generative models, and autoregressive models, offering readers a more comprehensive perspective. This paper presents a complete exposition of Masked Modeling's applications and theoretical aspects, detailing its use in various visual tasks as well as Beyond Vision tasks and discussing the current theoretical achievements and progress in Masked Modeling. Based on this, we propose promising future directions for Masked Modeling, aligned with current hot research topics in the artificial intelligence community, such as multimodality and large models, providing readers with ideas for proposing new methods based on this survey.
\else
    \section{Conclusion}
    \label{section:conclusion}
    This survey, grounded in CV, proposes a unified architecture for Masked Modeling, successfully integrating various technical details and data modalities within this framework. Additionally, we have meticulously organized and elucidated technologies related to Masked Modeling, such as CL, generative models, and autoregressive models, offering readers a more comprehensive perspective. This paper presents a complete exposition of Masked Modeling's applications and theoretical aspects, detailing its use in various visual tasks as well as Beyond Vision tasks and discussing the current theoretical achievements and progress in Masked Modeling. Based on this, we propose promising future directions for Masked Modeling, aligned with current hot research topics in the artificial intelligence community, such as multimodality and large models, providing readers with ideas for proposing new models and methods based on this survey.

    \section*{ACKNOWLEDGMENTS}
    This work was supported by the National Key R\&D Program of China (No. 2022ZD0115100), the National Natural Science Foundation of China Project (No. U21A20427), and Project (No. WU2022A009) from the Center of Synthetic Biology and Integrated Bioengineering of Westlake University.
    This work was done by Luyuan Zhang and Zedong Wang during their internship at Westlake University.
\fi


%

\ifCLASSOPTIONcaptionsoff
  \newpage
\fi



{
%
\bibliographystyle{abbrv}
\bibliography{ref}

\begin{thebibliography}{100}

\bibitem{Assran2022MaskedSN}
M.~Assran, M.~Caron, I.~Misra, P.~Bojanowski, F.~Bordes, P.~Vincent, A.~Joulin,
  M.~G. Rabbat, and N.~Ballas.
\newblock Masked siamese networks for label-efficient learning.
\newblock In {\em ECCV}, 2022.

\bibitem{cvpr2023IJEPA}
M.~Assran, Q.~Duval, I.~Misra, P.~Bojanowski, P.~Vincent, M.~Rabbat, Y.~LeCun,
  and N.~Ballas.
\newblock Self-supervised learning from images with a joint-embedding
  predictive architecture.
\newblock In {\em CVPR}, 2023.

\bibitem{2022MAEAST}
A.~Baade, P.~Peng, and D.~F. Harwath.
\newblock Mae-ast: Masked autoencoding audio spectrogram transformer.
\newblock {\em ArXiv}, 2022.

\bibitem{Bachmann2022MultiMAEMM}
R.~Bachmann, D.~Mizrahi, A.~Atanov, and A.~R. Zamir.
\newblock Multimae: Multi-modal multi-task masked autoencoders.
\newblock {\em ArXiv}, 2022.

\bibitem{2022Data2Vec2}
A.~Baevski, A.~Babu, W.-N. Hsu, and M.~Auli.
\newblock Efficient self-supervised learning with contextualized target
  representations for vision, speech and language.
\newblock 2022.

\bibitem{Baevski2022data2vecAG}
A.~Baevski, W.-N. Hsu, Q.~Xu, A.~Babu, J.~Gu, and M.~Auli.
\newblock data2vec: A general framework for self-supervised learning in speech,
  vision and language.
\newblock In {\em ICML}, 2022.

\bibitem{2019vqwav2vec}
A.~Baevski, S.~Schneider, and M.~Auli.
\newblock vq-wav2vec: Self-supervised learning of discrete speech
  representations.
\newblock {\em ArXiv}, 2019.

\bibitem{nips2020wav2vec2}
A.~Baevski, H.~Zhou, A.~rahman Mohamed, and M.~Auli.
\newblock wav2vec 2.0: A framework for self-supervised learning of speech
  representations.
\newblock {\em ArXiv}, 2020.

\bibitem{bai2023sequential}
Y.~Bai, X.~Geng, K.~Mangalam, A.~Bar, A.~Yuille, T.~Darrell, J.~Malik, and
  A.~A. Efros.
\newblock Sequential modeling enables scalable learning for large vision
  models, 2023.

\bibitem{Bai2022MaskedAE}
Y.~Bai, Z.~Wang, J.~Xiao, C.~Wei, H.~Wang, A.~L. Yuille, Y.~Zhou, and C.~Xie.
\newblock Masked autoencoders enable efficient knowledge distillers.
\newblock {\em CVPR}, pages 24256--24265, 2022.

\bibitem{CVPR2023AdaMAE}
W.~G.~C. Bandara, N.~Patel, A.~Gholami, M.~Nikkhah, M.~Agrawal, and V.~M.
  Patel.
\newblock Adamae: Adaptive masking for efficient spatiotemporal learning with
  masked autoencoders.
\newblock In {\em CVPR}, pages 14507--14517, 2023.

\bibitem{Bao2021BEiT}
H.~Bao, L.~Dong, and F.~Wei.
\newblock Beit: Bert pre-training of image transformers.
\newblock In {\em ICLR}, 2022.

\bibitem{Baraldi2023LearningTM}
L.~Baraldi, R.~Amoroso, M.~Cornia, A.~Pilzer, and R.~Cucchiara.
\newblock Learning to mask and permute visual tokens for vision transformer
  pre-training.
\newblock {\em ArXiv}, 2023.

\bibitem{bau2018gan}
D.~Bau, J.-Y. Zhu, H.~Strobelt, B.~Zhou, J.~B. Tenenbaum, W.~T. Freeman, and
  A.~Torralba.
\newblock Gan dissection: Visualizing and understanding generative adversarial
  networks.
\newblock {\em arXiv}, 2018.

\bibitem{BetkerImprovingIG}
J.~Betker, G.~Goh, L.~Jing, TimBrooks, J.~Wang, L.~Li, LongOuyang,
  JuntangZhuang, JoyceLee, YufeiGuo, WesamManassra, PrafullaDhariwal, CaseyChu,
  YunxinJiao, and A.~Ramesh.
\newblock Improving image generation with better captions.

\bibitem{Campos2016MSMA}
D.~F. Campos, T.~Nguyen, M.~Rosenberg, X.~Song, J.~Gao, S.~Tiwary, R.~Majumder,
  L.~Deng, and B.~Mitra.
\newblock Ms marco: A human generated machine reading comprehension dataset.
\newblock {\em ArXiv}, 2016.

\bibitem{Carreira2019ASN}
J.~Carreira, E.~Noland, C.~Hillier, and A.~Zisserman.
\newblock A short note on the kinetics-700 human action dataset.
\newblock {\em ArXiv}, 2019.

\bibitem{Casas2021MP3AU}
S.~Casas, A.~Sadat, and R.~Urtasun.
\newblock Mp3: A unified model to map, perceive, predict and plan.
\newblock {\em CVPR}, pages 14398--14407, 2021.

\bibitem{Chang2015ShapeNetAI}
A.~X. Chang, T.~A. Funkhouser, L.~J. Guibas, P.~Hanrahan, Q.-X. Huang, Z.~Li,
  S.~Savarese, M.~Savva, S.~Song, H.~Su, J.~Xiao, L.~Yi, and F.~Yu.
\newblock Shapenet: An information-rich 3d model repository.
\newblock {\em ArXiv}, 2015.

\bibitem{Chang2022MaskGITMG}
H.~Chang, H.~Zhang, L.~Jiang, C.~Liu, and W.~T. Freeman.
\newblock Maskgit: Masked generative image transformer.
\newblock {\em CVPR}, pages 11305--11315, 2022.

\bibitem{Changpinyo2021Conceptual1P}
S.~Changpinyo, P.~K. Sharma, N.~Ding, and R.~Soricut.
\newblock Conceptual 12m: Pushing web-scale image-text pre-training to
  recognize long-tail visual concepts.
\newblock In {\em CVPR}, pages 3557--3567, 2021.

\bibitem{cvpr2023PiMAE}
A.~Chen, K.~Zhang, R.~Zhang, Z.~Wang, Y.~Lu, Y.~Guo, and S.~Zhang.
\newblock Pimae: Point cloud and image interactive masked autoencoders for 3d
  object detection.
\newblock In {\em CVPR}, pages 5291--5301, June 2023.

\bibitem{Chen2023MaskedIT}
H.~Chen, J.~Gu, Y.~Liu, S.~A. Magid, C.~Dong, Q.~Wang, H.~Pfister, and L.~Zhu.
\newblock Masked image training for generalizable deep image denoising.
\newblock {\em CVPR}, pages 1692--1703, 2023.

\bibitem{Chen2023TrajMAEMA}
H.~Chen, J.~Wang, K.~Shao, F.~Liu, J.~Hao, C.~Guan, G.~Chen, and P.-A. Heng.
\newblock Traj-mae: Masked autoencoders for trajectory prediction.
\newblock {\em ArXiv}, 2023.

\bibitem{Chen2023AutoMAE}
H.~Chen, W.~Zhang, Y.~Wang, and X.~Yang.
\newblock Improving masked autoencoders by learning where to mask.
\newblock {\em ArXiv}, 2023.

\bibitem{chen2022efficient}
J.~Chen, M.~Hu, B.~Li, and M.~Elhoseiny.
\newblock Efficient self-supervised vision pretraining with local masked
  reconstruction.
\newblock {\em arXiv preprint}, 2022.

\bibitem{2021mam}
J.~Chen, M.~Ma, R.~Zheng, and L.~Huang.
\newblock Mam: Masked acoustic modeling for end-to-end speech-to-text
  translation.
\newblock {\em ArXiv}, 2020.

\bibitem{Chen2023MixedAF}
K.~Chen, Z.~Liu, L.~Hong, H.~Xu, Z.~Li, and D.-Y. Yeung.
\newblock Mixed autoencoder for self-supervised visual representation learning.
\newblock {\em ArXiv}, 2023.

\bibitem{Chen2023HumanMACMM}
L.~Chen, J.~Zhang, Y.~rong Li, Y.~Pang, X.~Xia, and T.~Liu.
\newblock Humanmac: Masked motion completion for human motion prediction.
\newblock {\em ArXiv}, 2023.

\bibitem{Chen2020GenerativePF}
M.~Chen, A.~Radford, J.~Wu, H.~Jun, P.~Dhariwal, D.~Luan, and I.~Sutskever.
\newblock Generative pretraining from pixels.
\newblock In {\em ICML}, 2020.

\bibitem{chen2020simple}
T.~Chen, S.~Kornblith, M.~Norouzi, and G.~Hinton.
\newblock A simple framework for contrastive learning of visual
  representations.
\newblock {\em arXiv preprint}, 2020.

\bibitem{Chen2022ContextAF}
X.~Chen, M.~Ding, X.~Wang, Y.~Xin, S.~Mo, Y.~Wang, S.~Han, P.~Luo, G.~Zeng, and
  J.~Wang.
\newblock Context autoencoder for self-supervised representation learning.
\newblock {\em ArXiv}, 2022.

\bibitem{Chen2022SdAESM}
Y.~Chen, Y.~Liu, D.~Jiang, X.~Zhang, W.~Dai, H.~Xiong, and Q.~Tian.
\newblock Sdae: Self-distillated masked autoencoder.
\newblock In {\em ECCV}, 2022.

\bibitem{Chen2022MaskguidedVT}
Y.~Chen, Z.~Xiao, L.~Zhao, L.~Zhang, H.~Dai, D.~Liu, Z.~Wu, C.~Li, T.~Zhang,
  C.~Li, D.~Zhu, T.~Liu, and X.~Jiang.
\newblock Mask-guided vision transformer (mg-vit) for few-shot learning.
\newblock {\em ArXiv}, 2022.

\bibitem{Chen2023InternVLSU}
Z.~Chen, J.~Wu, W.~Wang, W.~Su, G.~Chen, S.~Xing, Z.~Muyan, Q.~Zhang, X.~Zhu,
  L.~Lu, B.~Li, P.~Luo, T.~Lu, Y.~Qiao, and J.~Dai.
\newblock Internvl: Scaling up vision foundation models and aligning for
  generic visual-linguistic tasks.
\newblock 2023.

\bibitem{Cheng2023ForecastMAESP}
J.~Cheng, X.~Mei, and M.-Y. Liu.
\newblock Forecast-mae: Self-supervised pre-training for motion forecasting
  with masked autoencoders.
\newblock {\em ArXiv}, 2023.

\bibitem{chi2021audio}
P.-H. Chi, P.-H. Chung, T.-H. Wu, C.-C. Hsieh, Y.-H. Chen, S.-W. Li, and H.-y.
  Lee.
\newblock Audio albert: A lite bert for self-supervised learning of audio
  representation.
\newblock In {\em SLT}. IEEE, 2021.

\bibitem{2022MaskSpec}
D.~Chong, H.~Wang, P.~Zhou, and Q.~jie Zeng.
\newblock Masked spectrogram prediction for self-supervised audio pre-training.
\newblock {\em ArXiv}, 2022.

\bibitem{Chung2018Speech2VecAS}
Y.-A. Chung and J.~R. Glass.
\newblock Speech2vec: A sequence-to-sequence framework for learning word
  embeddings from speech.
\newblock {\em ArXiv}, 2018.

\bibitem{Coates2011AnAO}
A.~Coates, A.~Ng, and H.~Lee.
\newblock An analysis of single-layer networks in unsupervised feature
  learning.
\newblock In {\em AISTATS}, 2011.

\bibitem{2022SatMAE}
Y.~Cong, S.~Khanna, C.~Meng, P.~Liu, E.~Rozi, Y.~He, M.~Burke, D.~Lobell, and
  S.~Ermon.
\newblock Satmae: Pre-training transformers for temporal and multi-spectral
  satellite imagery.
\newblock {\em ArXiv}, 2022.

\bibitem{Cordts2016TheCD}
M.~Cordts, M.~Omran, S.~Ramos, T.~Rehfeld, M.~Enzweiler, R.~Benenson,
  U.~Franke, S.~Roth, and B.~Schiele.
\newblock The cityscapes dataset for semantic urban scene understanding.
\newblock {\em CVPR}, pages 3213--3223, 2016.

\bibitem{devlin2018bert}
J.~Devlin, M.-W. Chang, K.~Lee, and K.~Toutanova.
\newblock Bert: Pre-training of deep bidirectional transformers for language
  understanding.
\newblock {\em arXiv preprint}, 2018.

\bibitem{Doersch2015UnsupervisedVR}
C.~Doersch, A.~K. Gupta, and A.~A. Efros.
\newblock Unsupervised visual representation learning by context prediction.
\newblock {\em 2015 ICCV}, pages 1422--1430, 2015.

\bibitem{iclr2023act}
R.~Dong, Z.~Qi, L.~Zhang, J.~Zhang, J.~Sun, Z.~Ge, L.~Yi, and K.~Ma.
\newblock Autoencoders as cross-modal teachers: Can pretrained 2d image
  transformers help 3d representation learning?
\newblock In {\em ICLR}, 2023.

\bibitem{Dong2021PeCoPC}
X.~Dong, J.~Bao, T.~Zhang, D.~Chen, W.~Zhang, L.~Yuan, D.~Chen, F.~Wen, and
  N.~Yu.
\newblock Peco: Perceptual codebook for bert pre-training of vision
  transformers.
\newblock In {\em AAAI}, 2021.

\bibitem{Dong2022BootstrappedMA}
X.~Dong, J.~Bao, T.~Zhang, D.~Chen, W.~Zhang, L.~Yuan, D.~Chen, F.~Wen, and
  N.~Yu.
\newblock Bootstrapped masked autoencoders for vision bert pretraining.
\newblock In {\em ECCV}, 2022.

\bibitem{2022MaskCLIP}
X.~Dong, Y.~Zheng, J.~Bao, T.~Zhang, D.~Chen, H.~Yang, M.~Zeng, W.~Zhang,
  L.~Yuan, D.~Chen, F.~Wen, and N.~Yu.
\newblock Maskclip: Masked self-distillation advances contrastive
  language-image pretraining.
\newblock {\em ArXiv}, 2022.

\bibitem{Dosovitskiy2020AnII}
A.~Dosovitskiy, L.~Beyer, A.~Kolesnikov, D.~Weissenborn, X.~Zhai,
  T.~Unterthiner, M.~Dehghani, M.~Minderer, G.~Heigold, S.~Gelly, J.~Uszkoreit,
  and N.~Houlsby.
\newblock An image is worth 16x16 words: Transformers for image recognition at
  scale.
\newblock In {\em ICLR}, 2021.

\bibitem{2021splitmask}
A.~El-Nouby, G.~Izacard, H.~Touvron, I.~Laptev, H.~J{\'e}gou, and E.~Grave.
\newblock Are large-scale datasets necessary for self-supervised pre-training?
\newblock {\em ArXiv}, 2021.

\bibitem{esser2021taming}
P.~Esser, R.~Rombach, and B.~Ommer.
\newblock Taming transformers for high-resolution image synthesis, 2021.

\bibitem{pascal-voc-2007}
M.~Everingham, L.~Van~Gool, C.~K.~I. Williams, J.~Winn, and A.~Zisserman.
\newblock The {PASCAL} {V}isual {O}bject {C}lasses {C}hallenge 2007 {(VOC2007)}
  {R}esults.

\bibitem{Fan2023MotionGuidedMF}
D.~Fan, J.~Wang, S.~Liao, Y.~Zhu, V.~Bhat, H.~J. Santos-Villalobos, M.~V.
  Rohith, and X.~Li.
\newblock Motion-guided masking for spatiotemporal representation learning.
\newblock {\em ArXiv}, 2023.

\bibitem{fang2022corrupted}
Y.~Fang, L.~Dong, H.~Bao, X.~Wang, and F.~Wei.
\newblock Corrupted image modeling for self-supervised visual pre-training.
\newblock {\em arXiv preprint}, 2022.

\bibitem{Fang2023EVA02AV}
Y.~Fang, Q.~Sun, X.~Wang, T.~Huang, X.~Wang, and Y.~Cao.
\newblock Eva-02: A visual representation for neon genesis.
\newblock {\em ArXiv}, 2023.

\bibitem{Fang2022EVAET}
Y.~Fang, W.~Wang, B.~Xie, Q.-S. Sun, L.~Y. Wu, X.~Wang, T.~Huang, X.~Wang, and
  Y.~Cao.
\newblock Eva: Exploring the limits of masked visual representation learning at
  scale.
\newblock {\em ArXiv}, 2022.

\bibitem{Fang2022UnleashingVV}
Y.~Fang, S.~Yang, S.~Wang, Y.~Ge, Y.~Shan, and X.~Wang.
\newblock Unleashing vanilla vision transformer with masked image modeling for
  object detection.
\newblock {\em ArXiv}, 2022.

\bibitem{1384978}
L.~Fei-Fei, R.~Fergus, and P.~Perona.
\newblock Learning generative visual models from few training examples: An
  incremental bayesian approach tested on 101 object categories.
\newblock In {\em CVPRW}, pages 178--178, 2004.

\bibitem{Feichtenhofer2022MaskedAA}
C.~Feichtenhofer, H.~Fan, Y.~Li, and K.~He.
\newblock Masked autoencoders as spatiotemporal learners.
\newblock {\em ArXiv}, 2022.

\bibitem{2022MimCo}
Q.~feng Zhou, C.~Yu, H.~Luo, Z.~Wang, and H.~Li.
\newblock Mimco: Masked image modeling pre-training with contrastive teacher.
\newblock In {\em MM}, 2022.

\bibitem{Gan2023InstructCVIT}
Y.~Gan, S.~Park, A.~Schubert, A.~Philippakis, and A.~M. Alaa.
\newblock Instructcv: Instruction-tuned text-to-image diffusion models as
  vision generalists.
\newblock {\em ArXiv}, 2023.

\bibitem{Gandelsman2022TestTimeTW}
Y.~Gandelsman, Y.~Sun, X.~Chen, and A.~A. Efros.
\newblock Test-time training with masked autoencoders.
\newblock {\em ArXiv}, 2022.

\bibitem{gao2023pre}
K.~Gao, L.~Wu, J.~Zhu, T.~Peng, Y.~Xia, L.~He, S.~Xie, T.~Qin, H.~Liu, K.~He,
  et~al.
\newblock Pre-training antibody language models for antigen-specific
  computational antibody design.
\newblock In {\em SIGKDD}, pages 506--517, 2023.

\bibitem{Gao2022ConvMAEMC}
P.~Gao, T.~Ma, H.~Li, J.~Dai, and Y.~J. Qiao.
\newblock Convmae: Masked convolution meets masked autoencoders.
\newblock {\em ArXiv}, 2022.

\bibitem{Gao2021SimCSESC}
T.~Gao, X.~Yao, and D.~Chen.
\newblock Simcse: Simple contrastive learning of sentence embeddings.
\newblock {\em ArXiv}, 2021.

\bibitem{Gao2023VQPLVQ}
Z.~Gao, C.~Tan, and S.~Z. Li.
\newblock Vqpl: Vector quantized protein language.
\newblock {\em ArXiv}, 2023.

\bibitem{2022MILES}
Y.~Ge, Y.~Ge, X.~Liu, A.~Wang, J.~Wu, Y.~Shan, X.~Qie, and P.~Luo.
\newblock Miles: Visual bert pre-training with injected language semantics for
  video-text retrieval.
\newblock {\em ArXiv}, 2022.

\bibitem{7952261}
J.~F. Gemmeke, D.~P.~W. Ellis, D.~Freedman, A.~Jansen, W.~Lawrence, R.~C.
  Moore, M.~Plakal, and M.~Ritter.
\newblock Audio set: An ontology and human-labeled dataset for audio events.
\newblock In {\em ICASSP}, pages 776--780, 2017.

\bibitem{Geng2023InstructDiffusionAG}
Z.~Geng, B.~Yang, T.~Hang, C.~Li, S.~Gu, T.~Zhang, J.~Bao, Z.~Zhang, H.~Hu,
  D.~Chen, and B.~Guo.
\newblock Instructdiffusion: A generalist modeling interface for vision tasks.
\newblock {\em ArXiv}, 2023.

\bibitem{iccv2023audiovision}
M.-I. Georgescu, E.~Fonseca, R.~T. Ionescu, M.~Lucic, C.~Schmid, and A.~Arnab.
\newblock Audiovisual masked autoencoders.
\newblock In {\em ICCV}, pages 16144--16154, 2023.

\bibitem{2022OmniMAE}
R.~Girdhar, A.~El-Nouby, M.~Singh, K.~V. Alwala, A.~Joulin, and I.~Misra.
\newblock Omnimae: Single model masked pretraining on images and videos.
\newblock {\em ArXiv}, 2022.

\bibitem{grill2020bootstrap}
J.-B. Grill, F.~Strub, F.~Altch{\'e}, C.~Tallec, P.~H. Richemond,
  E.~Buchatskaya, C.~Doersch, B.~A. Pires, Z.~D. Guo, M.~G. Azar, et~al.
\newblock Bootstrap your own latent: A new approach to self-supervised
  learning.
\newblock {\em arXiv preprint}, 2020.

\bibitem{Gu2017AVAAV}
C.~Gu, C.~Sun, S.~Vijayanarasimhan, C.~Pantofaru, D.~A. Ross, G.~Toderici,
  Y.~Li, S.~Ricco, R.~Sukthankar, C.~Schmid, and J.~Malik.
\newblock Ava: A video dataset of spatio-temporally localized atomic visual
  actions.
\newblock In {\em CVPR}, pages 6047--6056, 2017.

\bibitem{Guo2022FastMIM}
J.~Guo, K.~Han, H.~Wu, Y.~Tang, Y.~Wang, and C.~Xu.
\newblock Fastmim: Expediting masked image modeling pre-training for vision.
\newblock 2022.

\bibitem{Gupta2019LVISAD}
A.~Gupta, P.~Doll{\'a}r, and R.~B. Girshick.
\newblock Lvis: A dataset for large vocabulary instance segmentation.
\newblock {\em CVPR}, pages 5351--5359, 2019.

\bibitem{Gupta2022MaskViTMV}
A.~Gupta, S.~Tian, Y.~Zhang, J.~Wu, R.~Mart'in-Mart'in, and L.~Fei-Fei.
\newblock Maskvit: Masked visual pre-training for video prediction.
\newblock {\em ArXiv}, 2022.

\bibitem{Gupta2023SiamMAE}
A.~Gupta, J.~Wu, J.~Deng, and L.~Fei-Fei.
\newblock Siamese masked autoencoders.
\newblock {\em ArXiv}, 2023.

\bibitem{Han2023RevColV2}
Q.~Han, Y.~Cai, and X.~Zhang.
\newblock Revcolv2: Exploring disentangled representations in masked image
  modeling.
\newblock {\em ArXiv}, 2023.

\bibitem{he2022masked}
K.~He, X.~Chen, S.~Xie, Y.~Li, P.~Doll{\'a}r, and R.~Girshick.
\newblock Masked autoencoders are scalable vision learners.
\newblock In {\em CVPR}, pages 16000--16009, 2022.

\bibitem{he2019momentum}
K.~He, H.~Fan, Y.~Wu, S.~Xie, and R.~Girshick.
\newblock Momentum contrast for unsupervised visual representation learning.
\newblock In {\em CVPR}, 2020.

\bibitem{2017iccvmaskrcnn}
K.~He, G.~Gkioxari, P.~Doll{\'a}r, and R.~Girshick.
\newblock Mask r-cnn.
\newblock In {\em ICCV}, 2017.

\bibitem{he2016deep}
K.~He, X.~Zhang, S.~Ren, and J.~Sun.
\newblock Deep residual learning for image recognition.
\newblock In {\em CVPR}, pages 770--778, 2016.

\bibitem{he2021pre}
L.~He, S.~Zhang, L.~Wu, H.~Xia, F.~Ju, H.~Zhang, S.~Liu, Y.~Xia, J.~Zhu,
  P.~Deng, et~al.
\newblock Pre-training co-evolutionary protein representation via a pairwise
  masked language model.
\newblock {\em arXiv preprint}, 2021.

\bibitem{Ho2020DenoisingDP}
J.~Ho, A.~Jain, and P.~Abbeel.
\newblock Denoising diffusion probabilistic models.
\newblock {\em ArXiv}, 2020.

\bibitem{Horn2017TheIC}
G.~V. Horn, O.~M. Aodha, Y.~Song, A.~Shepard, H.~Adam, P.~Perona, and S.~J.
  Belongie.
\newblock The inaturalist challenge 2017 dataset.
\newblock {\em ArXiv}, 2017.

\bibitem{hou2023graphmae2}
Z.~Hou, Y.~He, Y.~Cen, X.~Liu, Y.~Dong, E.~Kharlamov, and J.~Tang.
\newblock Graphmae2: A decoding-enhanced masked self-supervised graph learner.
\newblock In {\em WWW}, pages 737--746, 2023.

\bibitem{hou2022graphmae}
Z.~Hou, X.~Liu, Y.~Cen, Y.~Dong, H.~Yang, C.~Wang, and J.~Tang.
\newblock Graphmae: Self-supervised masked graph autoencoders.
\newblock In {\em SIGKDD}, pages 594--604, 2022.

\bibitem{Hou2022MILAN}
Z.~Hou, F.~Sun, Y.-K. Chen, Y.~Xie, and S.~Y. Kung.
\newblock Milan: Masked image pretraining on language assisted representation.
\newblock {\em ArXiv}, 2022.

\bibitem{taslp2021hubert}
W.-N. Hsu, B.~Bolte, Y.-H.~H. Tsai, K.~Lakhotia, R.~Salakhutdinov, and
  A.~Mohamed.
\newblock Hubert: Self-supervised speech representation learning by masked
  prediction of hidden units.
\newblock {\em TASLP}, 29:3451--3460, 2021.

\bibitem{hu2022exploring}
M.~Hu, F.~Yuan, K.~K. Yang, F.~Ju, J.~Su, H.~Wang, F.~Yang, and Q.~Ding.
\newblock Exploring evolution-based \&-free protein language models as protein
  function predictors.
\newblock {\em arXiv preprint}, 2022.

\bibitem{hu2019strategies}
W.~Hu, B.~Liu, J.~Gomes, M.~Zitnik, P.~Liang, V.~Pande, and J.~Leskovec.
\newblock Strategies for pre-training graph neural networks.
\newblock In {\em ICLR}, 2019.

\bibitem{hu2020gpt}
Z.~Hu, Y.~Dong, K.~Wang, K.-W. Chang, and Y.~Sun.
\newblock Gpt-gnn: Generative pre-training of graph neural networks.
\newblock In {\em SIGKDD}, pages 1857--1867, 2020.

\bibitem{Hua2022SelfsupervisionTR}
T.~Hua, Y.~Tian, S.~Ren, H.~Zhao, and L.~Sigal.
\newblock Self-supervision through random segments with autoregressive coding
  (randsac).
\newblock {\em ArXiv}, 2022.

\bibitem{Huang2023MGMAEMG}
B.~Huang, Z.~Zhao, G.~Zhang, Y.~Qiao, and L.~Wang.
\newblock Mgmae: Motion guided masking for video masked autoencoding.
\newblock {\em ArXiv}, 2023.

\bibitem{Huang2022GreenHV}
L.~Huang, S.~You, M.~Zheng, F.~Wang, C.~Qian, and T.~Yamasaki.
\newblock Green hierarchical vision transformer for masked image modeling.
\newblock {\em ArXiv}, 2022.

\bibitem{huang2022masked}
P.-Y. Huang, H.~Xu, J.~Li, A.~Baevski, M.~Auli, W.~Galuba, F.~Metze, and
  C.~Feichtenhofer.
\newblock Masked autoencoders that listen.
\newblock {\em NeurIPS}, 35:28708--28720, 2022.

\bibitem{nips2022AudioMAE}
P.-Y. Huang, H.~Xu, J.~B. Li, A.~Baevski, M.~Auli, W.~Galuba, F.~Metze, and
  C.~Feichtenhofer.
\newblock Masked autoencoders that listen.
\newblock {\em ArXiv}, 2022.

\bibitem{Huang2023ImprovingAR}
Q.~Huang, X.~Dong, D.~Chen, Y.~Chen, L.~Yuan, G.~Hua, W.~Zhang, N.~H. Yu, and
  M.~Reaserch.
\newblock Improving adversarial robustness of masked autoencoders via test-time
  frequency-domain prompting.
\newblock {\em ArXiv}, 2023.

\bibitem{cvpr2023G2SD}
W.~Huang, Z.~Peng, L.~Dong, F.~Wei, J.~Jiao, and Q.~Ye.
\newblock Generic-to-specific distillation of masked autoencoders.
\newblock {\em ArXiv}, 2023.

\bibitem{2022CMAE}
Z.~Huang, X.~Jin, C.~Lu, Q.~Hou, M.-M. Cheng, D.~Fu, X.~Shen, and J.~Feng.
\newblock Contrastive masked autoencoders are stronger vision learners.
\newblock {\em ArXiv}, 2022.

\bibitem{iclr2023layergrafted}
Z.~Jiang, Y.~Chen, M.~Liu, D.~Chen, X.~Dai, L.~Yuan, Z.~Liu, and Z.~Wang.
\newblock Layer grafted pre-training: Bridging contrastive learning and masked
  image modeling for label-efficient representations.
\newblock In {\em ICLR}, 2023.

\bibitem{Jing2022MaskedSC}
L.~Jing, J.~Zhu, and Y.~LeCun.
\newblock Masked siamese convnets.
\newblock {\em ArXiv}, 2022.

\bibitem{Mao2022SDMAE}
J.~ju~Mao, H.~Zhou, X.~Yin, Y.~Chang, B.~Nie, and R.~Xu.
\newblock Masked autoencoders are effective solution to transformer
  data-hungry.
\newblock {\em ArXiv}, 2022.

\bibitem{jumper2021highly}
J.~Jumper, R.~Evans, A.~Pritzel, T.~Green, M.~Figurnov, O.~Ronneberger,
  K.~Tunyasuvunakool, R.~Bates, A.~{\v{Z}}{\'\i}dek, A.~Potapenko, et~al.
\newblock Highly accurate protein structure prediction with alphafold.
\newblock {\em Nature}, 596(7873):583--589, 2021.

\bibitem{eccv2022attmask}
I.~Kakogeorgiou, S.~Gidaris, B.~Psomas, Y.~Avrithis, A.~Bursuc, K.~Karantzalos,
  and N.~Komodakis.
\newblock What to hide from your students: Attention-guided masked image
  modeling.
\newblock In {\em ECCV}, 2022.

\bibitem{Kay2017TheKH}
W.~Kay, J.~Carreira, K.~Simonyan, B.~Zhang, C.~Hillier, S.~Vijayanarasimhan,
  F.~Viola, T.~Green, T.~Back, A.~Natsev, M.~Suleyman, and A.~Zisserman.
\newblock The kinetics human action video dataset.
\newblock {\em ArXiv}, 2017.

\bibitem{Khan2023MeSaMG}
M.~O. Khan, J.~Liang, C.-K. Wang, S.~Yang, and Y.~Lou.
\newblock Mesa: Masked, geometric, and supervised pre-training for monocular
  depth estimation.
\newblock {\em ArXiv}, 2023.

\bibitem{cvpr2023understandingMAE}
L.~Kong, M.~Q. Ma, G.~Chen, E.~P. Xing, Y.~Chi, L.-P. Morency, and K.~Zhang.
\newblock Understanding masked autoencoders via hierarchical latent variable
  models.
\newblock In {\em CVPR}, pages 7918--7928, 2023.

\bibitem{2022UnderstandMIM}
X.~Kong and X.~Zhang.
\newblock Understanding masked image modeling via learning occlusion invariant
  feature.
\newblock {\em ArXiv}, 2022.

\bibitem{6755945}
J.~Krause, M.~Stark, J.~Deng, and L.~Fei-Fei.
\newblock 3d object representations for fine-grained categorization.
\newblock In {\em ICCV workshops}, pages 554--561, 2013.

\bibitem{Krizhevsky2009LearningML}
A.~Krizhevsky.
\newblock Learning multiple layers of features from tiny images.
\newblock 2009.

\bibitem{2022AnES}
G.~K. Kumar, S.~S. Mullappilly, and A.~S. Gehlot.
\newblock An empirical study of self-supervised learning approaches for object
  detection with transformers.
\newblock {\em ArXiv}, 2022.

\bibitem{Kwon2022MaskedVA}
G.~Kwon, Z.~Cai, A.~Ravichandran, E.~Bas, R.~Bhotika, and S.~. Soatto.
\newblock Masked vision and language modeling for multi-modal representation
  learning.
\newblock In {\em ICLR}, 2023.

\bibitem{Lai2019ContrastivePC}
C.-I. Lai.
\newblock Contrastive predictive coding based feature for automatic speaker
  verification.
\newblock {\em ArXiv}, 2019.

\bibitem{Lai2017RACELR}
G.~Lai, Q.~Xie, H.~Liu, Y.~Yang, and E.~H. Hovy.
\newblock Race: Large-scale reading comprehension dataset from examinations.
\newblock {\em ArXiv}, 2017.

\bibitem{lan2019albert}
Z.~Lan, M.~Chen, S.~Goodman, K.~Gimpel, P.~Sharma, and R.~Soricut.
\newblock Albert: A lite bert for self-supervised learning of language
  representations.
\newblock {\em arXiv preprint}, 2019.

\bibitem{iccv2023maekd}
S.~Lao, G.~Song, B.~Liu, Y.~Liu, and Y.~Yang.
\newblock Masked autoencoders are stronger knowledge distillers.
\newblock In {\em ICCV}, pages 6384--6393, 2023.

\bibitem{Lee2021VisionTF}
S.~H. Lee, S.~Lee, and B.~C. Song.
\newblock Vision transformer for small-size datasets.
\newblock {\em ArXiv}, 2021.

\bibitem{Lee2022RCMAE}
Y.~Lee, J.~Willette, J.~Kim, J.~Lee, and S.~J. Hwang.
\newblock Exploring the role of mean teachers in self-supervised masked
  auto-encoders.
\newblock {\em ArXiv}, 2022.

\bibitem{Lehner2023ContrastiveTA}
J.~Lehner, B.~Alkin, A.~F{\"u}rst, E.~Rumetshofer, L.~Miklautz, and
  S.~Hochreiter.
\newblock Contrastive tuning: A little help to make masked autoencoders forget.
\newblock {\em ArXiv}, 2023.

\bibitem{Li2023DreamTeacherPI}
D.~Li, H.~Ling, A.~Kar, D.~Acuna, S.~W. Kim, K.~Kreis, A.~Torralba, and
  S.~Fidler.
\newblock Dreamteacher: Pretraining image backbones with deep generative
  models.
\newblock {\em ArXiv}, 2023.

\bibitem{2022SemMAE}
G.~Li, H.~Zheng, D.~Liu, B.~Su, and C.~Zheng.
\newblock Semmae: Semantic-guided masking for learning masked autoencoders.
\newblock {\em ArXiv}, 2022.

\bibitem{Li2020PrototypicalCL}
J.~Li, P.~Zhou, C.~Xiong, R.~Socher, and S.~C.~H. Hoi.
\newblock Prototypical contrastive learning of unsupervised representations.
\newblock {\em ArXiv}, 2020.

\bibitem{Li2022MogaNet}
S.~Li, Z.~Wang, Z.~Liu, C.~Tan, H.~Lin, D.~Wu, Z.~Chen, J.~Zheng, and S.~Z. Li.
\newblock Efficient multi-order gated aggregation network.
\newblock {\em ArXiv}, 2022.

\bibitem{2022a2mim}
S.~Li, D.~Wu, F.~Wu, Z.~Zang, K.~Wang, L.~Shang, B.~Sun, H.~Li, and Stan.Z.Li.
\newblock Architecture-agnostic masked image modeling - from vit back to cnn.
\newblock In {\em ICML}, 2023.

\bibitem{cvpr2023mage}
T.~Li, H.~Chang, S.~K. Mishra, H.~Zhang, D.~Katabi, and D.~Krishnan.
\newblock Mage: Masked generative encoder to unify representation learning and
  image synthesis.
\newblock {\em arXiv preprint}, 2022.

\bibitem{Li2023SelfconditionedIG}
T.~Li, D.~Katabi, and K.~He.
\newblock Self-conditioned image generation via generating representations.
\newblock 2023.

\bibitem{Li2022mcBEiTMD}
X.~Li, Y.~Ge, K.~Yi, Z.~Hu, Y.~Shan, and L.~yu~Duan.
\newblock mc-beit: Multi-choice discretization for image bert pre-training.
\newblock In {\em ECCV}, 2022.

\bibitem{Li2022UniformME}
X.~Li, W.~Wang, L.~Yang, and J.~Yang.
\newblock Uniform masking: Enabling mae pre-training for pyramid-based vision
  transformers with locality.
\newblock {\em ArXiv}, 2022.

\bibitem{Li2022FLIP}
Y.~Li, H.~Fan, R.~Hu, C.~Feichtenhofer, and K.~He.
\newblock Scaling language-image pre-training via masking.
\newblock {\em ArXiv}, 2022.

\bibitem{Li2021MSTMS}
Z.~Li, Z.~Chen, F.~Yang, W.~Li, Y.~Zhu, C.~Zhao, R.~Deng, L.~Wu, R.~Zhao,
  M.~Tang, and J.~Wang.
\newblock Mst: Masked self-supervised transformer for visual representation.
\newblock In {\em NeurIPS}, 2021.

\bibitem{eccv2022MeshMAE}
Y.~Liang, S.~Zhao, B.~Yu, J.~Zhang, and F.~He.
\newblock Meshmae: Masked autoencoders for 3d mesh data analysis.
\newblock In {\em ECCV}, 2022.

\bibitem{Liao2021KITTI360AN}
Y.~Liao, J.~Xie, and A.~Geiger.
\newblock Kitti-360: A novel dataset and benchmarks for urban scene
  understanding in 2d and 3d.
\newblock {\em TPAMI}, pages 3292--3310, 2021.

\bibitem{Lin2014MicrosoftCC}
T.-Y. Lin, M.~Maire, S.~J. Belongie, J.~Hays, P.~Perona, D.~Ramanan,
  P.~Doll{\'a}r, and C.~L. Zitnick.
\newblock Microsoft coco: Common objects in context.
\newblock In {\em ECCV}, 2014.

\bibitem{liu2020mockingjay}
A.~T. Liu, S.-w. Yang, P.-H. Chi, P.-c. Hsu, and H.-y. Lee.
\newblock Mockingjay: Unsupervised speech representation learning with deep
  bidirectional transformer encoders.
\newblock In {\em ICASSP}. IEEE, 2020.

\bibitem{Liu2022MaskedPT}
B.~Liu, D.~Hsu, P.~Ravikumar, and A.~Risteski.
\newblock Masked prediction tasks: a parameter identifiability view.
\newblock {\em ArXiv}, 2022.

\bibitem{Liu2022MaskedDF}
H.~Liu, M.~Cai, and Y.~J. Lee.
\newblock Masked discrimination for self-supervised learning on point clouds.
\newblock In {\em ECCV}, 2022.

\bibitem{Liu2022TheDI}
H.~Liu, X.~Jiang, X.~Li, A.~Guo, D.~Jiang, and B.~Ren.
\newblock The devil is in the frequency: Geminated gestalt autoencoder for
  self-supervised visual pre-training.
\newblock In {\em AAAI}, 2022.

\bibitem{Liu2023LanguageQA}
H.~Liu, W.~Yan, and P.~Abbeel.
\newblock Language quantized autoencoders: Towards unsupervised text-image
  alignment.
\newblock {\em ArXiv}, 2023.

\bibitem{2022MixMIM}
J.~Liu, X.~Huang, Y.~Liu, and H.~Li.
\newblock Mixmim: Mixed and masked image modeling for efficient visual
  representation learning.
\newblock {\em ArXiv}, 2022.

\bibitem{Liu2023TowardsB3}
J.~Liu, T.~Wang, B.~Liu, Q.~Zhang, Y.~Liu, and H.~Li.
\newblock Towards better 3d knowledge transfer via masked image modeling for
  multi-view 3d understanding.
\newblock {\em ArXiv}, 2023.

\bibitem{icme2023DocMAE}
S.~Liu, H.~Feng, W.~gang Zhou, H.~Li, C.~Liu, and F.~Wu.
\newblock Docmae: Document image rectification via self-supervised
  representation learning.
\newblock 2023.

\bibitem{liu2021self}
X.~Liu, F.~Zhang, Z.~Hou, L.~Mian, Z.~Wang, J.~Zhang, and J.~Tang.
\newblock Self-supervised learning: Generative or contrastive.
\newblock {\em TKDE}, 35(1):857--876, 2021.

\bibitem{liu2022dBOT}
X.~Liu, J.~Zhou, T.~Kong, X.~Lin, and R.~Ji.
\newblock Exploring target representations for masked autoencoders.
\newblock {\em arXiv preprint}, 2022.

\bibitem{liu2019roberta}
Y.~Liu, M.~Ott, N.~Goyal, J.~Du, M.~Joshi, D.~Chen, O.~Levy, M.~Lewis,
  L.~Zettlemoyer, and V.~Stoyanov.
\newblock Roberta: A robustly optimized bert pretraining approach.
\newblock {\em arXiv preprint}, 2019.

\bibitem{Liu2023PixMIMRP}
Y.~Liu, S.~Zhang, J.~Chen, K.~Chen, and D.~Lin.
\newblock Pixmim: Rethinking pixel reconstruction in masked image modeling.
\newblock {\em ArXiv}, 2023.

\bibitem{Liu2023ImprovingPM}
Y.~Liu, S.~Zhang, J.~Chen, Z.~Yu, K.~Chen, and D.~Lin.
\newblock Improving pixel-based mim by reducing wasted modeling capability.
\newblock {\em ArXiv}, 2023.

\bibitem{2022convnet}
Z.~Liu, H.~Mao, C.-Y. Wu, C.~Feichtenhofer, T.~Darrell, and S.~Xie.
\newblock A convnet for the 2020s.
\newblock In {\em CVPR}, 2022.

\bibitem{lu2020self}
A.~X. Lu, H.~Zhang, M.~Ghassemi, and A.~Moses.
\newblock Self-supervised contrastive learning of protein representations by
  mutual information maximization.
\newblock {\em BioRxiv}, 2020.

\bibitem{Lu2023CMAEVCM}
C.~Lu, X.~Jin, Z.~Huang, Q.~Hou, M.-M. Cheng, and J.~Feng.
\newblock Cmae-v: Contrastive masked autoencoders for video action recognition.
\newblock {\em ArXiv}, 2023.

\bibitem{lu2022UnifiedIO}
J.~Lu, C.~Clark, R.~Zellers, R.~Mottaghi, and A.~Kembhavi.
\newblock Unified-io: A unified model for vision, language, and multi-modal
  tasks.
\newblock {\em arXiv preprint}, 2022.

\bibitem{Lugmayr2022RePaintIU}
A.~Lugmayr, M.~Danelljan, A.~Romero, F.~Yu, R.~Timofte, and L.~V. Gool.
\newblock Repaint: Inpainting using denoising diffusion probabilistic models.
\newblock {\em CVPR}, pages 11451--11461, 2022.

\bibitem{Luo2022SelfdistillationAM}
Y.~Luo, Z.~Chen, and X.~Gao.
\newblock Self-distillation augmented masked autoencoders for histopathological
  image classification.
\newblock {\em ArXiv}, 2022.

\bibitem{2022MaskOCR}
P.~Lyu, C.~Zhang, S.~Liu, M.~Qiao, Y.~Xu, L.~Wu, K.~Yao, J.~Han, E.~Ding, and
  J.~Wang.
\newblock Maskocr: Text recognition with masked encoder-decoder pretraining.
\newblock {\em ArXiv}, 2022.

\bibitem{Ma2022DisjointMW}
X.~Ma, C.-S. Liu, C.~Xie, L.~Ye, Y.~Deng, and X.~Ji.
\newblock Disjoint masking with joint distillation for efficient masked image
  modeling.
\newblock {\em ArXiv}, 2022.

\bibitem{Maji2013FineGrainedVC}
S.~Maji, E.~Rahtu, J.~Kannala, M.~B. Blaschko, and A.~Vedaldi.
\newblock Fine-grained visual classification of aircraft.
\newblock {\em ArXiv}, 2013.

\bibitem{Mao2023MaskedMP}
Y.~Mao, J.~Deng, W.~gang Zhou, Y.~Fang, W.~Ouyang, and H.~Li.
\newblock Masked motion predictors are strong 3d action representation
  learners.
\newblock {\em ArXiv}, 2023.

\bibitem{mcdermott2021adversarial}
M.~McDermott, B.~Yap, H.~Hsu, D.~Jin, and P.~Szolovits.
\newblock Adversarial contrastive pre-training for protein sequences.
\newblock {\em arXiv}, 2021.

\bibitem{Miech2020RareActAV}
A.~Miech, J.-B. Alayrac, I.~Laptev, J.~Sivic, and A.~Zisserman.
\newblock Rareact: A video dataset of unusual interactions.
\newblock {\em ArXiv}, 2020.

\bibitem{2022VoxelMAE}
C.~Min, X.~Xu, D.~Zhao, L.~Xiao, Y.~Nie, and B.~Dai.
\newblock Voxel-mae: Masked autoencoders for pre-training large-scale point
  clouds.
\newblock {\em ArXiv}, 2022.

\bibitem{min2022transformer}
E.~Min, R.~Chen, Y.~Bian, T.~Xu, K.~Zhao, W.~Huang, P.~Zhao, J.~Huang,
  S.~Ananiadou, and Y.~Rong.
\newblock Transformer for graphs: An overview from architecture perspective.
\newblock {\em arXiv preprint}, 2022.

\bibitem{Mishra2022ASE}
S.~K. Mishra, J.~Robinson, H.~Chang, D.~Jacobs, A.~Sarna, A.~Maschinot, and
  D.~Krishnan.
\newblock A simple, efficient and scalable contrastive masked autoencoder for
  learning visual representations.
\newblock {\em ArXiv}, 2022.

\bibitem{8014984}
S.~Moschoglou, A.~Papaioannou, C.~Sagonas, J.~Deng, I.~Kotsia, and
  S.~Zafeiriou.
\newblock Agedb: The first manually collected, in-the-wild age database.
\newblock In {\em CVPRW}, pages 1997--2005, 2017.

\bibitem{TGRS2023CMID}
D.~Muhtar, X.~liang Zhang, P.~Xiao, Z.~Li, and F.~Gu.
\newblock Cmid: A unified self-supervised learning framework for remote sensing
  image understanding.
\newblock {\em TGRS}, 2023.

\bibitem{Nguyen2023RMAERM}
D.-K. Nguyen, V.~Aggarwal, Y.~Li, M.~R. Oswald, A.~Kirillov, C.~G.~M. Snoek,
  and X.~Chen.
\newblock R-mae: Regions meet masked autoencoders.
\newblock {\em ArXiv}, 2023.

\bibitem{Nilsback08}
M.-E. Nilsback and A.~Zisserman.
\newblock Automated flower classification over a large number of classes.
\newblock In {\em ICVGIP}, Dec 2008.

\bibitem{Noroozi2016UnsupervisedLO}
M.~Noroozi and P.~Favaro.
\newblock Unsupervised learning of visual representations by solving jigsaw
  puzzles.
\newblock {\em ArXiv}, 2016.

\bibitem{oord2018representation}
A.~v.~d. Oord, Y.~Li, and O.~Vinyals.
\newblock Representation learning with contrastive predictive coding.
\newblock {\em arXiv preprint}, 2018.

\bibitem{pan2023img2vec}
H.~Pan, C.~Liu, W.~Wang, L.~Yuan, H.~Wang, Z.~Li, and W.~Liu.
\newblock Img2vec: A teacher of high token-diversity helps masked autoencoders,
  2023.

\bibitem{7178964}
V.~Panayotov, G.~Chen, D.~Povey, and S.~Khudanpur.
\newblock Librispeech: An asr corpus based on public domain audio books.
\newblock In {\em ICASSP}, pages 5206--5210, 2015.

\bibitem{Pang2022MaskedAF}
Y.~Pang, W.~Wang, F.~E.~H. Tay, W.~Liu, Y.~Tian, and L.~Yuan.
\newblock Masked autoencoders for point cloud self-supervised learning.
\newblock In {\em ECCV}, 2022.

\bibitem{2022BEiTV2}
Z.~Peng, L.~Dong, H.~Bao, Q.~Ye, and F.~Wei.
\newblock Beit v2: Masked image modeling with vector-quantized visual
  tokenizers.
\newblock {\em ArXiv}, 2022.

\bibitem{2022maskdistill}
Z.~Peng, L.~Dong, H.~Bao, Q.~Ye, and F.~Wei.
\newblock A unified view of masked image modeling.
\newblock 2022.

\bibitem{Qi2023ContrastWR}
Z.~Qi, R.~Dong, G.~Fan, Z.~Ge, X.~Zhang, K.~Ma, and L.~Yi.
\newblock Contrast with reconstruct: Contrastive 3d representation learning
  guided by generative pretraining.
\newblock {\em ArXiv}, 2023.

\bibitem{2022MAR}
Z.~Qing, S.~Zhang, Z.~Huang, X.~Wang, Y.~Wang, Y.~Lv, C.~Gao, and N.~Sang.
\newblock Mar: Masked autoencoders for efficient action recognition.
\newblock {\em ArXiv}, 2022.

\bibitem{2022gcmae}
H.~Quan, X.~Li, W.~Chen, Q.~Bai, M.~Zou, R.~Yang, T.~Zheng, R.~Qi, X.~Gao, and
  X.~Cui.
\newblock Global contrast masked autoencoders are powerful pathological
  representation learners.
\newblock {\em arXiv}, 2022.

\bibitem{radford2021learning}
A.~Radford, J.~W. Kim, C.~Hallacy, A.~Ramesh, G.~Goh, S.~Agarwal, G.~Sastry,
  A.~Askell, P.~Mishkin, J.~Clark, G.~Krueger, and I.~Sutskever.
\newblock Learning transferable visual models from natural language
  supervision.
\newblock In {\em ICML}, 2021.

\bibitem{Ramesh2022HierarchicalTI}
A.~Ramesh, P.~Dhariwal, A.~Nichol, C.~Chu, and M.~Chen.
\newblock Hierarchical text-conditional image generation with clip latents.
\newblock {\em ArXiv}, 2022.

\bibitem{Ramesh2021ZeroShotTG}
A.~Ramesh, M.~Pavlov, G.~Goh, S.~Gray, C.~Voss, A.~Radford, M.~Chen, and
  I.~Sutskever.
\newblock Zero-shot text-to-image generation.
\newblock {\em ArXiv}, 2021.

\bibitem{rao2019evaluating}
R.~Rao, N.~Bhattacharya, N.~Thomas, Y.~Duan, P.~Chen, J.~Canny, P.~Abbeel, and
  Y.~Song.
\newblock Evaluating protein transfer learning with tape.
\newblock {\em NeurIPS}, 2019.

\bibitem{Reed2022ScaleMAEAS}
C.~Reed, R.~Gupta, S.~Li, S.~Brockman, C.~Funk, B.~Clipp, S.~Candido,
  M.~Uyttendaele, and T.~Darrell.
\newblock Scale-mae: A scale-aware masked autoencoder for multiscale geospatial
  representation learning.
\newblock {\em ArXiv}, 2022.

\bibitem{ren2023rejuvenating}
S.~Ren, Z.~Wang, H.~Zhu, J.~Xiao, A.~Yuille, and C.~Xie.
\newblock Rejuvenating image-gpt as strong visual representation learners,
  2023.

\bibitem{Ren2023TinyMIM}
S.~Ren, F.~Wei, Z.~Zhang, and H.~Hu.
\newblock Tinymim: An empirical study of distilling mim pre-trained models.
\newblock 2023.

\bibitem{rives2021biological}
A.~Rives, J.~Meier, T.~Sercu, S.~Goyal, Z.~Lin, J.~Liu, D.~Guo, M.~Ott, C.~L.
  Zitnick, J.~Ma, et~al.
\newblock Biological structure and function emerge from scaling unsupervised
  learning to 250 million protein sequences.
\newblock {\em NAS}, 118(15):e2016239118, 2021.

\bibitem{Rohrbach2014CoherentMV}
A.~Rohrbach, M.~Rohrbach, W.~Qiu, A.~Friedrich, M.~Pinkal, and B.~Schiele.
\newblock Coherent multi-sentence video description with variable level of
  detail.
\newblock In {\em GCPR}, 2014.

\bibitem{Rohrbach2015ADF}
A.~Rohrbach, M.~Rohrbach, N.~Tandon, and B.~Schiele.
\newblock A dataset for movie description.
\newblock In {\em CVPR}, pages 3202--3212, 2015.

\bibitem{Rombach2021HighResolutionIS}
R.~Rombach, A.~Blattmann, D.~Lorenz, P.~Esser, and B.~Ommer.
\newblock High-resolution image synthesis with latent diffusion models.
\newblock {\em CVPR}, pages 10674--10685, 2021.

\bibitem{rong2020self}
Y.~Rong, Y.~Bian, T.~Xu, W.~Xie, Y.~Wei, W.~Huang, and J.~Huang.
\newblock Self-supervised graph transformer on large-scale molecular data.
\newblock {\em NeurIPS}, 2020.

\bibitem{ross2022large}
J.~Ross, B.~Belgodere, V.~Chenthamarakshan, I.~Padhi, Y.~Mroueh, and P.~Das.
\newblock Large-scale chemical language representations capture molecular
  structure and properties.
\newblock {\em Nature Machine Intelligence}, 4(12):1256--1264, 2022.

\bibitem{Russakovsky2014ImageNetLS}
O.~Russakovsky, J.~Deng, H.~Su, J.~Krause, S.~Satheesh, S.~Ma, Z.~Huang,
  A.~Karpathy, A.~Khosla, M.~S. Bernstein, A.~C. Berg, and L.~Fei-Fei.
\newblock Imagenet large scale visual recognition challenge.
\newblock {\em IJCV}, pages 211--252, 2014.

\bibitem{icml2023Hiera}
C.~K. Ryali, Y.-T. Hu, D.~Bolya, C.~Wei, H.~Fan, P.-Y. Huang, V.~Aggarwal,
  A.~Chowdhury, O.~Poursaeed, J.~Hoffman, J.~Malik, Y.~Li, and
  C.~Feichtenhofer.
\newblock Hiera: A hierarchical vision transformer without the
  bells-and-whistles.
\newblock In {\em ICML}, 2023.

\bibitem{Sameni2022RepresentationLB}
S.~Sameni, S.~Jenni, and P.~Favaro.
\newblock Representation learning by detecting incorrect location embeddings.
\newblock {\em AAAI}, 2022.

\bibitem{Schuhmann2022LAION5BAO}
C.~Schuhmann, R.~Beaumont, R.~Vencu, C.~Gordon, R.~Wightman, M.~Cherti,
  T.~Coombes, A.~Katta, C.~Mullis, M.~Wortsman, P.~Schramowski, S.~Kundurthy,
  K.~Crowson, L.~Schmidt, R.~Kaczmarczyk, and J.~Jitsev.
\newblock Laion-5b: An open large-scale dataset for training next generation
  image-text models.
\newblock {\em ArXiv}, 2022.

\bibitem{cvpr2023PointCMP}
Z.~Shen, X.~Sheng, L.~Wang, Y.~K. Guo, Q.~Liu, and X.~Zhou.
\newblock Pointcmp: Contrastive mask prediction for self-supervised learning on
  point cloud videos.
\newblock In {\em CVPR}, 2023.

\bibitem{shi2022adversarial}
Y.~Shi, N.~Siddharth, P.~Torr, and A.~R. Kosiorek.
\newblock Adversarial masking for self-supervised learning.
\newblock In {\em ICML}, 2022.

\bibitem{Sigurdsson2016HollywoodIH}
G.~A. Sigurdsson, G.~Varol, X.~Wang, A.~Farhadi, I.~Laptev, and A.~K. Gupta.
\newblock Hollywood in homes: Crowdsourcing data collection for activity
  understanding.
\newblock In {\em ECCV}, 2016.

\bibitem{Singh2023WSP}
M.~Singh, Q.~Duval, K.~V. Alwala, H.~Fan, V.~Aggarwal, A.~B. Adcock, A.~Joulin,
  P.~Doll'ar, C.~Feichtenhofer, R.~B. Girshick, R.~Girdhar, and I.~Misra.
\newblock The effectiveness of mae pre-pretraining for billion-scale
  pretraining.
\newblock {\em ArXiv}, 2023.

\bibitem{2022MAM2}
Y.~Song, M.~Yang, W.~Wu, D.~He, F.~Li, and J.~Wang.
\newblock It takes two: Masked appearance-motion modeling for self-supervised
  video transformer pre-training.
\newblock {\em ArXiv}, 2022.

\bibitem{Soomro2012UCF101AD}
K.~Soomro, A.~R. Zamir, and M.~Shah.
\newblock Ucf101: A dataset of 101 human actions classes from videos in the
  wild.
\newblock {\em ArXiv}, 2012.

\bibitem{su2023saprot}
J.~Su, C.~Han, Y.~Zhou, J.~Shan, X.~Zhou, and F.~Yuan.
\newblock Saprot: Protein language modeling with structure-aware vocabulary.
\newblock {\em bioRxiv}, 2023.

\bibitem{Su2019VLBERTPO}
W.~Su, X.~Zhu, Y.~Cao, B.~Li, L.~Lu, F.~Wei, and J.~Dai.
\newblock Vl-bert: Pre-training of generic visual-linguistic representations.
\newblock {\em ArXiv}, 2019.

\bibitem{tan2022mgae}
Q.~Tan, N.~Liu, X.~Huang, R.~Chen, S.-H. Choi, and X.~Hu.
\newblock Mgae: Masked autoencoders for self-supervised learning on graphs.
\newblock {\em arXiv preprint}, 2022.

\bibitem{Tao2022SiameseIM}
C.~Tao, X.~Zhu, G.~Huang, Y.~Qiao, X.~Wang, and J.~Dai.
\newblock Siamese image modeling for self-supervised vision representation
  learning.
\newblock {\em CVPR}, pages 2132--2141, 2022.

\bibitem{Tian2023SparK}
K.~Tian, Y.~Jiang, Q.~Diao, C.~Lin, L.~Wang, and Z.~Yuan.
\newblock Designing bert for convolutional networks: Sparse and hierarchical
  masked modeling.
\newblock {\em ArXiv}, 2023.

\bibitem{cvpr2023GeoMAE}
X.~Tian, H.~Ran, Y.~Wang, and H.~Zhao.
\newblock Geomae: Masked geometric target prediction for self-supervised point
  cloud pre-training.
\newblock In {\em CVPR}, pages 13570--13580, 2023.

\bibitem{Tian2022IntegrallyPT}
Y.~Tian, L.~Xie, Z.~Wang, L.~Wei, X.~Zhang, J.~Jiao, Y.~Wang, Q.~Tian, and
  Q.~Ye.
\newblock Integrally pre-trained transformer pyramid networks.
\newblock {\em CVPR}, pages 18610--18620, 2022.

\bibitem{Tong2022VideoMAEMA}
Z.~Tong, Y.~Song, J.~Wang, and L.~Wang.
\newblock Videomae: Masked autoencoders are data-efficient learners for
  self-supervised video pre-training.
\newblock {\em ArXiv}, 2022.

\bibitem{van2016conditional}
A.~Van~den Oord, N.~Kalchbrenner, L.~Espeholt, O.~Vinyals, A.~Graves, et~al.
\newblock Conditional image generation with pixelcnn decoders.
\newblock In {\em NeurIPS}, pages 4790--4798, 2016.

\bibitem{Oord2017NeuralDR}
A.~van~den Oord, O.~Vinyals, and K.~Kavukcuoglu.
\newblock Neural discrete representation learning.
\newblock {\em ArXiv}, 2017.

\bibitem{nature2023foldseek}
M.~van Kempen, S.~S. Kim, C.~Tumescheit, M.~Mirdita, J.~S{\"o}ding, and
  M.~Steinegger.
\newblock Foldseek: fast and accurate protein structure search.
\newblock 2022.

\bibitem{vaswani2017attention}
A.~Vaswani, N.~Shazeer, N.~Parmar, J.~Uszkoreit, L.~Jones, A.~N. Gomez,
  {\L}.~Kaiser, and I.~Polosukhin.
\newblock Attention is all you need.
\newblock In {\em NeurIPS}, pages 5998--6008, 2017.

\bibitem{wah2011caltech}
C.~Wah, S.~Branson, P.~Welinder, P.~Perona, and S.~Belongie.
\newblock The caltech-ucsd birds-200-2011 dataset.
\newblock 2011.

\bibitem{Wang2023DropPosPV}
H.~Wang, J.~Fan, Y.~Wang, K.~Song, T.~Wang, and Z.~Zhang.
\newblock Droppos: Pre-training vision transformers by reconstructing dropped
  positions.
\newblock {\em ArXiv}, 2023.

\bibitem{cvpr2023HPM}
H.~Wang, K.~Song, J.~Fan, Y.~Wang, J.~Xie, and Z.~Zhang.
\newblock Hard patches mining for masked image modeling.
\newblock In {\em CVPR}, 2023.

\bibitem{Wang2023MaskedIM}
H.~Wang, Y.~Tang, Y.~Wang, J.~Guo, Z.~Deng, and K.~Han.
\newblock Masked image modeling with local multi-scale reconstruction.
\newblock {\em CVPR}, pages 2122--2131, 2023.

\bibitem{Wang2022FaceMAEPF}
K.~Wang, B.~Zhao, X.~Peng, Z.~H. Zhu, J.~Deng, X.~Wang, H.~Bilen, and Y.~You.
\newblock Facemae: Privacy-preserving face recognition via masked autoencoders.
\newblock {\em ArXiv}, 2022.

\bibitem{cvpr2023VideoMAEV2}
L.~Wang, B.~Huang, Z.~Zhao, Z.~Tong, Y.~He, Y.~Wang, Y.~Wang, and Y.~Qiao.
\newblock Videomae v2: Scaling video masked autoencoders with dual masking.
\newblock In {\em CVPR}, 2023.

\bibitem{Wang2022RePreIS}
L.~Wang, F.~Liang, Y.~Li, H.~Zhang, W.~Ouyang, and J.~Shao.
\newblock Repre: Improving self-supervised vision transformer with
  reconstructive pre-training.
\newblock In {\em IJCAI}, 2022.

\bibitem{Wang2022OracleMNISTAR}
M.~Wang and W.~Deng.
\newblock Oracle-mnist: a realistic image dataset for benchmarking machine
  learning algorithms.
\newblock {\em ArXiv}, 2022.

\bibitem{cvpr2022BEVT}
R.~Wang, D.~Chen, Z.~Wu, Y.~Chen, X.~Dai, M.~Liu, Y.-G. Jiang, L.~Zhou, and
  L.~Yuan.
\newblock Bevt: Bert pretraining of video transformers.
\newblock In {\em CVPR}, pages 14713--14723, 2022.

\bibitem{cvpr2023MaskedVD}
R.~Wang, D.~Chen, Z.~Wu, Y.~Chen, X.~Dai, M.~Liu, L.~Yuan, and Y.-G. Jiang.
\newblock Masked video distillation: Rethinking masked feature modeling for
  self-supervised video representation learning.
\newblock In {\em CVPR}, pages 6312--6322, 2023.

\bibitem{Wang2022ACL}
S.~Wang, J.~Gao, Z.~Li, J.~Sun, and W.~Hu.
\newblock A closer look at self-supervised lightweight vision transformers.
\newblock {\em ArXiv}, 2022.

\bibitem{wang2019smiles}
S.~Wang, Y.~Guo, Y.~Wang, H.~Sun, and J.~Huang.
\newblock Smiles-bert: large scale unsupervised pre-training for molecular
  property prediction.
\newblock In {\em ICBCB}, pages 429--436, 2019.

\bibitem{2022BEiTV3}
W.~Wang, H.~Bao, L.~Dong, J.~Bjorck, Z.~Peng, Q.~Liu, K.~Aggarwal, O.~Mohammed,
  S.~Singhal, S.~Som, and F.~Wei.
\newblock Image as a foreign language: Beit pretraining for all vision and
  vision-language tasks.
\newblock {\em ArXiv}, 2022.

\bibitem{2023FreMAE}
W.~Wang, J.~Wang, C.~Chen, J.~Jiao, L.~Sun, Y.~Cai, S.~Song, and J.~Li.
\newblock Fremae: Fourier transform meets masked autoencoders for medical image
  segmentation.
\newblock 2023.

\bibitem{Wang2022ImagesSI}
X.~Wang, W.~Wang, Y.~Cao, C.~Shen, and T.~Huang.
\newblock Images speak in images: A generalist painter for in-context visual
  learning.
\newblock {\em CVPR}, pages 6830--6839, 2022.

\bibitem{acmmm2022fmnet}
Y.~Wang, Z.~Pan, X.~Li, Z.~CAO, K.~Xian, and J.~Zhang.
\newblock Less is more: Consistent video depth estimation with masked frames
  modeling.
\newblock {\em ArXiv}, 2022.

\bibitem{Wei2021MaskedFP}
C.~Wei, H.~Fan, S.~Xie, C.~Wu, A.~L. Yuille, and C.~Feichtenhofer.
\newblock Masked feature prediction for self-supervised visual pre-training.
\newblock {\em CVPR}, pages 14648--14658, 2021.

\bibitem{Wei2022MVPMV}
L.~Wei, L.~Xie, W.~gang Zhou, H.~Li, and Q.~Tian.
\newblock Mvp: Multimodality-guided visual pre-training.
\newblock {\em ArXiv}, 2022.

\bibitem{Wei2022ContrastiveLR}
Y.~Wei, H.~Hu, Z.~Xie, Z.~Zhang, Y.~Cao, J.~Bao, D.~Chen, and B.~Guo.
\newblock Contrastive learning rivals masked image modeling in fine-tuning via
  feature distillation.
\newblock {\em ArXiv}, 2022.

\bibitem{wen2023road}
L.~Wen, X.~Yang, D.~Fu, X.~Wang, P.~Cai, X.~Li, T.~Ma, Y.~Li, L.~Xu, D.~Shang,
  Z.~Zhu, S.~Sun, Y.~Bai, X.~Cai, M.~Dou, S.~Hu, B.~Shi, and Y.~Qiao.
\newblock On the road with gpt-4v(ision): Early explorations of visual-language
  model on autonomous driving, 2023.

\bibitem{Woo2023ConvNeXtV2}
S.~Woo, S.~Debnath, R.~Hu, X.~Chen, Z.~Liu, I.-S. Kweon, and S.~Xie.
\newblock Convnext v2: Co-designing and scaling convnets with masked
  autoencoders.
\newblock {\em ArXiv}, 2023.

\bibitem{Wu2022ObjectwiseMA}
J.~Wu and S.~Mo.
\newblock Object-wise masked autoencoders for fast pre-training.
\newblock {\em ArXiv}, 2022.

\bibitem{Wu2023DropMAEMA}
Q.~Wu, T.~Yang, Z.~Liu, B.~Wu, Y.~Shan, and A.~B. Chan.
\newblock Dropmae: Masked autoencoders with spatial-attention dropout for
  tracking tasks.
\newblock {\em CVPR}, 2023.

\bibitem{Wu2022DenoisingMA}
Q.~Wu, H.~Ye, Y.~Gu, H.~Zhang, L.~Wang, and D.~He.
\newblock Denoising masked autoencoders are certifiable robust vision learners.
\newblock {\em ArXiv}, 2022.

\bibitem{Wu2022ExtremeMF}
Z.~Wu, Z.~Lai, X.~Sun, and S.~Lin.
\newblock Extreme masking for learning instance and distributed visual
  representations.
\newblock {\em ArXiv}, 2022.

\bibitem{Wu2018UnsupervisedFL}
Z.~Wu, Y.~Xiong, S.~X. Yu, and D.~Lin.
\newblock Unsupervised feature learning via non-parametric instance-level
  discrimination.
\newblock {\em ArXiv}, 2018.

\bibitem{Xia2016AIDAB}
G.-S. Xia, J.~Hu, F.~Hu, B.~Shi, X.~Bai, Y.~Zhong, L.~Zhang, and X.~Lu.
\newblock Aid: A benchmark data set for performance evaluation of aerial scene
  classification.
\newblock {\em TGRS}, 55:3965--3981, 2016.

\bibitem{xia2022mole}
J.~Xia, C.~Zhao, B.~Hu, Z.~Gao, C.~Tan, Y.~Liu, S.~Li, and S.~Z. Li.
\newblock Mole-bert: Rethinking pre-training graph neural networks for
  molecules.
\newblock In {\em ICLR}, 2022.

\bibitem{Xiao2017FashionMNISTAN}
H.~Xiao, K.~Rasul, and R.~Vollgraf.
\newblock Fashion-mnist: a novel image dataset for benchmarking machine
  learning algorithms.
\newblock {\em ArXiv}, 2017.

\bibitem{5539970}
J.~Xiao, J.~Hays, K.~A. Ehinger, A.~Oliva, and A.~Torralba.
\newblock Sun database: Large-scale scene recognition from abbey to zoo.
\newblock In {\em 2010 IEEE Computer Society Conference on Computer Vision and
  Pattern Recognition}, pages 3485--3492, 2010.

\bibitem{Xiao2023MaskedIA}
Y.~Xiao, Z.~Tang, P.~Wei, C.~Liu, and L.~Lin.
\newblock Masked images are counterfactual samples for robust fine-tuning.
\newblock {\em CVPR}, 2023.

\bibitem{2022MFM}
J.~Xie, W.~Li, X.~Zhan, Z.~Liu, Y.~S. Ong, and C.~C. Loy.
\newblock Masked frequency modeling for self-supervised visual pre-training.
\newblock {\em ArXiv}, 2022.

\bibitem{Xie2022RevealingTD}
Z.~Xie, Z.~Geng, J.~Hu, Z.~Zhang, H.~Hu, and Y.~Cao.
\newblock Revealing the dark secrets of masked image modeling.
\newblock {\em ArXiv}, 2022.

\bibitem{Xie2021SimMIMAS}
Z.~Xie, Z.~Zhang, Y.~Cao, Y.~Lin, J.~Bao, Z.~Yao, Q.~Dai, and H.~Hu.
\newblock Simmim: a simple framework for masked image modeling.
\newblock {\em CVPR}, 2021.

\bibitem{Xie2022OnDS}
Z.~Xie, Z.~Zhang, Y.~Cao, Y.~Lin, Y.~Wei, Q.~Dai, and H.~Hu.
\newblock On data scaling in masked image modeling.
\newblock {\em ArXiv}, 2022.

\bibitem{Xu2022MaskedAA}
H.~Xu, S.~Ding, X.~Zhang, H.~Xiong, and Q.~Tian.
\newblock Masked autoencoders are robust data augmentors.
\newblock {\em ArXiv}, 2022.

\bibitem{Xue2022StareAW}
H.~Xue, P.~Gao, H.~Li, Y.~J. Qiao, H.~Sun, H.~Li, and J.~Luo.
\newblock Stare at what you see: Masked image modeling without reconstruction.
\newblock {\em ArXiv}, 2022.

\bibitem{Yan2023SkeletonMAEGM}
H.~Yan, Y.~Liu, Y.~Wei, Z.~Li, G.~Li, and L.~Lin.
\newblock Skeletonmae: Graph-based masked autoencoder for skeleton sequence
  pre-training.
\newblock {\em ArXiv}, 2023.

\bibitem{Yan2021VideoGPTVG}
W.~Yan, Y.~Zhang, P.~Abbeel, and A.~Srinivas.
\newblock Videogpt: Video generation using vq-vae and transformers.
\newblock {\em ArXiv}, 2021.

\bibitem{Yang2022motionmae}
H.~Yang, D.~Huang, B.~Wen, J.~Wu, H.~Yao, Y.~Jiang, X.~Zhu, and Z.~Yuan.
\newblock Self-supervised video representation learning with motion-aware
  masked autoencoders.
\newblock 2022.

\bibitem{Yang2023UniPADAU}
H.~Yang, S.~Zhang, D.~Huang, X.~Wu, H.~Zhu, T.~He, S.~Tang, H.~Zhao, Q.~Qiu,
  B.~Lin, X.~He, and W.~Ouyang.
\newblock Unipad: A universal pre-training paradigm for autonomous driving.
\newblock {\em ArXiv}, 2023.

\bibitem{yang2022masked}
K.~K. Yang, N.~Zanichelli, and H.~Yeh.
\newblock Masked inverse folding with sequence transfer for protein
  representation learning.
\newblock {\em bioRxiv}, 2022.

\bibitem{iccv2023MRM}
Q.~Yang, W.~Li, B.~Li, and Y.~Yuan.
\newblock Mrm: Masked relation modeling for medical image pre-training with
  genetics.
\newblock In {\em ICCV}, 2023.

\bibitem{Yang2022AttentiveCLIP}
Y.~Yang, W.~Huang, Y.~Wei, H.~Peng, X.~Jiang, H.~Jiang, F.~Wei, Y.~Wang, H.~Hu,
  L.~Qiu, and Y.~Yang.
\newblock Attentive mask clip.
\newblock 2022.

\bibitem{Yao2022MACRL}
Y.~Yao, N.~Desai, and M.~S. Palaniswami.
\newblock Masked contrastive representation learning.
\newblock {\em ArXiv}, 2022.

\bibitem{Yao2023MOMADF}
Y.~Yao, N.~Desai, and M.~S. Palaniswami.
\newblock Moma: Distill from self-supervised teachers.
\newblock {\em ArXiv}, 2023.

\bibitem{2022ConMAE}
K.~Yi, Y.~Ge, X.~Li, S.~Yang, D.~Li, J.~Wu, Y.~Shan, and X.~Qie.
\newblock Masked image modeling with denoising contrast.
\newblock {\em ArXiv}, 2022.

\bibitem{you2022cross}
Y.~You and Y.~Shen.
\newblock Cross-modality and self-supervised protein embedding for
  compound--protein affinity and contact prediction.
\newblock {\em Bioinformatics}, 38:ii68--ii74, 2022.

\bibitem{cvpr2023MAGVIT}
L.~Yu, Y.~Cheng, K.~Sohn, J.~Lezama, H.~Zhang, H.~Chang, A.~G. Hauptmann, M.-H.
  Yang, Y.~Hao, I.~Essa, and L.~Jiang.
\newblock Magvit: Masked generative video transformer.
\newblock In {\em CVPR}, 2023.

\bibitem{Yu2023SPAESP}
L.~Yu, Y.~Cheng, Z.~Wang, V.~Kumar, W.~Macherey, Y.~Huang, D.~A. Ross, I.~Essa,
  Y.~Bisk, M.~Yang, K.~P. Murphy, A.~G. Hauptmann, and L.~Jiang.
\newblock Spae: Semantic pyramid autoencoder for multimodal generation with
  frozen llms.
\newblock {\em ArXiv}, 2023.

\bibitem{Yu2020GradientSF}
T.~Yu, S.~Kumar, A.~Gupta, S.~Levine, K.~Hausman, and C.~Finn.
\newblock Gradient surgery for multi-task learning.
\newblock {\em ArXiv}, 2020.

\bibitem{cvpr2022pointbert}
X.~Yu, L.~Tang, Y.~Rao, T.~Huang, J.~Zhou, and J.~Lu.
\newblock Point-bert: Pre-training 3d point cloud transformers with masked
  point modeling.
\newblock In {\em CVPR}, 2022.

\bibitem{Yue2023ObjectRA}
K.~Yue, B.-C. Chen, J.~Geiping, H.~Li, T.~Goldstein, and S.-N. Lim.
\newblock Object recognition as next token prediction.
\newblock 2023.

\bibitem{Zhai2023MaskedAA}
J.-T. Zhai, X.~Liu, A.~D. Bagdanov, K.-C. Li, and M.-M. Cheng.
\newblock Masked autoencoders are efficient class incremental learners.
\newblock {\em ArXiv}, 2023.

\bibitem{Zhai2022PositionPA}
S.~Zhai, N.~Jaitly, J.~Ramapuram, D.~Busbridge, T.~Likhomanenko, J.~Y. Cheng,
  W.~A. Talbott, C.~Huang, H.~Goh, and J.~M. Susskind.
\newblock Position prediction as an effective pretraining strategy.
\newblock In {\em ICML}, 2022.

\bibitem{Zhai2021ScalingVT}
X.~Zhai, A.~Kolesnikov, N.~Houlsby, and L.~Beyer.
\newblock Scaling vision transformers.
\newblock {\em CVPR}, pages 1204--1213, 2021.

\bibitem{ijcai2023MIMsurvey}
C.~Zhang, C.~Zhang, J.~Song, J.~S.~K. Yi, K.~Zhang, and I.-S. Kweon.
\newblock A survey on masked autoencoder for self-supervised learning in vision
  and beyond.
\newblock In {\em IJCAI}, 2023.

\bibitem{Zhang2022iMAE}
K.~Zhang and Z.~Shen.
\newblock i-mae: Are latent representations in masked autoencoders linearly
  separable?
\newblock {\em ArXiv}, 2022.

\bibitem{Zhang2022HowMM}
Q.~Zhang, Y.~Wang, and Y.~Wang.
\newblock How mask matters: Towards theoretical understandings of masked
  autoencoders.
\newblock {\em ArXiv}, 2022.

\bibitem{Zhang2022PointM2AEMM}
R.~Zhang, Z.~Guo, P.~Gao, R.~Fang, B.~Zhao, D.~Wang, Y.~J. Qiao, and H.~Li.
\newblock Point-m2ae: Multi-scale masked autoencoders for hierarchical point
  cloud pre-training.
\newblock {\em ArXiv}, 2022.

\bibitem{cvpr2022Learning3R}
R.~Zhang, L.~Wang, Y.~J. Qiao, P.~Gao, and H.~Li.
\newblock Learning 3d representations from 2d pre-trained models via
  image-to-point masked autoencoders.
\newblock In {\em CVPR}, 2023.

\bibitem{zhang2022graph}
S.~Zhang, H.~Chen, H.~Yang, X.~Sun, P.~S. Yu, and G.~Xu.
\newblock Graph masked autoencoders with transformers.
\newblock {\em arXiv preprint}, 2022.

\bibitem{iclr2023ccMIM}
S.~Zhang, F.~Zhu, R.~Zhao, and J.~Yan.
\newblock Contextual image masking modeling via synergized contrasting without
  view augmentation for faster and better visual pretraining.
\newblock In {\em ICLR}, 2023.

\bibitem{Zhang2022CAEVC}
X.~Zhang, J.~Chen, J.~Yuan, Q.~Chen, J.~Wang, X.~Wang, S.~Han, X.~Chen, J.~Pi,
  K.~Yao, J.~Han, E.~Ding, and J.~Wang.
\newblock Cae v2: Context autoencoder with clip target.
\newblock {\em ArXiv}, 2022.

\bibitem{Zhang2022IntegrallyMP}
X.~Zhang, F.~Liu, Z.~Peng, Z.~Guo, F.~Wan, X.-W. Ji, and Q.~Ye.
\newblock Integrally migrating pre-trained transformer encoder-decoders for
  visual object detection.
\newblock 2022.

\bibitem{Zhang2022HiViTHV}
X.~Zhang, Y.~Tian, W.~Huang, Q.~Ye, Q.~Dai, L.~Xie, and Q.~Tian.
\newblock Hivit: Hierarchical vision transformer meets masked image modeling.
\newblock {\em ArXiv}, 2022.

\bibitem{Zhang2023MetaTransformerAU}
Y.~Zhang, K.~Gong, K.~Zhang, H.~Li, Y.~J. Qiao, W.~Ouyang, and X.~Yue.
\newblock Meta-transformer: A unified framework for multimodal learning.
\newblock {\em ArXiv}, 2023.

\bibitem{zhang2022protein}
Z.~Zhang, M.~Xu, A.~Jamasb, V.~Chenthamarakshan, A.~Lozano, P.~Das, and
  J.~Tang.
\newblock Protein representation learning by geometric structure pretraining.
\newblock In {\em ICLR}, 2023.

\bibitem{iccv2023mrt}
Z.~Zhao, S.~Wei, Q.~Chen, D.~Li, Y.~Yang, Y.~Peng, and Y.~Liu.
\newblock Masked retraining teacher-student framework for domain adaptive
  object detection.
\newblock In {\em ICCV}, pages 19039--19049, 2023.

\bibitem{Zheng2022CIMCI}
X.~Zheng, X.~Ma, and C.~Wang.
\newblock Cim: Constrained intrinsic motivation for sparse-reward continuous
  control.
\newblock {\em ArXiv}, 2022.

\bibitem{iccv2023sparsemae}
A.~Zhou, Y.~Li, Z.~Qin, J.~Liu, J.~Pan, R.~Zhang, R.~Zhao, P.~Gao, and H.~Li.
\newblock Sparsemae: Sparse training meets masked autoencoders.
\newblock In {\em ICCV}, pages 16176--16186, 2023.

\bibitem{Zhou2016PlacesAI}
B.~Zhou, A.~Khosla, {\`A}.~Lapedriza, A.~Torralba, and A.~Oliva.
\newblock Places: An image database for deep scene understanding.
\newblock {\em ArXiv}, 2016.

\bibitem{8100027}
B.~Zhou, H.~Zhao, X.~Puig, S.~Fidler, A.~Barriuso, and A.~Torralba.
\newblock Scene parsing through ade20k dataset.
\newblock In {\em CVPR}, pages 5122--5130, 2017.

\bibitem{iclr2022ibot}
J.~Zhou, C.~Wei, H.~Wang, W.~Shen, C.~Xie, A.~Yuille, and T.~Kong.
\newblock ibot: Image bert pre-training with online tokenizer.
\newblock {\em ICLR}, 2022.

\bibitem{Zhou2022SelfPW}
L.~Zhou, H.~Liu, J.~Bae, J.~He, D.~Samaras, and P.~Prasanna.
\newblock Self pre-training with masked autoencoders for medical image
  analysis.
\newblock {\em ArXiv}, 2022.

\bibitem{ieee2023MAEsurvey}
Z.~Zhou and X.~Liu.
\newblock Masked autoencoders in computer vision: A comprehensive survey.
\newblock {\em IEEE Access}, 2023.

\bibitem{Zhu2023VLGPT}
J.~Zhu, X.~Ding, Y.~Ge, Y.~Ge, S.~Zhao, H.~Zhao, X.~Wang, and Y.~Shan.
\newblock Vl-gpt: A generative pre-trained transformer for vision and language
  understanding and generation.
\newblock 2023.

\end{thebibliography}
}

%

\let\oldaddcontentsline\addcontentsline
\renewcommand{\addcontentsline}[3]{}
\let\addcontentsline\oldaddcontentsline




\renewcommand\thefigure{A\arabic{figure}}
\renewcommand\thetable{A\arabic{table}}
\setcounter{table}{0}
\setcounter{figure}{0}

\newpage

\onecolumn  

\if\submission\submissionTPAMI  
    \section*{Acknowledgments}
    This work was supported by the National Key R\&D Program of China (No. 2022ZD0115100), the National Natural Science Foundation of China Project (No. U21A20427), and Project (No. WU2022A009) from the Center of Synthetic Biology and Integrated Bioengineering of Westlake University.
    This work was done by Luyuan Zhang and Zedong Wang during their internship at Westlake University.

    \section{Appendix}
    \label{app}
    \vspace{-0.5em}
    In the Appendix sections, we provide detailed information on MIM methods for fundamental pre-training in Table~\ref{tab:full_mim_categoty} and CV downstream tasks in Table~\ref{tab:cv_downstream}.
%
\else  
%
    \section{Appendix}
    \label{app}
    In the Appendix sections, we provide detailed information on MIM methods for fundamental pre-training in Table~\ref{tab:full_mim_categoty} and CV downstream tasks in Table~\ref{tab:cv_downstream}, and datasets for masked modeling tasks in Table~\ref{tab:dataset}.
\fi

\begin{table*}[ht]
\setlength{\tabcolsep}{0.6mm}
\resizebox{\linewidth}{!}{
    \begin{tabular}{lcccccccc}
    \toprule
                  & EVA\cite{Fang2022EVAET} & EVA-02\cite{Fang2023EVA02AV} & WSP\cite{Singh2023WSP} & Painter\cite{Wang2022ImagesSI} & ViT-G\cite{Zhai2021ScalingVT} & MAE(ViT-L)~\cite{he2022masked} & LVM\cite{bai2023sequential} & InternVL\cite{Chen2023InternVLSU}
                  \\ \hline
Layer             & 40                      & 24                           & 24                     & 24                             & 48                            & 16     & 26 &    48                    \\
Attention Head    & 16                      & 16                           & 32                     & 16                             & 16                            & 24    & 32 &  25                        \\
Parameters        & 1011M                   & 304M                         & 1.89B                  & 307M                           & 1.84B                         & 307M     & 3B & 5903M                     \\ \hline
Pre-training      & IN-21K, CC3M,           & IN-21K, CC3M,                & IN-1K, IN-Real,        & ADE20K                         & IN-1K                         & IN-1K    & UVD &  LAION-COCO,COYO                   \\
Dataset           & CC12M                   & CC12M                        &                        & NYUv2                          & JFT-3B                        & ADE20K   &  &   CC12M                  \\ \hline
Downstream        & ADE, COCO,              & ADE, COCO,                   & COCO, ObjectNet        & COCO, Rain,                    &          ObjectNet                     & COCO       & IN-1K &        IN-1K            \\
Dataset           & Object365, Kinitics     & Object365, Kinetics          & Kinetics               & SIDD                           &    Real                           &    & Kinetics &                ADE20K            \\ \hline
Segmentation      & 62.3 mIoU               & 63.8 mIoU                    & 51.8 mIoU              & 49.9 mIoU                      & -                             & 53.6 mIoU     & - & 58.9 mIoU                 \\
Detection         & 64.7 AP                 & 65.9 AP                      & 58.0 AP                & 72.2AP                         & -                             & 53.3AP     &  -&                 -    \\
Video Recognition & 89.8 acc                & -                            & 86.0 acc               & -                              & -                             & -        & - &                     71.5 acc  \\
Classification    & 84.0 acc                & 85.5 acc                     & 90.9 acc               & -                              & 84.86 acc                     & 87.8 acc        & - &            82.5 acc   \\ 
     \bottomrule
     \end{tabular}
     }
     \caption{Experimental details and results of vision foundation models. IN denotes ImageNet datasets. LVM only performs comparison experiments of visual prompting and lacks standard benchmark results.}
     \label{tab:scaling_up}
\end{table*}
\begin{table*}[ht]
\setlength{\tabcolsep}{0.8mm}
\resizebox{\linewidth}{!}{
    \begin{tabular}{lcccc}
    \toprule
Model                                 & Modality & Pre-trained Method & Pre-trained Dataset        & Downstream Task                                              \\ \hline
BEiT.v3\cite{2022BEiTV3}              & CV, NLP  & MIM, MLM           & IN-1K, ADE20K,             & Classification, Detection,                                   \\
                                      &          &                    & COCO, NLVR2                & Segmentation                                                 \\
MaskVLM\cite{Kwon2022MaskedVA}        & CV, NLP  & MIM,MLM,CL         & CC,COCO,                   & Image-Text Retrieval, Natural Language for Visual Reasoning, \\
                                      &          &                    & SBU, Flickr30K             & Visual Entailment, Visual Question Answering                 \\
FLIP\cite{Li2022FLIP}                 & CV, NLP  & MIM,CL,MLM         & LAION-5B, IN-1K,           & Classification, Image-Text Retrieval,                        \\
                                      &          &                    & COCO, Flickr30K            & Image Captioning, Visual Question Answering                  \\
A-CLIP\cite{Yang2022AttentiveCLIP}    & CV, NLP  & MIM,CL             & IN-1K, YFCC100M, COCO,     & Classification (Zero-shot),                                  \\
                                      &          &                    & Flickr30K, Aircraft, MNIST & Image-Text Retrieval                                         \\
VL-BERT\cite{Su2019VLBERTPO}          & CV, NLP  & MLM,MIM            & COCO, RefCOCO+, VCR        & Classification, Segmentation,                                \\
                                      &          &                    &                            & Visual Question Answering                                    \\
MaskCLIP\cite{2022MaskCLIP}           & CV, NLP  & MIM,MLM,CL         & IN-1K, ADE20K,             & Classification (Zero-shot),                                  \\
                                      &          &                    & COCO, Flickr30K            & Detection, Segmentation                                      \\
MaskGIT\cite{Chang2022MaskGITMG}      & CV, NLP  & MIM                & IN-1K                      & Image-Text Generation                                        \\
VL-GPT\cite{Zhu2023VLGPT}             & CV, NLP  & MIM                & CC3M,LAION-COCO,MMC4       & Image Generation, Text-to-Image Generation                   \\
DALLE\cite{Ramesh2021ZeroShotTG}      & CV, NLP  & MIM,MLM            & IN-1K, CC, COCO, CUB200    & Text-Image Generation                                        \\
LQAE\cite{Liu2023LanguageQA}          & CV, NLP  & MIM,MLM            & IN-1K                      & Text-Image Alignment                                         \\
SPAE\cite{Yu2023SPAESP}               & CV, NLP  & MLM,MIM            & IN-1K, Kinetics            & Text-Image Generation                                        \\
InstructCV \cite{Gan2023InstructCVIT} & CV, NLP  & MLM                & IN-1K, MSCOCO, ADE20K      & Text-Image Generation                                        \\
     \bottomrule
    \end{tabular}
    }
    \caption{Details of MIM methods with both image and text data modalities.}
    \label{tab:multimodality}
\end{table*}

\twocolumn  

\begin{table*}[ht]
\setlength{\tabcolsep}{0.5mm}
\resizebox{\linewidth}{!}{
    \begin{tabular}{lccccccccc}
    \toprule
Model                                    & Category & Type   & Mask              & Encoder             & Target               & MIM Head        & CL Head & Loss                & Publish    \\ \hline
iGPT~\cite{Chen2020GenerativePF}          & BTTM     & AR     & AR Mask          & Transformer         & Offline, Tokenizer   & Linear          & -       & CE                  & ICML'2020  \\
VL-BERT~\cite{Su2019VLBERTPO}             & BTTM     & AE     & Random           & Tansformer          & Tokenizer            & Linear          & -       & CE                  & ICLR'2020  \\
MST~\cite{Li2021MSTMS}                    & ATPM     & AE     & Attention        & Transformer         & Feature, Pixel       & MLP             & -       & CE, MSE             & NIPS'2021  \\
SplitMask~\cite{2021splitmask}            & BTTM     & AE     & Random           & Transformer         & Tokenizer            & -               & Softmax & CE                  & arXiv'2021 \\
BEiT~\cite{Bao2021BEiT}                   & BTTM     & AE     & Random           & Transformer         & Offline Tokenizer    & Linear          & -       & CE                  & ICLR'2022  \\
iBOT~\cite{iclr2022ibot}                  & BTTM     & AE     & Random           & Transformer         & Tokenizer            & MLP             & -       & CE                  & ICLR'2022  \\
data2vec~\cite{Baevski2022data2vecAG}     & BTFM     & AE     & Random           & Transformer         & Feature              & Linear          & -       & $\ell_1$            & ICML'2022  \\
ADIOS~\cite{shi2022adversarial}           & ATPM     & AE     & Adversarial      & ResNet, Transformer & Pixel                & MLP             & -       & MSE                 & ICML'2022  \\
MP3~\cite{Casas2021MP3AU}                 & BTFM     & AE     & Random           & Transformer         & Feature              & Linear          & -       & MSE                 & ICML'2022  \\
MAE~\cite{he2022masked}                   & BTPM     & AE     & Random           & Transformer         & Pixel                & Transformer     & -       & MSE                 & CVPR'2022  \\
SimMIM~\cite{Xie2021SimMIMAS}             & BTPM     & AE     & Random           & Transformer         & Pixel                & Linear          & -       & MSE                 & CVPR'2022  \\
MaskFeat~\cite{Wei2021MaskedFP}           & BTFM     & AE     & Random           & Transformer         & Feature              & Linear          & -       & MSE                 & CVPR'2022  \\
MaskGIT~\cite{Chang2022MaskGITMG}         & BTTM     & AR     & Random           & Transformer         & Tokenizer            & Transformer     & -       & CE                  & CVPR'2022  \\
AttMask~\cite{eccv2022attmask}            & ATFM     & AE     & Attention        & Transformer         & Feature              & Transformer     & -       & CE                  & ECCV'2022  \\
mc-BEiT~\cite{Li2022mcBEiTMD}             & BTTM     & AE     & Random           & Transformer         & Tokenizer            & MLP             & -       & CE                  & ECCV'2022  \\
BootMAE~\cite{Dong2022BootstrappedMA}     & BTPM     & AE     & Random           & Transformer         & Pixel, Feature       & Transformer     & -       & MSE                 & ECCV'2022  \\
SdAE~\cite{Chen2022SdAESM}                & BTPM     & AE     & Random           & Transformer         & Pixel                & Transformer     & -       & Cosine              & ECCV'2022  \\
MultiMAE~\cite{Bachmann2022MultiMAEMM}    & BTFM     & AE     & Random           & Transformer         & Feature              & Transformer     & -       & MSE                 & ECCV'2022  \\
CAE~\cite{Chen2022ContextAF}              & BTFM     & AE     & Random           & Transformer         & Feature              & Transformer     & -       & CE, MSE             & IJCV'2023  \\
CAE.v2~\cite{Zhang2022CAEVC}              & BTFM     & AE     & Random           & Transformer         & Feature              & FC              & -       & Cosine              & arXiv'2022 \\
SemMAE~\cite{2022SemMAE}                  & ATPM     & AE     & Semantic Guided  & Transformer         & Pixel                & Transformer     & -       & MSE                 & NIPS'2022  \\
TTT-MAE~\cite{Gandelsman2022TestTimeTW}   & BTPM     & AE     & Random           & Transformer         & Pixel                & Transformer     & -       & MSE                 & NIPS'2022  \\
GreenMIM~\cite{Huang2022GreenHV}          & BTPM     & AE     & Random           & Transformer         & Pixel                & Transformer     & -       & MSE                 & NIPS'2022  \\
ConvMAE~\cite{Gao2022ConvMAEMC}           & BCPM     & AE     & Random           & Transformer,CNN     & Pixel                & Transformer     & -       & MSE                 & NIPS'2022  \\
MSN~\cite{Assran2022MaskedSN}             & BTFC     & AE     & Random           & Transformer         & Feature              & -               & Softmax & CE                  & arXiv'2022 \\
RePre~\cite{Wang2022RePreIS}              & BTPM     & AE     & Random           & Transformer         & Pixel                & CNN Transformer & -       & MSE                 & arXiv'2022 \\
MACRL~\cite{Yao2022MACRL}                 & BTPM     & AE     & Random           & Transformer         & Pixel                & Transformer     & MLP     & InfoNCE, MSE        & arXiv'2022 \\
Unified-IO~\cite{lu2022UnifiedIO}         & BTFM     & AE     & Binary           & Transformer         & Feature              & Transformer     & -       & InfoNCE             & arXiv'2022 \\
UnMAE~\cite{Li2022UniformME}              & ATPM     & AE     & Uniform Sampling & Transformer         & Pixel                & Transformer     & -       & MSE                 & arXiv'2022 \\
SIM~\cite{Tao2022SiameseIM}               & BTFM     & AE     & Random           & Transformer         & Feature              & Transformer     & -       & MSE                 & arXiv'2022 \\
ExtreMA~\cite{Wu2022ExtremeMF}            & BTFC     & AE     & Random           & Transformer         & Feature              & -               & FC      & InfoNCE             & arXiv'2022 \\
LoMaR~\cite{chen2022efficient}            & ATPM     & AE     & Local Mask       & Transformer         & Pixel                & Transformer     & -       & MSE                 & arXiv'2022 \\
CMAE~\cite{2022CMAE}                      & ATPC     & AE     & Local Mask       & Transformer         & Pixel                & -               & FC      & InfoNCE, MSE        & arXiv'2022 \\
MaskCLIP~\cite{2022MaskCLIP}              & BTFB     & AE     & Random           & Transformer         & Feature              & Transformer     & FC      & InfoNCE, MSE        & arXiv'2022 \\
BEiT.v2~\cite{2022BEiTV2}                 & BTTM     & AE     & Random           & Transformer         & Offline Tokenizer    & Linear          & -       & CE                  & arXiv'2022 \\
BEiT.v3~\cite{2022BEiTV3}                 & BTTM     & AE     & Random           & Transformer         & Tokenizer            & Linear          & -       & CE                  & arXiv'2022 \\
DMAE~\cite{Wu2022DenoisingMA}             & BTPM     & AE     & Random           & Transformer         & Pixel                & Transformer     & -       & MSE                 & arXiv'2022 \\
MILAN~\cite{Hou2022MILAN}                 & ATFM     & AE     & Attention        & Transformer         & Feature              & Transformer     & -       & MSE                 & arXiv'2022 \\
MimCo~\cite{2022MimCo}                    & BTFC     & AE     & Random           & Transformer         & Feature              & -               & FC      & InfoNCE             & arXiv'2022 \\
dBOT~\cite{liu2022dBOT}                   & BTFM     & AE     & Random           & Transformer         & Feature              & Transformer     & -       & $\ell_1$            & arXiv'2022 \\
RC-MAE~\cite{Lee2022RCMAE}                & BTPM     & AE     & Random           & Transformer         & Pixel                & Transformer     & -       & MSE                 & arXiv'2022 \\
MaskDistill~\cite{2022maskdistill}        & BTFM     & AE     & Random           & Transformer         & Feature              & Transformer     & -       & $\ell_1$, Cosine    & arXiv'2022 \\
i-MAE~\cite{Zhang2022iMAE}                & ATPM     & AE     & Mixture          & Transformer         & Pixel                & Transformer     & -       & MSE                 & arXiv'2022 \\
CAE.V2~\cite{Zhang2022CAEVC}              & BTFM     & AE     & Random           & Transformer         & Feature              & FC              & -       & Cosine              & arXiv'2022 \\
FastMIM~\cite{Guo2022FastMIM}             & BTFM     & AE     & Random           & Transformer         & HOG Feature          & Transformer     & -       & MSE                 & arXiv'2022 \\
A-CLIP~\cite{Yang2022AttentiveCLIP}       & ATFC     & AE     & Attention        & Transformer         & Feature              & -               & FC      & InfoNCE             & arXiv'2022 \\
MixMIM~\cite{2022MixMIM}                  & ATPM     & AE     & Mixture          & Transformer         & Pixel                & Transformer     & -       & MSE                 & arXiv'2022 \\
MVP~\cite{Wei2022MVPMV}                   & BTTM     & AE     & Random           & Transformer         & Token                & Linear          & -       & CE                  & arXiv'2022 \\
FD~\cite{Wei2022ContrastiveLR}            & BTFM     & AE     & Random           & Transformer         & Feature              & FC              & -       & $\ell_1$            & arXiv'2022 \\
ObjMAE~\cite{Wu2022ObjectwiseMA}          & ATPM     & AE     & Hard Sampling    & Transformer         & Pixel                & Transformer     & -       & MSE                 & arXiv'2022 \\
SDMAE~\cite{Mao2022SDMAE}                 & ATFB     & AE     & Contextual       & Transformer         & Pixel, Feature       & Transformer     & FC      & InfoNCE, MSE        & arXiv'2022 \\
Ge2AE~\cite{Liu2022TheDI}                 & BTFB     & AE     & Random           & Transformer         & Fourier Feature      & Transformer     & FC      & Focal FFT, MSE      & AAAI'2023  \\
DILEMMA~\cite{Sameni2022RepresentationLB} & BTFM     & AE     & Random           & Transformer         & Feature              & Transformer     & -       & CE                  & AAAI'2023  \\
PeCo~\cite{Dong2021PeCoPC}                & BTTM     & AE     & Random           & Transformer         & Token                & Linear          & -       & CE                  & AAAI'2023  \\
data2vec2.0~\cite{2022Data2Vec2}          & ATFM     & AE     & Multi-Masking    & Transformer         & Feature              & CNN             & -       & MSE                 & ICML'2023  \\
A2MIM~\cite{2022a2mim}                    & BCFM     & AE     & Random           & Transformer, CNN    & Fourier, HOG Feature & Linear          & -       & $\ell_1$, Focal FFT & ICML'2023  \\
Hiera~\cite{icml2023Hiera}                & BTPM     & AE     & Random           & Transformer         & Pixel                & Transformer     & -       & MSE                 & ICML'2023  \\
MAE-Lite~\cite{Wang2022ACL}               & BTPM     & AE     & Random           & Transformer         & Pixel                & Transformer     & -       & MSE                 & ICML'2023  \\
ConMIM~\cite{2022ConMAE}                  & BTPC     & AE     & Random           & Transformer         & Pixel                & -               & FC      & InfoNCE             & ICLR'2023  \\
HiViT~\cite{Zhang2022HiViTHV}             & BTPM     & AE     & Random           & Transformer         & Pixel                & Transformer     & -       & MSE                 & ICLR'2023  \\
Layer Grafted~\cite{iclr2023layergrafted} & BTPC     & AE     & Random           & Transformer         & Pixel                & -               & FC      & InfoNCE, MSE        & ICLR'2023  \\
ccMIM~\cite{iclr2023ccMIM}                & ATPM     & AE     & Attention        & Transformer         & Pixel                & Transformer     & -       & MSE                 & ICLR'2023  \\
RandSAC~\cite{Hua2022SelfsupervisionTR}   & BTTM     & AR     & Random           & Transformer         & Tokenizer            & Transformer     & -       & CE                  & ICLR'2023  \\
Spark~\cite{Tian2023SparK}                & BCPM     & AE     & Random           & CNN                 & Pixel                & CNN             & -       & MSE                 & ICLR'2023  \\
CIM~\cite{Zheng2022CIMCI}                 & BCTM     & AE     & Random           & Transformer,CNN     & Tokenizer            & Transformer     & -       & CE                  & ICLR'2023  \\
MaskVLM~\cite{Kwon2022MaskedVA}           & BTPM     & AE     & Random           & Transformer         & Pixel, Feature       & Transformer     & -       & MSE                 & ICLR'2023  \\
ConvNext.v2~\cite{Woo2023ConvNeXtV2}      & BCPM     & AE     & Random           & CNN                 & Pixel                & CNN             & -       & MSE                 & CVPR'2023  \\
MAGE~\cite{cvpr2023mage}                  & BTTB     & AE, AR & Random           & Transformer         & Tokenizer            & Transformer     & MLP     & CE, InfoNCE         & CVPR'2023  \\
I-JEPA~\cite{cvpr2023IJEPA}               & ATPM     & AE     & Contextual       & Transformer         & Pixel                & Transformer     & -       & L2                  & CVPR'2023  \\
HPM~\cite{cvpr2023HPM}                    & ATPM     & AE     & Hard Sampling    & Transformer         & Pixel                & Transformer     & -       & MSE                 & CVPR'2023  \\
FLIP~\cite{Li2022FLIP}                    & BTFC     & AE     & Random           & Transformer         & Text, Feature        & -               & FC      & InfoNCE             & CVPR'2023  \\
AutoMAE \cite{Chen2023AutoMAE}            & ATPM     & AE     & Adversarial      & Transformer         & Pixel                & Transformer     & -       & MSE                 & CVPR'2023  \\
LocalMAE~\cite{Wang2023MaskedIM}          & BTFM     & AE     & Random           & Transformer         & Feature              & Transformer     & -       & MSE                 & CVPR'2023  \\
MaskAlign~\cite{Xue2022StareAW}           & ATFM     & AE     & Attention        & Transformer         & Feature              & MLP             & -       & MSE                 & CVPR'2023  \\
MFM~\cite{Liu2023ImprovingPM}             & BTFM     & AE     & Random           & Transformer         & Feature              & Transformer     & -       & MSE                 & ICCV'2023  \\
SparseMAE~\cite{iccv2023sparsemae}        & BTFM     & AE     & Random           & Transformer         & Pixel                & Transformer     & -       & MSE                 & ICCV'2023  \\
MFM~\cite{2022MFM}                        & BCFM     & AE     & Random           & Transformer, CNN    & Fourier Feature      & Linear          & -       & Fourier Loss        & ICCV'2023  \\
SparseMAE~\cite{iccv2023sparsemae}        & BTPM     & AE     & Random           & Transformer         & Pixel                & Transformer     & -       & MSE                 & ICCV'2023  \\
RobustMAE~\cite{Huang2023ImprovingAR}     & BTFM     & AE     & Random           & Transformer         & Feature              & Transformer     & -       & CE                  & ICCV'2023  \\
CAN~\cite{Mishra2022ASE}                  & BTPB     & AE     & Random           & Transformer         & Pixel                & Transformer     & FC      & InfoNCE, MSE        & ICCV'2023  \\
    \bottomrule
    \end{tabular}
    }
    \caption{Detailed information of fundamental masked image modeling (MIM) methods  (\textbf{view Table~\ref{tab:full_mim_categoty_1} to continue}).}
    \label{tab:full_mim_categoty}
\end{table*}

\begin{table*}[ht]
\setlength{\tabcolsep}{0.8mm}
\resizebox{\linewidth}{!}{
    \begin{tabular}{lccccccccc}
    \toprule
Model                                     & Category & Type   & Mask             & Encoder             & Target               & MIM Head        & CL Head & Loss                & Publish    \\ \hline
DropPos~\cite{Wang2023DropPosPV}          & BTFM     & AE     & Random           & Transformer         & Feature              & MLP             & -       & CE                  & NIPS'2023  \\
RevColV2~\cite{Han2023RevColV2}           & BTPM     & AE     & Random           & Transformer         & Pixel                & Transformer     & -       & MSE                 & NIPS'2023  \\
MaPeT~\cite{Baraldi2023LearningTM}        & BTTM     & AE, AR & Random   & Transformer & Tokenizer      & Transformer & -       & Likehood & arXiv'2023 \\
R-MAE~\cite{Nguyen2023RMAERM}             & BCPM     & AE     & Random   & Transformer & Pixel          & Transformer & -       & CE       & arXiv'2023 \\
DMJD~\cite{Ma2022DisjointMW}              & ATFM     & AE     & Disjoint & Transformer & Feature        & Transformer & -       & MSE      & arXiv'2023 \\
MOMA~\cite{Yao2023MOMADF}                 & BTFC     & AE     & Random   & Transformer & Feature        & -           & FC      & InfoNCE  & arXiv'2023 \\
PixMIM~\cite{Liu2023PixMIMRP}             & BTFM     & AE     & Random   & Transformer & Feature        & Transformer & -       & MSE      & arXiv'2023 \\
TinyMIM~\cite{Ren2023TinyMIM}             & BTFM     & AE     & Random   & Transformer & Feature        & Transformer & -       & MSE      & arXiv'2023 \\
MSCN~\cite{Jing2022MaskedSC}              & BTFM     & AE     & Random   & Transformer & Feature        & MLP         & -       & MSE      & arXiv'2023 \\
Img2vec~\cite{pan2023img2vec}             & BTFM     & AE     & Random   & Transformer & Feature        & MLP         & -       & MSE      & arXiv'2023 \\
DeepMIM~\cite{ren2023rejuvenating}        & BTFM     & AE     & Random   & Transformer & Pixel, Feature & Transformer & -       & MSE      & arXiv'2023 \\
D-iGPT~\cite{ren2023rejuvenating}         & BTTB     & AE     & Random   & Transformer & Tokenizer      & Transformer & -       & CE       & arXiv'2023 \\
VL-GPT~\cite{Zhu2023VLGPT}                & BTTM     & AE     & Random   & Transformer & Tokenizer      & Transformer & -       & CE, MSE  & arXiv'2023 \\
LVM\cite{bai2023sequential}               & BTTM     & AR     & AR Mask  & Transformer & Tokenizer      & Transformer & -       & CE       & arXiv'2023 \\
    \bottomrule
    \end{tabular}
    }
    \caption{Detailed information of fundamental masked image modeling (MIM) methods (\textbf{continue Table}~\ref{tab:full_mim_categoty}).}
    \label{tab:full_mim_categoty_1}
\end{table*}

\begin{table*}[ht]
    \centering
\setlength{\tabcolsep}{0.6mm}
\resizebox{\linewidth}{!}{
    \begin{tabular}{lcccccccc}
    \toprule
Model                                      & Task                    & Type   & Category & Mask                   & Encoder     & Target          & Head           & Publication \\ \hline
MIMDet~\cite{Fang2022UnleashingVV}         & Detection               & AE     & RTTM     & Random                 & Transformer & Token           & MIM Head       & arXiv'2022  \\
iTPN~\cite{Tian2022IntegrallyPT}           & Detection, Segmentation & AE     & BTFM     & Random                 & Transformer & Feature         & MIM Head       & CVPR'2023   \\
imTED~\cite{Zhang2022IntegrallyMP}         & Detection               & AE     & BTFM     & Random                 & Transformer & Feature         & MIM Head       & CVPR'2023   \\
PiMAE~\cite{cvpr2023PiMAE}                 & Detection               & AE     & BTFM     & Random                 & Transformer & Feature         & MIM Head       & ICCV'2023   \\
MRT~\cite{iccv2023mrt}                     & Detection               & AE     & ATFM     & Hard Sampling          & Transformer & Feature         & MIM Head       & ICCV'2023   \\
NXTP~\cite{Yue2023ObjectRA}                & Detection               & AR     & BTTM     & AR Mask                & Transformer & Token           & MIM Head       & arXiv'2023  \\
FreMAE~\cite{2023FreMAE}                   & Medical Image           & AE     & BTFM     & Random                 & Transformer & Fourier Feature & MIM Head       & arXiv'2023  \\
G2SD~\cite{cvpr2023G2SD}                   & KD                      & AE     & BTFM     & Random                 & Transformer & Feature         & MIM Head       & CVPR'2023   \\
MKD~\cite{iccv2023maekd}                   & KD                      & AE     & BTFM     & Random                 & Transformer & Feature         & MIM Head       & ICCV'2023   \\
VideoGPT~\cite{Yan2021VideoGPTVG}          & Video                   & AR     & BTTM     & AR Mask                & Transformer & Token           & MIM Head       & arXiv'2021  \\
BEVT~\cite{cvpr2022BEVT}                   & Video                   & AE     & BTTM     & Random                 & Transformer & Token           & MIM Head       & CVPR'2022   \\
MAE~\cite{Feichtenhofer2022MaskedAA}       & Video                   & AE     & BTPM     & Random                 & Transformer & Pixel           & MIM Head       & NIPS'2022   \\
VideoMAE~\cite{Tong2022VideoMAEMA}         & Video                   & AE     & BTPM     & Random                 & Transformer & Pixel           & MIM Head       & NIPS'2022   \\
FMNet~\cite{acmmm2022fmnet}                & Video                   & AE     & BTFM     & Random                 & Tranformer  & Feature         & MIM Head       & ACMMM'2022  \\
MILES~\cite{2022MILES}                     & Video                   & AE     & ATFM     & Contextual             & Transformer & Feature         & MIM Head       & arXiv'2022  \\
MAR~\cite{2022MAR}                         & Video                   & AE     & ATPM     & Cell Running           & Transformer & Pixel           & MIM Head       & arXiv'2022  \\
OmniMAE~\cite{2022OmniMAE}                 & Video                   & AE     & BTPM     & Random                 & Transformer & Pixel           & MIM Head       & arXiv'2022  \\
MotionMAE~\cite{Yang2022motionmae}         & Video                   & AE     & BTPM     & Random                 & Transformer & Pixel           & MIM Head       & arXiv'2022  \\
MAM2~\cite{2022MAM2}                       & Video                   & AE     & BTTM     & Random                 & Transformer & Token           & MIM Head       & arXiv'2022  \\
MaskViT~\cite{Gupta2022MaskViTMV}          & Video                   & AE, AR & BTTM     & Random                 & Transformer & Token           & MIM Head       & CVPR'2023   \\
DropMAE~\cite{Wu2023DropMAEMA}             & Video                   & AE     & BTPM     & Random                 & Transformer & Pixel           & MIM Head       & CVPR'2023   \\
MAGVIT~\cite{cvpr2023MAGVIT}               & Video                   & AE, AR & BTTM     & Random                 & Transformer & Token           & MIM Head       & CVPR'2023   \\
AdaMAE~\cite{CVPR2023AdaMAE}               & Video                   & AE     & BTPM     & Random                 & Transformer & Pixel           & MIM Head       & CVPR'2023   \\
VideoMAE.v2~\cite{cvpr2023VideoMAEV2}      & Video                   & AE     & BTPM     & Random                 & Transformer & Pixel           & MIM Head       & CVPR'2023   \\
MVD~\cite{cvpr2023MaskedVD}                & Video                   & AE     & BTPM     & Random                 & Transformer & Pixel, Feature  & MIM Head       & CVPR'2023   \\
MGMAE~\cite{Huang2023MGMAEMG}              & Video                   & AE     & BTFM     & Random                 & Transformer & Feature         & MIM Head       & ICCV'2023   \\
Forecast-MAE~\cite{Cheng2023ForecastMAESP} & Video                   & AE     & BTFM     & Random                 & Transformer & Feature         & MIM Head       & ICCV'2023   \\
Traj-MAE~\cite{Chen2023TrajMAEMA}          & Video                   & AE     & BTFM     & Random                 & Transformer & Feature         & MIM Head       & ICCV'2023   \\
MGM~\cite{Fan2023MotionGuidedMF}           & Video                   & AE     & ATPM     & Motion Guided          & Transformer & Pixel           & MIM Head       & ICCV'2023   \\
HumanMAC~\cite{Mao2023MaskedMP}            & Video                   & AE     & BTFM     & Random                 & Transformer & Feature         & MIM Head       & ICCV'2023   \\
SkeletonMAE~\cite{Yan2023SkeletonMAEGM}    & Video                   & AE     & ATFM     & Joint Mask             & Transformer & Feature         & MIM Head       & ICCV'2023   \\
MAMP~\cite{Chen2023HumanMACMM}             & Video                   & AE     & ATFM     & Motion Aware           & Transformer & Feature         & MIM Head       & ICCV'2023   \\
GeoMIM~\cite{Liu2023TowardsB3}             & Video                   & AE     & BTFM     & Random                 & Transformer & Feature         & MIM Head       & ICCV'2023   \\
SiamMAE~\cite{Gupta2023SiamMAE}            & Video                   & AE     & BTPM     & Random                 & Transformer & Pixel           & MIM Head       & arXiv'2023  \\
CMAE-V~\cite{Lu2023CMAEVCM}                & Video                   & AE     & BTPB     & Random                 & Transformer & Pixel           & CL \& MIM Head & arXiv'2023  \\
MRM~\cite{iccv2023MRM}                     & Medical Image           & AE     & ATPM     & Relation Mask          & Transformer & Pixel           & MIM Head       & ICCV'2023   \\
SD-MAE~\cite{Mao2022SDMAE}                 & Medical Image           & AE     & BTPM     & Random                 & Transformer & Pixel           & MIM Head       & arXiv'2022  \\
MedMAE~\cite{Zhou2022SelfPW}               & Medical Image           & AE     & BTPM     & Random                 & Transformer & Pixel           & MIM Head       & arXiv'2022  \\
GCMAE~\cite{2022gcmae}                     & Medical Image           & AE     & BTPM     & Random                 & Transformer & Pixel           & MIM Head       & arXiv'2022  \\
SatMAE~\cite{2022SatMAE}                   & Remote Sensing          & AE     & BTPM     & Consistent Independent & Transformer & Pixel           & MIM Head       & arXiv'2022  \\
Scale-MAE~\cite{Reed2022ScaleMAEAS}        & Remote Sensing          & AE     & BTPM     & Random                 & Transformer & Pixel           & MIM Head       & ICCV'2023   \\
CMID~\cite{TGRS2023CMID}                   & Remote Sensing          & AE     & BTFB     & Random                 & Transformer & Fourier Feature & CL \& MIM Head & TGRS'2023   \\
DocMAE~\cite{icme2023DocMAE}               & OCR                     & AE     & BTPM     & Random                 & Transformer & Pixel           & MIM Head       & ICME'2023   \\
MGViT~\cite{Chen2022MaskguidedVT}          & Few Shot                & AE     & BTPM     & Random                 & Transformer & Pixel           & MIM Head       & NIPS'2022   \\
MeshMAE~\cite{eccv2022MeshMAE}             & 3D Mesh                 & AE     & BTPM     & Random                 & Transformer & Pixel           & MIM Head       & ECCV'2022   \\
VoxelMAE~\cite{2022VoxelMAE}               & 3D Point                & AE     & BTFM     & Random                 & Transformer & Voxel           & MIM Head       & arXiv'2022  \\
PointBERT~\cite{cvpr2022pointbert}         & 3D Point                & AE     & BTTM     & Random                 & Transformer & Token           & MIM Head       & CVPR'2022   \\
PointMAE~\cite{Pang2022MaskedAF}           & 3D Point                & AE     & BTFM     & Random                 & Transformer & Feature         & MIM Head       & ECCV'2022   \\
MaskPoint~\cite{Liu2022MaskedDF}           & 3D Point                & AE     & BTFM     & Random                 & Transformer & Real \& Fake    & MIM Head       & ECCV'2022   \\
Point-M2AE~\cite{Zhang2022PointM2AEMM}     & 3D Point                & AE     & BTPM     & Random                 & Transformer & Pixel           & MIM Head       & NIPS'2022   \\
PointCMP~\cite{cvpr2023PointCMP}           & 3D Point                & AE     & BTTB     & Random                 & Transformer & Token           & CL \& MIM Head & CVPR'2023   \\
I2P-MAE~\cite{cvpr2022Learning3R}          & 3D Point                & AE     & BTFM     & Random                 & Transformer & Feature         & MIM Head       & CVPR'2023   \\
GeoMAE~\cite{cvpr2023GeoMAE}               & 3D Point                & AE     & BTPM     & Random                 & Transformer & Pixel           & MIM Head       & CVPR'2023   \\
ACT~\cite{iclr2023act}                     & 3D Point                & AE     & BTFM     & Random                 & Transformer & Feature         & MIM Head       & ICLR'2023   \\
ReCon~\cite{Qi2023ContrastWR}              & 3D Point                & AE     & BTFB     & Random                 & Transformer & Feature         & CL \& MIM Head & ICML'2023   \\
MGM~\cite{Fan2023MotionGuidedMF}           & 3D Point                & AE     & BTPM     & Random                 & Transformer & Pixel           & MIM Head       & ICCV'2023   \\
    
    \bottomrule
    \end{tabular}
    }
    \caption{Detailed information of MIM methods for vision downstream tasks.}
    \label{tab:cv_downstream}
\end{table*}

\if\submission\submissionarXiv  
    \newpage
    \begin{table*}[ht]
    \centering
\setlength{\tabcolsep}{1.0mm}
\resizebox{\linewidth}{!}{
    \begin{tabular}{lcccccc}
    \toprule
Dataset       & Modality   & Type           & Pre-training & Downstream Task              & Training Set & Link                                                                                                                                                \\ \hline

ImageNet-1K\cite{Russakovsky2014ImageNetLS}      & CV         & Image          & CL MIM       & Classification               & 1,281,167   & \href{http://www.image-net.org/challenges/LSVRC/2012/}{ImageNet}                                                                                    \\
COCO 2014 Detection\cite{Lin2014MicrosoftCC}         & CV         & Image          & CL MIM       & Detection, Segmentation      & 83000      & \href{https://cocodataset.org/\#home}{COCO2014}                                                                                                         \\
COCO 2017 Detection\cite{Lin2014MicrosoftCC} & CV & Image & CL MIM & Detection, Segmentation & 118,000 &\href{https://cocodataset.org/dataset/detection-2017.htm}{COCO2017} \\

PASCAL Content  & CV & Image & CL MIM & Segmentation & 4998 & \href{https://www.cs.stanford.edu/~roozbeh/pascal-context/}{PASCAL Content} \\ 

MNIST\cite{Wang2022OracleMNISTAR}         & CV         & Image          & -            & Classification               & 60,000       & \href{http://yann.lecun.com/exdb/mnist/}{MNIST}                                                                                                     \\
Cityscapes\cite{Cordts2016TheCD}    & CV         & Image          & CL           & Segmentation                 & 2975       & \href{https://www.cityscapes-dataset.com/dataset-overview/}{Cityscapes}                                                                             \\

Kinetics700\cite{Kay2017TheKH}     & CV         & Video          & CL, MIM      & Action Recognition           & 494,801      & \href{https://deepmind.com/research/open-source/kinetics}{Kinetics}                                                                                 \\
UCF101\cite{Soomro2012UCF101AD}        & CV         & Video          & CL, MIM      & Action Recognition           & 9,537       & \href{https://www.crcv.ucf.edu/data/UCF101.php}{UCF-101}                                                                                            \\

RareAct\cite{Miech2020RareActAV} & CV & Video & CL MIM & Action Recognition & 7,607 & \href{https://github.com/antoine77340/RareActhttps://github.com/antoine77340/RareAct}{RareAct}
\\

AID\cite{Xia2016AIDAB}           & CV         & Image          & CL, MIM     & Classification               & 10,000       & \href{https://captain-whu.github.io/AID/}{AID}                                                                                                      \\

PASCAL VOC 2007 \cite{pascal-voc-2007}  &  CV  &  Image &  CL,MIM & Classification, Detection &  5011 &  \href{http://host.robots.ox.ac.uk/pascal/VOC/voc2007/}{PASCAL VOC}
\\

Oxford 102 Folwers \cite{Nilsback08} &  CV & Image & CL & Classification & 2040 & \href{https://www.robots.ox.ac.uk/~vgg/data/flowers/102/}{Oxford 102 Flowers}\\

SUN397\cite{5539970} & CV  & Image & CL,MIM & Classification &  19,850 & \href{https://vision.princeton.edu/projects/2010/SUN/}{SUN397}\\

Tiny-ImageNet\cite{Lee2021VisionTF} & CV         & Image          & CL MIM       & Classification               & 100,000      & \href{http://cs231n.stanford.edu/tiny-imagenet-200.zip}{TinyIN}                                                                                     \\

CIFAR-10\cite{Krizhevsky2009LearningML}      & CV         & Image          & CL           & Classification               & 50,000       & \href{https://www.cs.toronto.edu/$\sim$kriz/cifar.html}{CIFAR-10}                                                                                   \\
CIFAR-100\cite{Krizhevsky2009LearningML} & CV & Image & CL & Classification & 50,000 &       \href{https://www.cs.toronto.edu/~kriz/cifar.html}{CIFAR-100}                                                                        \\

STL-10\cite{Coates2011AnAO}        & CV         & Image          & CL MIM       & Classification               & 1,000      & \href{https://cs.stanford.edu/$\sim$acoates/stl10/}{STL}                                                                                            \\
CUB-200-2011\cite{wah2011caltech}  & CV         & Image          & CL MIM       & Classification               & 11,788       & \href{http://www.vision.caltech.edu/datasets/}{CUB-200-2011}                                                                                        \\
FGVC-Aircraft\cite{Maji2013FineGrainedVC} & CV         & Imgae          & CL MIM       & Classification               &  6,770       & \href{https://www.robots.ox.ac.uk/$\sim$vgg/data/fgvc-aircraft/}{Aircraft}                                                                          \\
StanfordCars\cite{6755945}  & CV         & Image          & CL MIM       & Classification               & 8,144       & \href{https://ai.stanford.edu/$\sim$jkrause/cars/car\_dataset.html}{StanfordCars}                                                                   \\
Places205\cite{Zhou2016PlacesAI}     & CV         & Image          & CL MIM       & Recognition                  & 2,500,000    & \href{http://places.csail.mit.edu/downloadData.html}{Places205}                                                                                     \\
iNaturalist\cite{Horn2017TheIC}   & CV         & Image          & CL MIM       & Classification               & 675,170      & \href{https://github.com/visipedia/inat\_comp/tree/master/2017}{iNaturalist}                                                                        \\
AgeDB\cite{8014984}         & CV         & Image          & MIM          & Age Estimation               & 16,488       & \href{https://ibug.doc.ic.ac.uk/resources/agedb/}{AgeDB}                                                                                            \\
Fashion-MNIST\cite{Xiao2017FashionMNISTAN} & CV         & Image          & MIM          & Classification               & 70,000       & \href{https://github.com/zalandoresearch/fashion-mnist}{Fashion-MNIST}                                                                              \\

KITTI-360\cite{Liao2021KITTI360AN}  & CV         & 3D Point Cloud & CL MIM       & Detection, Segmentation      & 43552        & \href{https://www.cvlibs.net/datasets/kitti/}{KITTI Vision}                                                                                         \\
ShapeNet\cite{Chang2015ShapeNetAI}      & CV         & 3D PointCloud  & CL MIM       & Recognition, Classification  & 220,000      & \href{https://www.shapenet.org/}{ShapeNet}                                                                                                          \\

Caltech-101\cite{1384978} & CV & Image & CL MIM & Classification & 3060   &  \href{https://data.caltech.edu/records/mzrjq-6wc02}{Caltech-101}\\
Charades\cite{Sigurdsson2016HollywoodIH}      & CV         & Video          & CL MIM       & Recognition                  & 66,500       & \href{http://vuchallenge.org/charades.html}{Charades}                                                                                               \\
AVA\cite{Gu2017AVAAV}           & CV         & Video          & CL MIM       & Detection                    &   211,000           & \href{http://research.google.com/ava/}{AVA}      \\

LVIS \cite{Gupta2019LVISAD} & CV & Image  & CL MIM & Detection &118,000 & \href{https://www.lvisdataset.org}{LVIS} \\ 

CC12M\cite{Changpinyo2021Conceptual1P}         & CV, NLP    & Image, Text    & MM CL        & Classification               & 12,000,000   & \href{https://arxiv.org/pdf/2102.08981v1.pdf}{CC12M}                                                                                                \\
LAION-5B\cite{Schuhmann2022LAION5BAO}         & CV, NLP    & Image, Text    & MM CL        & Classification               & 400,000,000  & \href{https://laion.ai/laion-400-open-dataset/}{LAION}                                                                                              \\

Flickr30k\cite{} \cite{Carreira2019ASN} & CV, NLP & Image, Text & MM CL & Image-Text  Retrieval & 31783 & \href{https://shannon.cs.illinois.edu/DenotationGraph/}{Flickr30k} \\ 

COCO Caption & CV, NLP & Image, Text & MM CL & Image-Text  Retrieval & 82783 & \href{https://github.com/tylin/coco-caption}{COCO Caption} \\

LSMDC\cite{Rohrbach2015ADF}         & CV, NLP    & Video, Text    & MM CL        & Movie Description            & 118,081      & \href{https://sites.google.com/site/describingmovies/}{LSMDC} \\

ADE20K\cite{8100027}       & CV, NLP    & Image, Text    & CL, MIM      & Scene Parsing                & 20,000       & \href{https://groups.csail.mit.edu/vision/datasets/ADE20K/}{ADE-20K}                                                                                \\

TACoS\cite{Rohrbach2014CoherentMV}         & CV, NLP    & Text, Video    & CL, MM       & Detection                    & 2,600             & \href{https://www.mpi-inf.mpg.de/departments/computer-vision-and-machine-learning/research/vision-and-language/tacos-multi-level-corpus}{TACoS} \\

RACE\cite{Lai2017RACELR}          & NLP        & Text           & MLM          & Reading Comprehension        & 28,000       & \href{https://www.cs.cmu.edu/$\sim$glai1/data/race/}{RACE}                                                                                          \\
MS MARCO\cite{Campos2016MSMA}      & NLP        & Text           & MLM          & Question Answering           & 1,000,000    & \href{https://microsoft.github.io/msmarco/}{MSMAECO}                                                                                                \\

AudioSet\cite{7952261}      & Audio, NLP & Speech, Text   & MM, MLM      & Sound Classification         & 2,000,000    & \href{https://research.google.com/audioset/index.html}{AudioSet}                                                                                    \\
LibriSpeech\cite{7178964}   & Audio      & Speech         & MLM          & Speech Recognition           &         1,789,621     & \href{http://www.openslr.org/12}{LibriSpeech}                                                                                                   \\

    \bottomrule
    \end{tabular}
    }
    \caption{Summary of datasets for MIM pre-training and vision downstream tasks. \textcolor{magenta}{Link} to dataset websites is provided.}
    \label{tab:dataset}
\end{table*}

\fi


\end{document}